\documentclass[10pt,twocolumn,letterpaper]{article}
\usepackage[dvipsnames]{xcolor}

\usepackage{iccv}
\usepackage{times}
\usepackage{epsfig}
\usepackage{graphicx}
\usepackage{amsmath}
\usepackage{amssymb}

\usepackage{booktabs}
\usepackage{enumitem}
\usepackage{glossaries}
\usepackage{graphicx}
\usepackage{multirow}
\usepackage{tikz}
\usetikzlibrary{positioning}
\usepackage{colortbl}
\usepackage[symbol,hang,flushmargin]{footmisc}

\usepackage[pagebackref=true,breaklinks=true,colorlinks,bookmarks=false]{hyperref}

\definecolor{linksblue}{rgb}{0.16, 0.36, 0.69}
\definecolor{linksorange}{rgb}{0.95, 0.4, 0.}
\definecolor{mygray}{gray}{0.97}
\hypersetup{colorlinks, linkcolor=linksorange, citecolor=linksblue}

\usepackage[capitalize]{cleveref}
\crefname{section}{Sec.}{Secs.}
\Crefname{section}{Section}{Sections}
\Crefname{table}{Table}{Tables}

\iccvfinalcopy 
\ificcvfinal\fi

\newacronym{vit}{ViT}{Vision Transformer}
\newacronym{mhsa}{MHSA}{Multiheaded Self-Attention}
\newacronym{ffn}{FFNs}{Feed-Forward Neural Networks}
\newacronym{nlp}{NLP}{Natural Language Processing}
\newacronym{cv}{CV}{Computer Vision}
\newacronym{cnn}{CNN}{Convolutional Neural Network}

\usepackage{xspace}
\newcommand{\idxfine}{f}
\newcommand{\idxcoarse}{c}
\newcommand{\mixedwhat}{scale\xspace}
\newcommand{\mixedwhats}{scales\xspace}
\newcommand{\bsloss}{BaS}
\newcommand{\gbsloss}{GBaS}

\usepackage{pifont}
\definecolor{byzantine}{rgb}{0.74, 0.2, 0.64}
\definecolor{dollarbill}{rgb}{0.52, 0.73, 0.4}

\DeclareRobustCommand\tikzinlinecoarse{\begin{tikzpicture}[baseline=-2.8pt] \node[rectangle, line width=1.0, draw=orange!40, fill=yellow!15, minimum width=0.28cm, minimum height=0.28cm] () {}; \end{tikzpicture}\hspace{1pt}}

\DeclareRobustCommand\tikzinlinefine{\begin{tikzpicture}[baseline=-2.8pt] \node[rectangle, line width=1.0, draw=orange!40, fill=yellow!1, minimum width=0.2cm, minimum height=0.2cm] () {};\end{tikzpicture}\hspace{1pt}}

\begin{document}

\title{MSViT: Dynamic Mixed-\mixedwhat Tokenization for Vision Transformers}

\renewcommand{\thefootnote}{\fnsymbol{footnote}}
\author{Jakob Drachmann Havtorn\textsuperscript{$\dagger \ast$}\\
Technical University of Denmark \& Corti.ai\\
Copenhagen, Denmark\\
{\tt\small jakobhavtorn.github.io}
\and
Am\'elie Royer\textsuperscript{$\ast$}\\
Qualcomm AI Research\\
Amsterdam, The Netherlands\\
{\tt\small aroyer@qti.qualcomm.com } 
\and
Tijmen Blankevoort\\
Qualcomm AI Research\\
Amsterdam, The Netherlands\\
{\tt\small  tijmen@qti.qualcomm.com } 
\and
Babak Ehteshami Bejnordi\\
Qualcomm AI Research\\
Amsterdam, The Netherlands\\
{\tt\small behtesha@qti.qualcomm.com} 
}
\maketitle
\ificcvfinal\thispagestyle{empty}\fi

\footnotetext{\textsuperscript{$\ast$}equal contribution}
\footnotetext{\textsuperscript{$\dagger$}work done as an intern at Qualcomm AI Research}
\footnotetext{Qualcomm AI Research is an initiative of Qualcomm Technologies, Inc.}

\renewcommand{\thefootnote}{\arabic{footnote}}
\begin{abstract}
    The input tokens to Vision Transformers carry little semantic meaning as they are defined as regular equal-sized patches of the input image, regardless of its content.
    However, processing uniform background areas of an image should not necessitate as much compute as dense, cluttered areas. 
    To address this issue, we propose a dynamic mixed-scale tokenization scheme for ViT, MSViT. Our method introduces a conditional gating mechanism that selects the optimal token scale for every image region, such that the number of tokens is dynamically determined per input. 
   In addition, to enhance the conditional behavior of the gate during training, we introduce a novel generalization of the batch-shaping loss. We show that our gating module is able to learn meaningful semantics despite operating locally at the coarse  patch-level.
  The proposed gating module is lightweight, agnostic to the choice of transformer backbone, and trained within a few epochs with little training overhead. Furthermore, in contrast to token pruning, MSViT does not lose information about the input, thus can be readily applied for dense tasks. 
   We validate MSViT on the tasks of classification and segmentation where it leads to improved accuracy-complexity trade-off. 
\end{abstract}

\section{Introduction}
\label{sec:intro}
\begin{figure}
    \centering
    \includegraphics[width=0.49\textwidth]{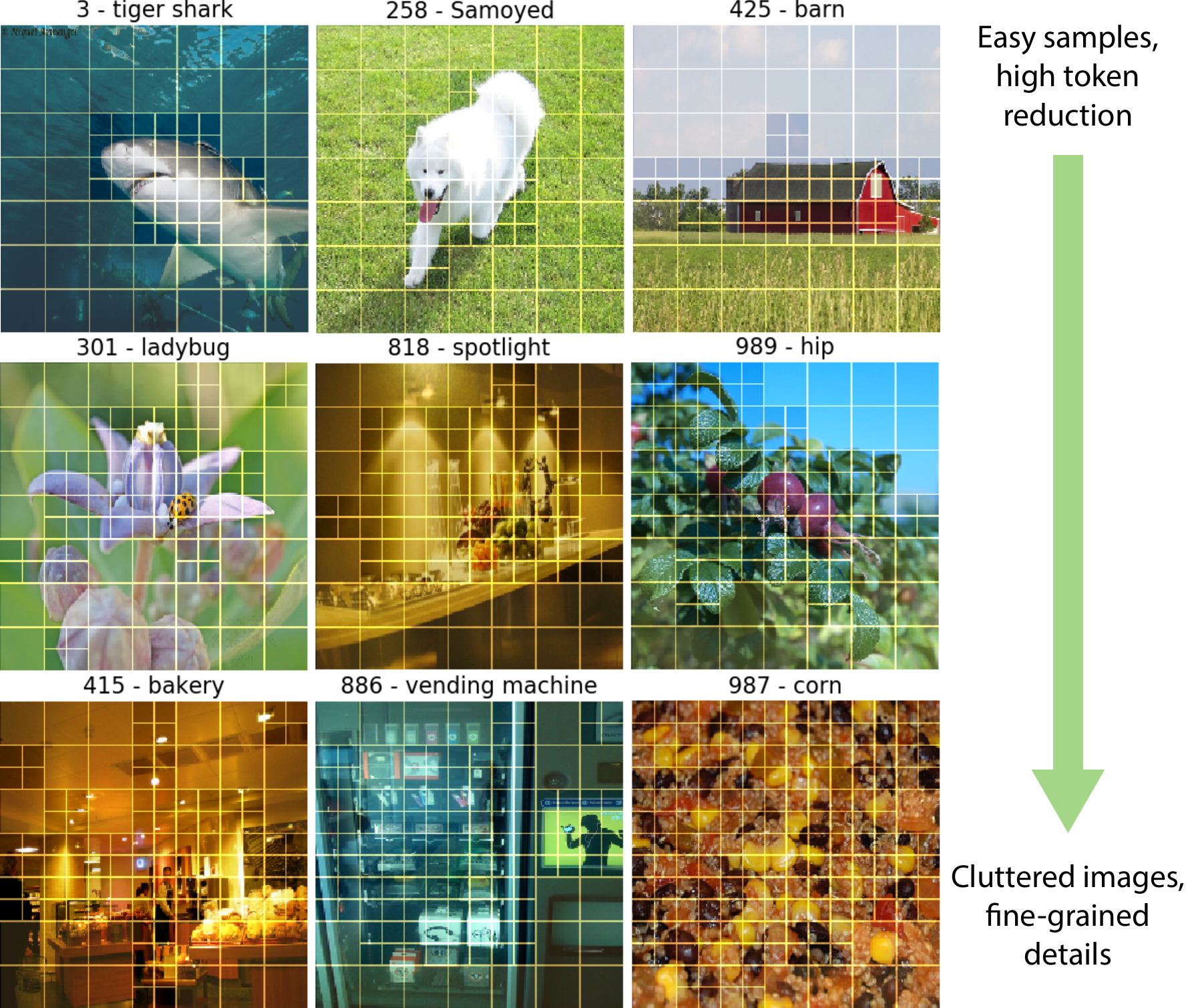}
    \caption{We introduce a learnable module to dynamically select the optimal token scale for each region. This module can be plugged in as a preprocessing step to any Vision Transformer. 
    Here we illustrate some mixed-scale masks on ImageNet samples with varying levels of clutter, output by the scale selection module, trained alongside a pretrained ViT-S/16 for 20 epochs to choose between a coarse 
    (32px, \tikzinlinecoarse{}) and a fine (16px, \tikzinlinefine{})	token scale.
    \vspace{-0.4cm}
    }
    \label{fig:teaser}
\end{figure}
%
%
%
%
The Transformer architecture \cite{vaswani_attention_2017} has seen widespread success across Natural Language Processing (\textbf{NLP}) tasks and more recently in Computer Vision (\textbf{CV}) \cite{dosovitskiy_image_2021,liu2021Swin,touvron2021training}. 
However, the quadratic time and memory complexity of transformer-based models poses a challenge when deploying such models on compute constrained devices or scaling them to large image sizes.
In particular, the number of input tokens and the tokenization method are defining aspects of the computational complexity of transformers. 
In NLP, it is generally straightforward to use semantic units, such as words or sentences, as input  tokens: This leads to little redundancy in the information carried by individual tokens. 
Conversely, in CV, tokenization is usually achieved by slicing an image into equal-sized, square patches without considering their content.
This introduces redundant information across tokens, leading to computational waste: For instance, trivial background regions (e.g. sky and grass) are often expressed by a large number of tokens, dominating the bulk of compute in the model.
Nonetheless, it remains unclear how to design a more efficient tokenization that reduces input redundancy compared to such uniform patching. 
In fact, most successful token reduction methods in the literature, such as token pruning~\cite{xu_evovit_2021, rao_dynamicvit_2021, yin_avit_2022, meng_adavit_2022, kong_spvit_2022, pan_fast_2022} or token merging~\cite{ renggli_learning_2022,Ryoo2021TokenLearnerWC}, only act on intermediate layers of the transformer, while earlier layers still inefficiently operate with a large number of redundant tokens.

In this work, we propose a novel, orthogonal approach: We predict the \textit{tokenization scale} for each image region as a pre-processing step before the transformer. 
Intuitively, uninformative image regions such as background can be processed at a coarser scale than the foreground, without loss of information, leading to a smaller total number of tokens.
To capture this behavior, we introduce a lightweight conditional gating MLP trained to select the optimal tokenization scale for every coarse local image region, as illustrated in \hyperref[fig:teaser]{Figure \ref{fig:teaser}}, leading to a dynamic number of tokens per image.
Because it operates at the input level, the gate is agnostic to the choice of transformer backbone.
Furthermore, mixed-scale tokenization is lossless, as every input region is covered by a token, making it well suited for dense prediction tasks in contrast to other methods such as pruning. 
Nevertheless, learning such a scale selection module raises several issues: \textbf{(i)} Current multi-scale ViT architectures are often trained with extra parameters for each scale or have cumbersome training pipelines with multiple  stages~\cite{chen_crossvit_2021a,Zhu2021MakeAL,chen_cfvit_2022}. Instead, we design a unified, single-stage model by maximizing parameter sharing across scales.
\textbf{(ii)} The gating module may learn a bad local minimum such as always outputting the same trivial static pattern. To combat this, we introduce a novel training loss that enables finer control over the learned gating distribution, enhancing the dynamic behavior of the mixed-scale tokenization. 
Finally, \textbf{(iii)} the cost of training grows with the total number of fine and coarse tokens. To reduce  training costs, we employ an adaptive trimming strategy at training time which relies on the underlying mapping between coarse and fine tokens. The main contributions of this work are as follows:

\begin{enumerate}[topsep=2pt]
    \setlength{\itemsep}{2pt}
    \setlength{\parskip}{0pt}
    \setlength{\parsep}{0pt}
    \item We design a dynamic scale selection gating mechanism that acts as a \textit{preprocessing} stage, agnostic to the choice of transformer backbone, and trained jointly with the transformer \textit{in a single stage} with  mixed-scale tokens as inputs. 
    We show in experiments that this dynamic tokenization process leads to improved computational costs by reducing the number of input tokens.
    \item We propose a generalization of batch-shaping~\cite{Bejnordi2020BatchshapingFL} to better handle \textit{multi-dimensional distributions} when training dynamic gates: The resulting loss provides better control over the learned scale distribution, and allows for easier and better initialization of the gates.   
    \item We reduce the training overhead incurred from handling a set of tokens for each scale by (i) defining the gate locally at the coarse token level only and (ii) employing an adaptive trimming strategy during training.
\end{enumerate}

\section{Proposed method}
\label{sec:method}
In this work, we enhance the standard Vision Transformer (\textbf{ViT}) formalism with mixed-\mixedwhat tokens that are dynamically selected for each input image.
In this section, we briefly introduce ViT, then describe how we handle tokens extracted at different \mixedwhats, with a focus on keeping the architecture parameter-efficient (\hyperref[sec:mixedresarch]{Section \ref{sec:mixedresarch}}) and reducing training overhead (\hyperref[sec:mixedresarch]{Section \ref{sec:mixedrestraining}}).
Finally, we present the generalized batch-shaping loss for training the mixed-\mixedwhat selection module (\hyperref[sec:gatelearning]{Section \ref{sec:gatelearning}}). 
%

\begin{figure*}[htb!]
\centering
\includegraphics[width=0.98\linewidth]{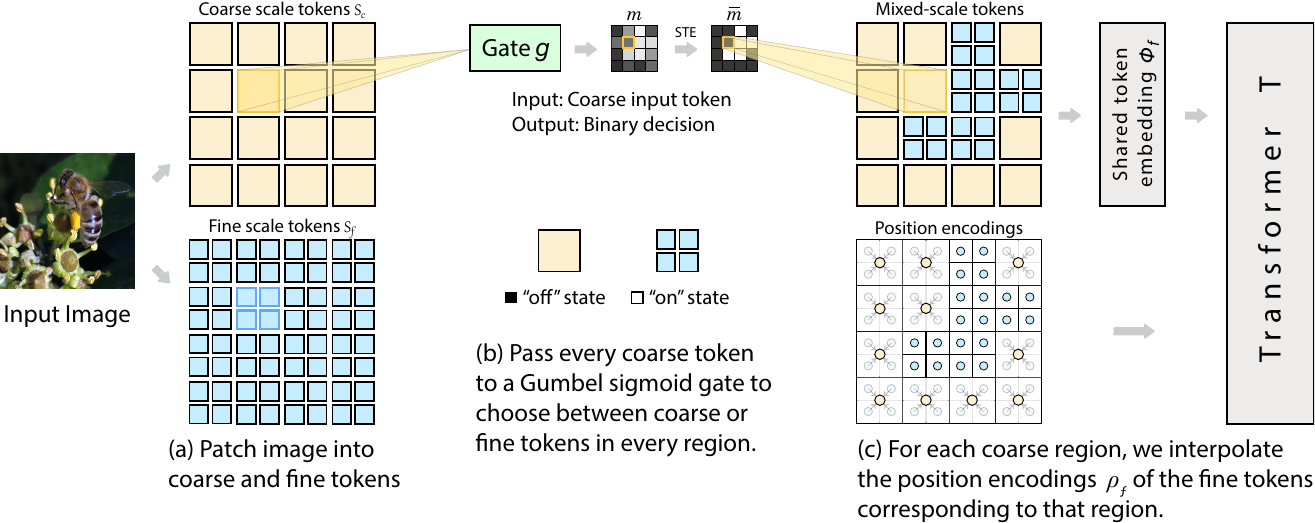}
\caption{\label{fig:modeloverview}
\label{fig:vit}Overview of the proposed dynamic mixed-scale tokenization scheme for ViT, MSViT.
\textbf{(a)} The input image is first patched into coarse image regions of size $S_{\idxcoarse}\times S_{\idxcoarse}$. \textbf{(b)} Each coarse region is processed by a small 4-layer MLP, the gate $g$, outputting a binary decision on whether the region should be processed at a coarse or fine scale.
\textbf{(c)} The resulting mask, $\overline{m}$, defines the set of mixed-scale tokens for the input image. The corresponding mixed-scale position encodings are obtained by linearly interpolating the fine scale position encodings to the coarse scale, when needed. 
Finally, the tokens are sent to a standard  transformer backbone $T$ which outputs the task-relevant prediction. 
}
\vspace{-0.45cm}
\end{figure*}

\subsection{Parameter-efficient mixed-\mixedwhat ViT}
\label{sec:mixedresarch}

Given an input image of size $W \times W$, a ViT first splits the image into  square \textit{patches} of equal size, $S$, resulting in a total of $N_S = \lfloor W / S \rfloor ^2$  tokens.
These tokens are flattened, and individually embedded to the target dimension $d$.
A position encoding is then added to each token, which is a vector capturing the initial 2D spatial location of the token.
Finally, the tokens are fed to a transformer, $T$, which is a sequence of Multi-Headed Self-Attention (\textbf{MHSA}) blocks, that compute global attention across the set of tokens, followed by FFNs, which process each token independently \cite{vaswani_attention_2017}.
Our work is agnostic to the choice of the transformer backbone $T$, thus, in the rest of the section, we only describe changes made to the patching, token embedding, and position encoding mechanisms to handle mixed-\mixedwhat tokens.  

\vspace{-0.21cm}
\paragraph{Dynamic mixed-\mixedwhat ViT.} 
An overview of the proposed mixed-scale vision transformer model (MSViT) is presented in \hyperref[fig:modeloverview]{Figure \ref{fig:modeloverview}}.
In the scope of this paper, we consider the case of two  \mixedwhats  ($S_{\idxfine} < S_{\idxcoarse}$).
We refer to $S_{\idxfine}$ (resp. $S_{\idxcoarse}$) as the \textit{fine} (resp. \textit{coarse}) \mixedwhat.
First, we extract square patches at both \mixedwhats, for a total of $N = N_{S_{\idxfine}} + N_{S_{\idxcoarse}}$ tokens.
We then introduce a discrete \textit{gating} mechanism, $g$, which selects active tokens across both \mixedwhats, for a given input image: These tokens are further sent to the  transformer, while inactive ones are discarded at this stage. %

In practice, we define the learned gate as a local operation, at the level of coarse tokens: The gate parses each coarse image region individually and outputs a binary decision on whether the region should be tokenized at either the coarse or  fine scale.
We consider the case where the fine-scale $S_{\idxfine}$ evenly divides the coarse scale  $S_{\idxcoarse}$. This way, for all $i$, the $i$-th fine token can be mapped to the unique coarse token $C(i) = j$ it belongs to. 
Using this mapping, we recover the complete binary mixed-scale mask at the fine token level, $\overline{m}$, using the coarse-level gate outputs:  
\begin{align}
    \forall j \in [1, N_{S_{\idxcoarse}}],\  &m_j = \text{GumbelSigmoid}(g(x_j)) \in [0, 1]\\
    &\overline{m}_j = \text{STE}(m_j) \in \{0, 1\}\\
    \forall i \in [N_{S_{\idxcoarse}}& + 1, N_{S_{\idxcoarse}} + N_{S_{\idxfine}} ],\  \overline{m}_i = 1 -  \overline{m}_{C(i)}
\end{align}
Here, we distinguish between the soft outputs of the gate, $m \in [0, 1]$, used to constrain the gate during training, and the discretized outputs $\overline{m} \in \{0 , 1\}$ used during the forward pass. 
In order, to estimate gradients for the discrete gate operation, we use the Gumbel-Sigmoid relaxation of binary variables during training~\cite{Maddison2017TheCD} with the straight-through gradient estimator (\textbf{STE}) \cite{hinton_neural_2012, bengio_estimating_2013}. 

While this design choices for the gate may limit representational power, as $g$ only sees local regions of the image as inputs, we find that it works well in practice and yields a very lightweight gating strategy. Moreover, as in the original ViT tokenization, token overlap is prevented by design, as every image region can only be captured by a unique scale.

\vspace{-0.21cm}
\paragraph{Sharing parameters across \mixedwhats.}
Previous mixed-scale ViTs usually introduce extra parameters to handle each \mixedwhat~\cite{chen_crossvit_2021a,wang_not_2021} or train a shared backbone stage by stage for each scale separately~\cite{chen_cfvit_2022,Zhu2021MakeAL}. 
Instead, our intention is \textbf{(i)} to fully share the token embedding, position encodings, and the transformer backbone across \mixedwhats, and \textbf{(ii)} to directly train the model in one stage with batches of mixed-scale tokens, rather than treating each scale individually. 
This allows us to avoid extra parameter costs and makes our method architecture agnostic. In addition, due to the dynamic nature of the gating mechanism, defining separate branches per \mixedwhat instead of sharing may lead to common issues of training conditional models such as imbalanced routing and data starvation~\cite{moe1,moe2,moe3}. 

To implement sharing across scales, we draw inspiration from ViT~\cite{dosovitskiy_image_2021,Beyer2022FlexiViTOM}: At inference, the authors scale a trained model to a different input image size by linearly interpolating its position encodings to match the size of the new grid. 
We extend this idea to our  setting by defining the learned embedding $\phi_{\idxfine}$ and position encoding parameters $\rho_{\idxfine}$ relative to the \textit{fine scale} only (\hyperref[fig:modeloverview]{Figure \ref{fig:modeloverview} (c)}). We then deterministically infer their equivalent for the coarse scale as: 
\begin{align}
    \phi_{\idxfine}: x \in \mathbb{R}^{S_{\idxfine} \times S_{\idxfine} \times 3} \mapsto \mathbb{R}^d,\ \ \rho_{\idxfine} \in \mathbb{R}^{N_{S_{\idxfine}} \times d}\\
    \phi_{\idxcoarse} = \phi_{\idxfine} \circ\ \text{resize}(S_{\idxcoarse} \rightarrow S_{\idxfine}),\ \rho_{\idxcoarse} = \text{\small interpolate}(\rho_{\idxfine})
\end{align}
In \hyperref[app:resizetrick]{Appendix \ref{app:resizetrick}}, we show that this simple linear interpolation scheme works very well in practice, but may suffer when rescaling to a very low token resolution: For instance, directly training with the coarse patch size 32 on inputs of size 224px yields higher accuracy than the model with fine patch size 16, rescaled for 112px inputs to reach the same number of 49 tokens.
Nevertheless, this is not an issue for the image and patch sizes we consider in our experiments.

\begin{figure*}[htb!]
\centering
\includegraphics[width=0.96\linewidth]{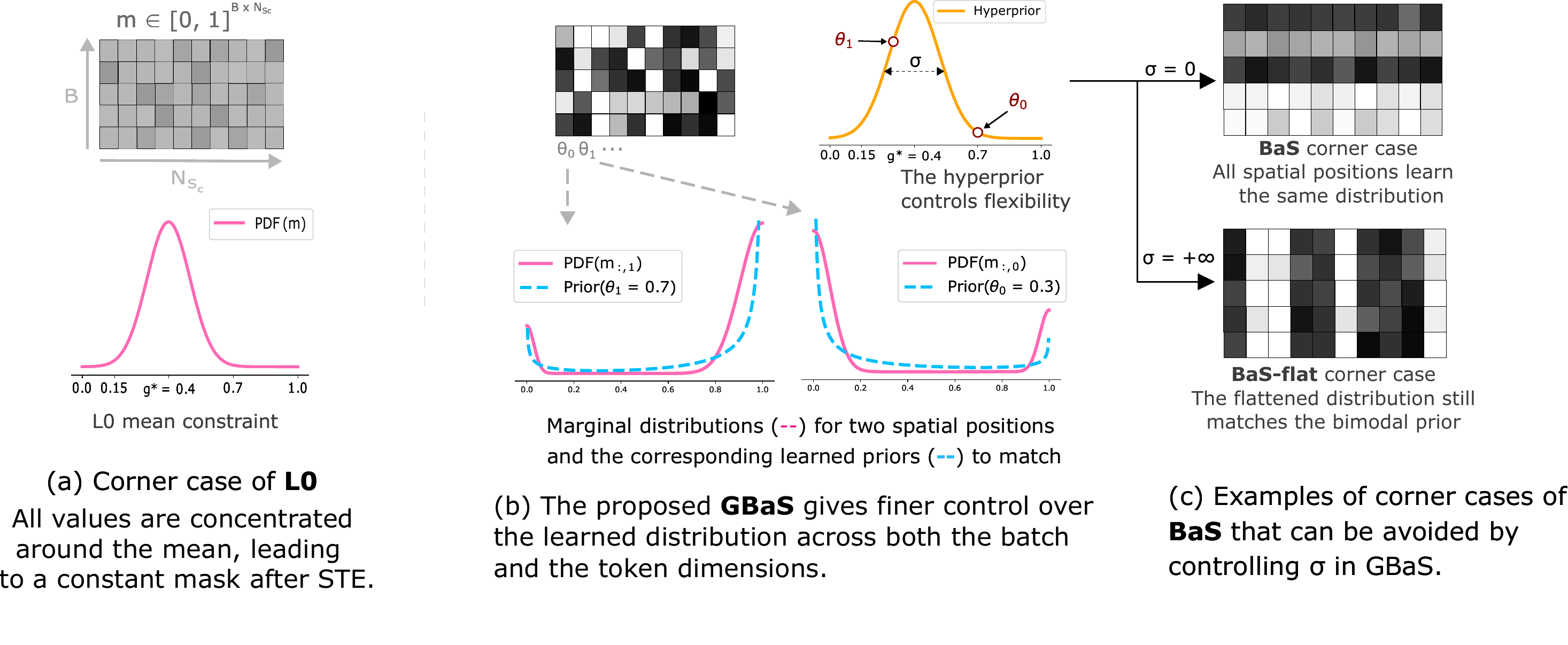}
\vspace{-0.55cm}
\caption{\label{fig:lossesexplained} Our proposed generalized batch-shaping (\gbsloss) allows for fine control over the learned distribution via a hyperprior \textbf{(b)}: GBaS allows for learning different distributions for each token position in contrast to \bsloss{} \textbf{(c, top)}; In addition, \gbsloss{} explicitly controls this flexibility through the variance hyperparameter $\sigma$, hence avoiding corner cases of \bsloss-flat \textbf{(c, bottom)} or L0 \textbf{(a)} }
\vspace{-0.4cm}
\end{figure*}

\subsection{Learning the mixed-\mixedwhat gating}
\label{sec:gatelearning}
We jointly train the transformer and the gate by balancing the model performance with computational efficiency, forcing the model to only select a few tokens at fine scale:
\begin{align}
    \mathcal{L} (x_{1\dots N}, m_{1\dots N}, y) = \mathcal{L}_{task}(x, y; m) + \lambda \mathcal{L}_{gate}(m)
\end{align}
where $\mathcal L_{task}$ is the task loss (e.g., cross-entropy) applied on the masked transformer outputs, $\mathcal{L}_{gate}$ is a sparsity constraint on the gate output $m$ (before STE), which directly controls the model's computational cost, and $\lambda$ is a hyperparameter balancing both losses. In the next paragraphs, we will motivate and define a novel gate loss to use for $\mathcal{L}_{gate}$.

\vspace{-0.31cm}
\paragraph{Common gate sparsity losses.} The $L_0$ loss is often used as sparsity loss in the conditional computing literature\cite{veit2018convolutional}.
Given the 2-dimensional active token mask for the current batch, $m \in [0, 1]^{B \times N}$, we define:
\begin{align}
    \mathcal{L}^{L_0}_{gate}(m) = \frac{1}{B} \sum_{b=1}^B \min \left(0, \frac{1}{N_{S_{\idxcoarse}}} \sum_{i=1}^{N_{S_{\idxcoarse}}} m_{b, i} - g^{\ast}\right)
\end{align}
where the hyperparameter $g^\ast \in [0, 1]$ is the target rate for gate sparsity.
However, $L_0$ only penalizes the \textit{mean} of the distribution, and can not prevent the model from learning static patterns, such as assigning the same probability to all tokens independent of input, as illustrated in \hyperref[fig:lossesexplained]{Figure \ref{fig:lossesexplained} (a)}. 

To enhance the desired conditional behavior, the recently proposed batch-shaping loss~\cite{Bejnordi2020BatchshapingFL} (\bsloss)  constrains the distribution of the gate outputs, across the batch, to match a certain prior $p$.
In our setting, this means enforcing the \textit{same prior} across each spatial position. This lacks the necessary flexibility for our use-case, as the gate could not learn for instance that edges of the image are less likely to contain fine-scale details.
As a more flexible alternative, we apply \bsloss{} directly on the flattened distribution of the gate outputs: 
\begin{align}
    \mathcal{L}^{\bsloss}_{gate}(m) = \left[ \text{CDF}(\left\{m_{b, i},\  \forall b, i\right\}) - \text{CDF}(p(g^\ast)) \right]^2
\end{align}
where CDF is the cumulative distribution function, and $p$ is a prior with mean $g^{\ast}$. 
Unfortunately, this variant is now too flexible, e.g. it cannot prevent spatial positions from being constantly on or off regardless of the input patch. 
Corner cases for both variants of \bsloss{} are illustrated in   \hyperref[fig:lossesexplained]{Figure  \ref{fig:lossesexplained} (c)}.

\vspace{-0.3cm}
\paragraph{Generalized batch-shaping loss.}
To address these shortcomings, we introduce the \textit{generalized batch-shaping loss} (\textbf{\gbsloss})  for finer control over the learned mask distribution, across both the batch and token dimensions.
Like \bsloss{}, \gbsloss{}  constrains the marginal distribution at each token spatial position, $m_{:, i}\ \forall i \in [1, N_{S_{\idxcoarse}}]$, but with a dedicated independent prior instead of a shared one.
Manually setting the prior for each position would be tedious; Instead, we let the model learn each of these independent prior's parameters, while controlling their distribution using a \textit{hyperprior} $\mathcal P$ with mean equal to the target sparsity $g^{\ast}$ (\hyperref[fig:lossesexplained]{Figure \ref{fig:lossesexplained} (b)}): 
\begin{align}
    \mathcal{L}^{\gbsloss}_{gate}(m) = &\sum_{i=1}^{N_{S}} \left[ \text{CDF}(\left\{m_{b, i},\  \forall b \right\}) - \text{CDF}(p(\theta_i)) \right]^2 \nonumber\\
    + &\left[ \text{CDF}(\left\{\theta_i,\  \forall i\right\}) - \text{CDF}(\mathcal{P}(g^\ast; \sigma)) \right]^2
\end{align}
where $\theta$ are learned parameters defining each prior, and $\sigma$ is a variance hyperparameter controlling the spread of the learned priors.
When $\sigma = 0$, all priors are identical; hence we recover the original \bsloss{}; When $\sigma \rightarrow + \infty$, there is little constraint on the learned $\theta$ and we may encounter the same corner cases as for \bsloss{} applied to the flattened distribution. 

In summary, \gbsloss{} enables fine-grained control over the learned distribution through the hyperprior.
Another benefit of \gbsloss{} is that we can easily inject prior knowledge about which spatial positions are more likely to be kept at fine/coarse \mixedwhat by initializing the $\theta$ parameters accordingly. 
In contrast, achieving a similar initialization with \bsloss{} would require pretraining the gate to match the desired prior. For instance, in most of our experiments with \gbsloss, we initialize the learned prior parameters $\theta$ with the inverse normalized distances of each spatial position to the center. 
We further compare \bsloss{} and \gbsloss{} in ablation experiments in \hyperref[sec:ablation]{Section \ref{sec:ablation}} and \hyperref[app:gbas]{Appendix \ref{app:gbas}}.
We use the Relaxed Bernoulli~\cite{Maddison2017TheCD} distribution for the prior $p$, as we found it easier to parametrize than the Beta distribution used in \bsloss. We use a Gaussian for the hyperprior $\mathcal{P}$ with mean $g^{\ast}$ and variance given by the hyperparameter $\sigma$.
Our implementation for the Batch-Shaping and generalized Batch-Shaping losses is available on \href{https://github.com/Qualcomm-AI-research/batchshaping}{github}\footnote{\url{https://github.com/Qualcomm-AI-research/batchshaping}}.

\subsection{Reducing the training overhead}
\label{sec:mixedrestraining}
When executing the model with batches of data, inactive tokens ($\overline{m}_i = 0$) cannot be pruned statically, as the masking pattern output by the gate $g$ varies across the batch.
Instead, we explicitly mask the inactive tokens in the attention layers and the output of the transformer backbone; the FFN layers are applied individually to every token and hence are not affected. 
Given the set of tokens across all \mixedwhats, $x \in \mathbb{R}^{N \times d}$ and the current binary mask output by the gate, $\overline{m} \in \{0, 1\}^{N}$, we must apply masking in every attention block, such that the inactive tokens are ignored when updating the representations of active ones: 

\begin{align}
\label{eq:maskedatt}
    \forall i, j \in [1, N],\ A^{\text{mask}}(x_i, x_j) = \frac{\overline{m}_j\ e^{Q_i K_j^T}}{\sum_{p=1}^N \overline{m}_p\ e^{Q_i K_p^T} }
\end{align}
where $A^\text{mask}(x_i,x_j)$ is the normalized attention score from token $i$ to $j$ and $Q$ and $K$ denote the query and key embeddings of the tokens.
Unfortunately, with this naive masking approach the increased total number of tokens, $N = N_{S_{\idxfine}} + N_{S_{\idxcoarse}}$, leads to higher  training costs.

To address this issue, we employ an \textit{adaptive trimming} (\textbf{AT}) strategy at training time: 
For each image in the batch, we first reorder the tokens in descending order of the corresponding gate outputs $m$, omitting the class token or any task-specific token. This reordering step takes advantage of the fact that the transformer is not affected by the order of the tokens.
We then trim the token dimension to only keep $k$ tokens for each image, where $k$ is the maximum number of active tokens in any image in the current batch. 
As a result, the number of tokens (and hence the computational cost) is lower bounded by $N_{S_{\idxfine}}$, i.e., the number of fine \mixedwhat tokens. 
This strategy does impact the gradients received by the gate, effectively making the gating module less robust to tokens flipping from the coarse to the fine scale during training (see \hyperref[app:trimvsfullmask]{Appendix \ref{app:trimvsfullmask}}).
Nevertheless, as we show in \hyperref[app:trimming]{Appendix \ref{app:trimming}}, this only leads to a small drop in accuracy in practice but a clear reduction in training time ($\sim$1.16-1.35 times per-epoch speedup, depending on the target sparsity). For this reason, we always use AT in our training pipeline.

\section{Related work}
\label{sec:related}
\paragraph{Self-Attention for computer vision.} 
Starting from \gls{vit} \cite{dosovitskiy_image_2021, cordonnier_relationship_2020,carion_endtoend_2020a,ramachandran2019stand}, \gls{mhsa} has been successfully applied in many vision tasks such as image classification \cite{dosovitskiy_image_2021, touvron2021training}, object detection \cite{carion_endtoend_2020, zhu2021deformable} or semantic segmentation \cite{duke_sstvos_2021, zheng_rethinking_2021, liu2021Swin}. 
While \glspl{vit} are often able to match CNN-based models' performance with fewer computational resources \cite{dosovitskiy_image_2021}, the number of input tokens remains an important bottleneck to achieve efficient transformers. 
Several works~\cite{Tay2020Efficient} have focused on reducing the cost of the attention operation, which scales quadratically with the number of tokens, by using low-rank approximations~\cite{choromanski_rethinking_2021, wang_linformer_2020, peng_random_2021} or exploiting redundant or sparse structures~ \cite{kitaev_reformer_2020, vyas_fast_2020, fedus_switch_2022, lepikhin_gshard_2020, liu_swin_2022, vyas_fast_2020}.
However, unlike for NLP, the cost incurred by the \gls{ffn} in \glspl{vit} is often significant due in part to the generally smaller number of tokens. 
Hence, instead of focusing only on attention layers, a number of techniques have been proposed to reduce the total number of tokens.

\vspace{-0.35cm}
\paragraph{Token pruning and merging.} 
Token pruning~\cite{xu_evovit_2021, rao_dynamicvit_2021, yin_avit_2022, meng_adavit_2022, kong_spvit_2022, pan_fast_2022, liang_not_2022} and merging~\cite{renggli_learning_2022,Ryoo2021TokenLearnerWC,extra1} are some of the most successful token reduction approaches in the literature.
These methods usually prune away a fixed number of tokens in intermediate layers of the transformer based on their class attention score~\cite{liang_not_2022, yin_avit_2022} or on the previous layer's features~\cite{rao_dynamicvit_2021}, or merge tokens into a fixed smaller number of tokens using a cross-attention layer or projection~\cite{renggli_learning_2022,Ryoo2021TokenLearnerWC}.

Orthogonal to these methods, our mixed-scale selection scheme outputs a dynamic number of tokens, tailored to the input image content. 
It is also designed as a preprocessing module acting on the token set before the first Transformer layer, and hence can be combined with methods such as token pruning or early-exiting which act on the intermediate transformer layers. 
Finally, in contrast to pruning, mixed-scale models are lossless in the sense that every input image region is covered by a token. This is important for dense tasks such as segmentation where the final spatial predictions are usually directly reconstructed from the tokens.

\vspace{-0.35cm}
\paragraph{Mixed-scale ViTs.} 
Mixing features from different scales has shown positive results for convolutional networks~\cite{Lin2017FeaturePN,Lin2020FocalLF}. 
Following this insight, recent works have started to investigate ways to incorporate mixed-scale information in \glspl{vit} as well: 
Quadtree Attention \cite{Tang2022QuadTreeAF} uses hierarchical structures to improve the efficiency of \gls{mhsa}.  
CrossViT~\cite{chen_crossvit_2021a} defines separate  branches for each scale, which occasionally communicate through cross-attention. 
CF-ViT~\cite{chen_cfvit_2022} and DVT~\cite{wang_not_2021} combine early-exiting with a two-stage cascade of transformers, one for each scale. 
Finally ReViT~\cite{Zhu2021MakeAL} learns a global input patch scale for each image with an EfficientNet backbone trained with precomputed proxy labels. 
The majority of these works treat each scale separately, either by  incorporating extra parameters (entire branch~\cite{chen_crossvit_2021a,wang_not_2021} or layernorm parameters~\cite{Zhu2021MakeAL}) or by training for each scale in separate stages~\cite{chen_cfvit_2022,Zhu2021MakeAL}. 
In contrast, we design a simpler single-stage model which directly handles having mixed-scale tokens in one batch, for both training and inference. 
Closest to our work is \cite{extra2}, which  leverages saliency maps from a pretrained model to design a quadtree structure on token scales and enable mixed-scale token selection.

\section{Experiments}
\label{sec:exp}
\subsection{ImageNet classification}

We first benchmark the proposed mixed-scale tokenization on  ImageNet~\cite{Russakovsky2015ImageNetLS}: We use publicly available SotA ViT backbones pretrained on ImageNet-21k~\cite{Steiner2021HowTT,dosovitskiy_image_2021,vitrepo}, and DeiT backbones pretrained on ImageNet~\cite{touvron2021training,deitrepo}.
We implement the gate as a lightweight 4-layer MLP with a scalar output in [0, 1], applied to every coarse token individually. After the first layer, a learned position encoding, specific to the gate, is also added to the token representations. Finally, the bias of the last layer is initialized  such that the gate outputs 1: i.e., all patches are extracted at the fine scale at the beginning of training.
We set all other hyperparameters to that of the original ViT (resp. DeiT)  pipeline and finetune all models for 20 epochs with a batch size of 512 on a single device (see additional training details in \hyperref[app:hyperparameter]{Appendix \ref{app:hyperparameter}}).

\begin{table}[!htb]
    \centering
    
      \resizebox{0.48\textwidth}{!}{  
    \begin{tabular}{|c||c|c||c|c|}
    \hline
         DeiT-Small & Avg \# & GMACs  & \multicolumn{2}{c|}{accuracy}\\
         \cline{4-5}
          backbone & tokens & (avg) & top-1 & top-5\\
         \hline
         \hline
        \rowcolor{mygray} DeiT-S/16 in=160 & 100 & \textbf{2.27} & 75.86 & 92.84\\
        \hline
        \textbf{MSDeiT}-S/16,32 in=224 & \textbf{97} & \textbf{2.27} & \textbf{76.99} & \textbf{93.38}\\
         \hline
         \hline
        \rowcolor{mygray} DeiT-S/16 in=192 & 144 & \textbf{3.32} & 77.79 & 93.96\\
        \hline
        \textbf{MSDeiT}-S/16,32 in=224 & \textbf{142} & \textbf{3.32} & \textbf{78.76} & \textbf{94.32}\\
         \hline
         \hline
        \rowcolor{mygray} DeiT-S/16 in=224 & 196 & 4.60 & \textbf{79.85} & \textbf{94.57}\\
        \hline
        \textbf{MSDeiT}-S/16,32 in=224 & \textbf{173} & \textbf{4.08} & 79.38 & 94.38\\
        \hline
    \end{tabular}
    }
    \vspace{0.2cm}
    
      \resizebox{0.48\textwidth}{!}{  
    \begin{tabular}{|c||c|c||c|c|}
    \hline
         ViT-Tiny & Avg \# & GMACs  & \multicolumn{2}{c|}{accuracy}\\
         \cline{4-5}
          backbone & tokens & (avg) & top-1 & top-5\\
        \hline
        \hline
        \rowcolor{mygray} ViT-Ti/16 in=160 & 100 & \textbf{0.60} & 71.63 & 90.68\\
            \hline
        \textbf{MSViT}-Ti/16,32 in=224 & \textbf{95} & \textbf{0.60} & \textbf{72.57} & \textbf{91.32}\\
        \hline
        \hline
        \rowcolor{mygray} ViT-Ti/16 in=192 & 144 & 0.89 & 74.24 & 92.22\\
            \hline
       \textbf{ MSViT}-Ti/16,32 in=224 & \textbf{138} & \textbf{0.88} & \textbf{74.93} & \textbf{92.54}\\
        \hline
        \hline
        \rowcolor{mygray} ViT-Ti/16 in=224 & 196 & 1.25 & \textbf{76.00} & \textbf{93.26}\\
            \hline
        \textbf{MSViT}-Ti/16,32 in=224 & \textbf{154} & \textbf{0.98} & 75.51 & 92.98\\
         \hline
    \end{tabular}
    }
    \vspace{0.2cm}
    
      \resizebox{0.48\textwidth}{!}{  
    \begin{tabular}{|c||c|c||c|c|}
    \hline
         ViT-Small & Avg \# & GMACs  & \multicolumn{2}{c|}{accuracy}\\
         \cline{4-5}
          backbone & tokens & (avg) & top-1 & top-5\\
        \hline
        \hline
            \rowcolor{mygray} ViT-S/16 in=128 & \textbf{64} & \textbf{1.44} & 75.48 & 93.08\\
            \hline
            \textbf{MSViT}-S/16,32 in=224 & 75 & 1.76 & \textbf{77.16} & \textbf{94.14}\\
        \hline
        \hline
            \rowcolor{mygray} ViT-S/16 in=160 & 100 & \textbf{2.27} & 78.88 & 94.95\\
            \hline
            \textbf{MSViT}-S/16,32 in=224 & \textbf{98 }& 2.30 & \textbf{79.51} & \textbf{95.33}\\
        \hline
        \hline
        \rowcolor{mygray} ViT-S/16 in=192 & 144 & 3.32 & 80.75 & 95.86\\
            \hline
        \textbf{MSViT}-S/16,32 in=224 & \textbf{138} & \textbf{3.23} & \textbf{81.47} & \textbf{96.14}\\
        \hline
        \hline
        \rowcolor{mygray} ViT-S/16 in=224 & 196 & 4.60 & \textbf{82.02} & \textbf{96.4}5\\
            \hline
        \textbf{MSViT}-S/16,32 in=224 & \textbf{187} & \textbf{4.43} & \textbf{82.02 }& 96.44\\
         \hline
    \end{tabular}
    }
    \vspace{0.12cm}
    
    \caption{Comparison of our dynamic mixed-scale model with the corresponding backbone baseline evaluated at different input image sizes.
    For ease of reading, the results are sorted by MACs, and grouped by backbones. Inside each table, we group results by comparable MAC counts or  accuracy. 
    We refer to models as ``\texttt{arch/S in=X}'', where \texttt{arch} is the backbone architecture, \texttt{X} is the input image size, and \texttt{S} is the patch scale(s). 
    The prefix \texttt{MS} (Multi-Scale) refers to our mixed-scale models:
    We sweep over values of the gate target $g^{\ast} \in \{0.5, 0.25, 0.1 \}$ and loss weight $\lambda \in \{ 1, 4, 20\}$ to obtain dynamic models with various MACs counts and report their GMACs and number of tokens  averaged over the evaluation set (For reference, the additional computational cost induced by the gate for ViT-S is 0.017 GMACs).
    Additional results for all hyperparameters and different input image sizes, and including latency measurements, can be found in \hyperref[app:fullclass]{Appendix \ref{app:fullclass}}.
    \vspace{-0.5cm}
    }
    \label{tab:main_results}
\end{table}


In \hyperref[tab:main_results]{Table \ref{tab:main_results}}, we evaluate the proposed mixed-scale MSViT across different backbone architectures (ViT-S and ViT-Tiny), pretraining strategies (ViT and DeiT), and input image sizes. We report top-1 accuracy results as well as MACs counts calculated via deepseed~\cite{deepspeed}.

From the quantitative results, we observe that the mixed-scale models consistently reach higher accuracy at equivalent MACs across different compute budgets and input sizes. 
We also display qualitative examples of the mixed-\mixedwhat selection patterns learned by the gate in \hyperref[fig:teaser]{Figure \ref{fig:teaser}} and \hyperref[app:extra_results]{Appendix \ref{app:extra_results}}: 
Despite having a limited field of view, the learned gate picks up on meaningful  local features such as background/foreground distinction to select tokens' scales.
Furthermore, we observe that the learned mixed-scale pattern is very similar across experimental settings: Two gates with the same number of active tokens, trained for MSViT-S/16 and MSViT-L/16 respectively, select the same scale for \textbf{78.4\% }of the tokens on the ImageNet validation set. 
Similarly, the gates of a MSViT-S/16 model trained with 224px and 192px inputs respectively, agree for \textbf{87.1\%} of the tokens.
Motivated by this observation, we investigate in the next section whether the learned mixed-scale gate can be directly transferred as an off-the-shelf lightweight preprocessing module to other vision transformer-based models.

\subsection{Transferring mixed-scale tokenization across tasks and backbones}
\label{sec:transfer}

\subsubsection{Mixed-scale tokens for segmentation}
\label{sec:seg}
To verify whether MSViT also benefits dense prediction tasks, we augment the standard Segmenter training pipeline~\cite{segmenterrepo,strudel2021} on ADE20k~\cite{Zhou2017ScenePT} with one of our gates, pretrained on ImageNet and frozen. 
The change is easy to implement: we replace the standard patch embedding of the ViT encoder with our own mixed-scale tokenization (\hyperref[sec:mixedresarch]{Section \ref{sec:mixedresarch}}) and keep it frozen during training. We then propagate the mixed-scale mask into further layers using masked attention (Equation \eqref{eq:maskedatt}), and finally reshape the stream of mixed-scale tokens to the original 2D image shape before feeding it to the decoder (see \hyperref[app:reshapes]{Appendix \ref{app:reshapes}} for  details).

\begin{figure}
    \begin{center}
      \resizebox{0.5\textwidth}{!}{  
    \begin{tabular}{|c|c||c|c|c||c|}
        \hline
         Backbone & $g^{\ast}$ & \# tokens & MACs & time & mIoU \\
         &  & avg & x 1e10 & ms & \footnotesize single-scale \\
         \hline
         \rowcolor{mygray} Seg-T/16 {\footnotesize (512px)}  & - & 1024 & 1.04 & 113.68& 38.1 \\
         \hline
         \multirow{3}{*}{\textbf{MS}Seg-T/16} & 
         0.5 & 655 & 0.56 & 86.12& 37.9\\
          \cline{2-5}
          & 
         0.25 & 565 & 0.46 & 75.96& 37.3 \\
          \cline{2-5}
          & 
         0.1 & 525 & 0.42 & 69.13 & 36.8 \\
          \hline
          \hline
         \rowcolor{mygray} Seg-S/16 {\footnotesize (512px)}  & - & 1024 & 3.17 &  252.09 & 45.3 \\
         \hline
         \multirow{3}{*}{\textbf{MS}Seg-S/16} & 
         0.5 & 684 & 1.92 & 184.81 & 44.9\\
          \cline{2-5}
          & 0.25 & 586 & 1.59 & 153.12 & 44.1\\
          \cline{2-5}
          & 
         0.1 & 552 & 1.48  & 144.02 & 43.3 \\
         \hline
    \end{tabular}
    }
    \end{center}
    \vspace{-0.2cm}
    {\small \textbf{(a)} Single-scale segmentation results of our mixed-scale model with ViT-S and ViT-Ti backbones finetuned on ADE20K~\cite{Zhou2017ScenePT}. We measure the computational cost of the  encoder, as the decoder cost is the same for both MSViT and ViT backbones; We also report the average runtime per image measured on a Geforce 2080 Ti GPU}\\[0.2cm]
    %
    \newcommand\idx{592}
    \setlength{\fboxsep}{0pt}%
    \setlength{\fboxrule}{0.4pt}%
    \begin{minipage}[b]{0.155\textwidth}
    \centering
     \footnotesize Mixed-scale mask 
     \vspace{0.023cm}
    \fbox{\includegraphics[width=\textwidth]{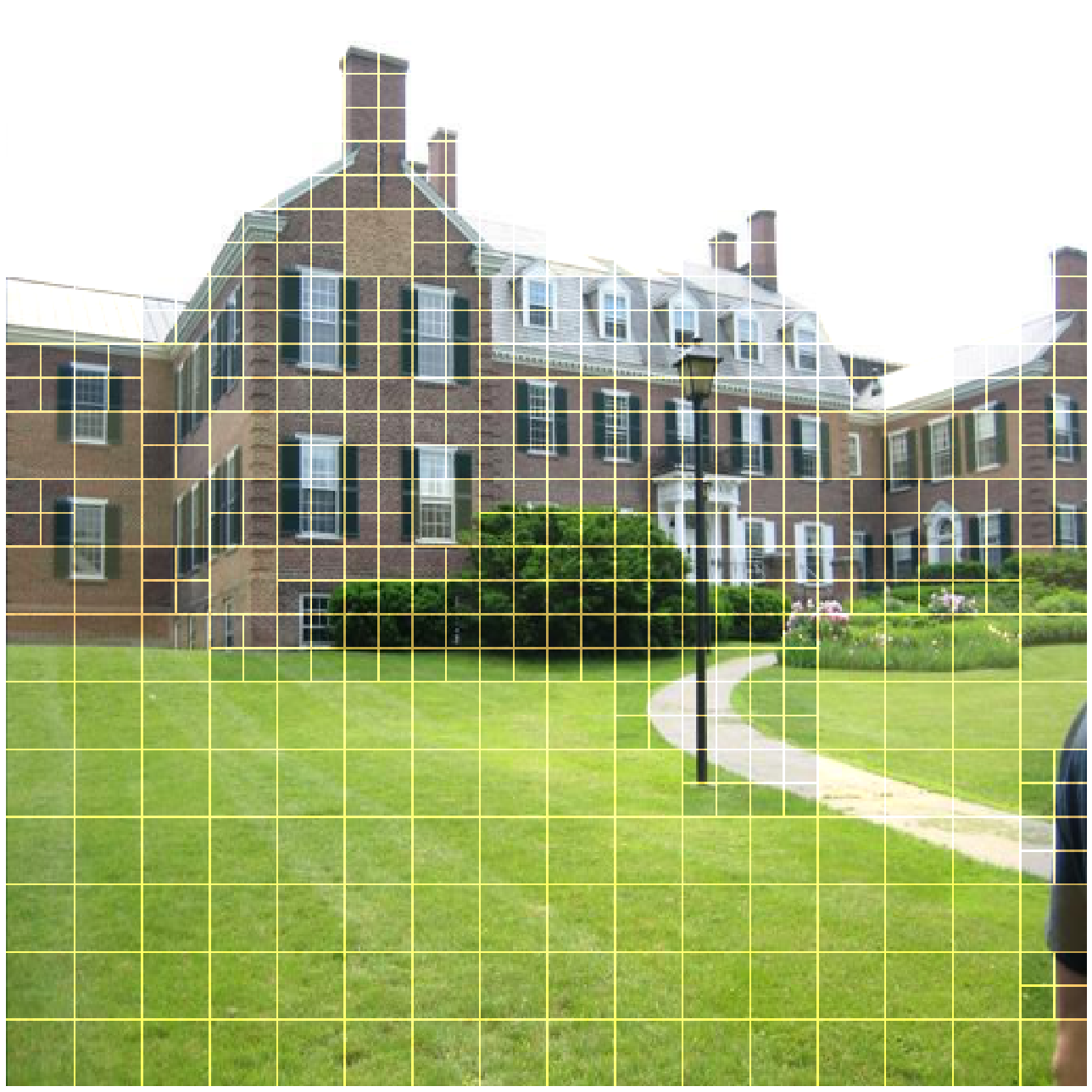}}
    \end{minipage}
    \begin{minipage}[b]{0.155\textwidth}
    \centering
    \small \footnotesize \textbf{MS}Seg-S/16 \tiny ($g^{\ast}$ = 0.25)
    \fbox{\includegraphics[width=\textwidth, height=\textwidth]{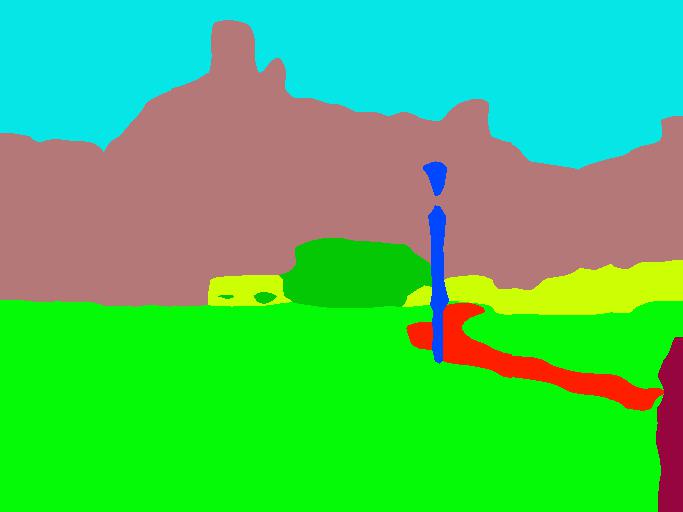}}
    \end{minipage}
    \begin{minipage}[b]{0.155\textwidth}
    \centering
    \footnotesize Seg-S/16
    \fbox{\includegraphics[width=\textwidth, height=\textwidth]{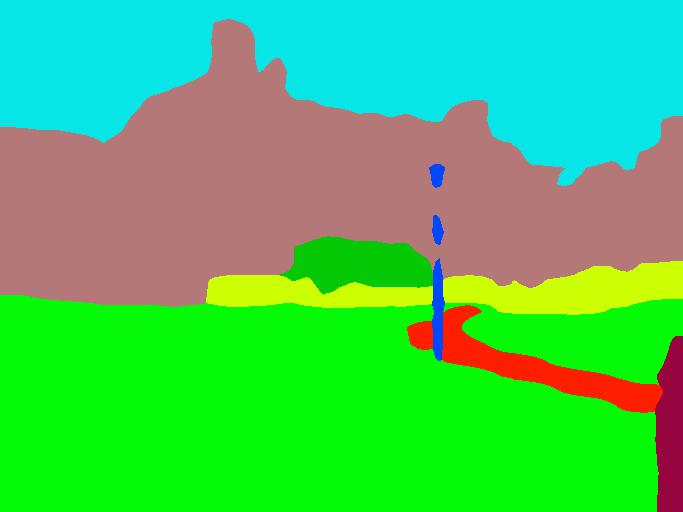}}
    \end{minipage}
    {\small \textbf{(b)} Example of a mixed-scale mask and segmentation output, as well as the baseline backbone's output (\textit{best seen zoomed}). We report additional qualitative results in \hyperref[app:adek]{Appendix \ref{app:adek}}.
 }
%
    \vspace{-0.1cm}
    \begin{center}
        \includegraphics[width=0.46\textwidth]{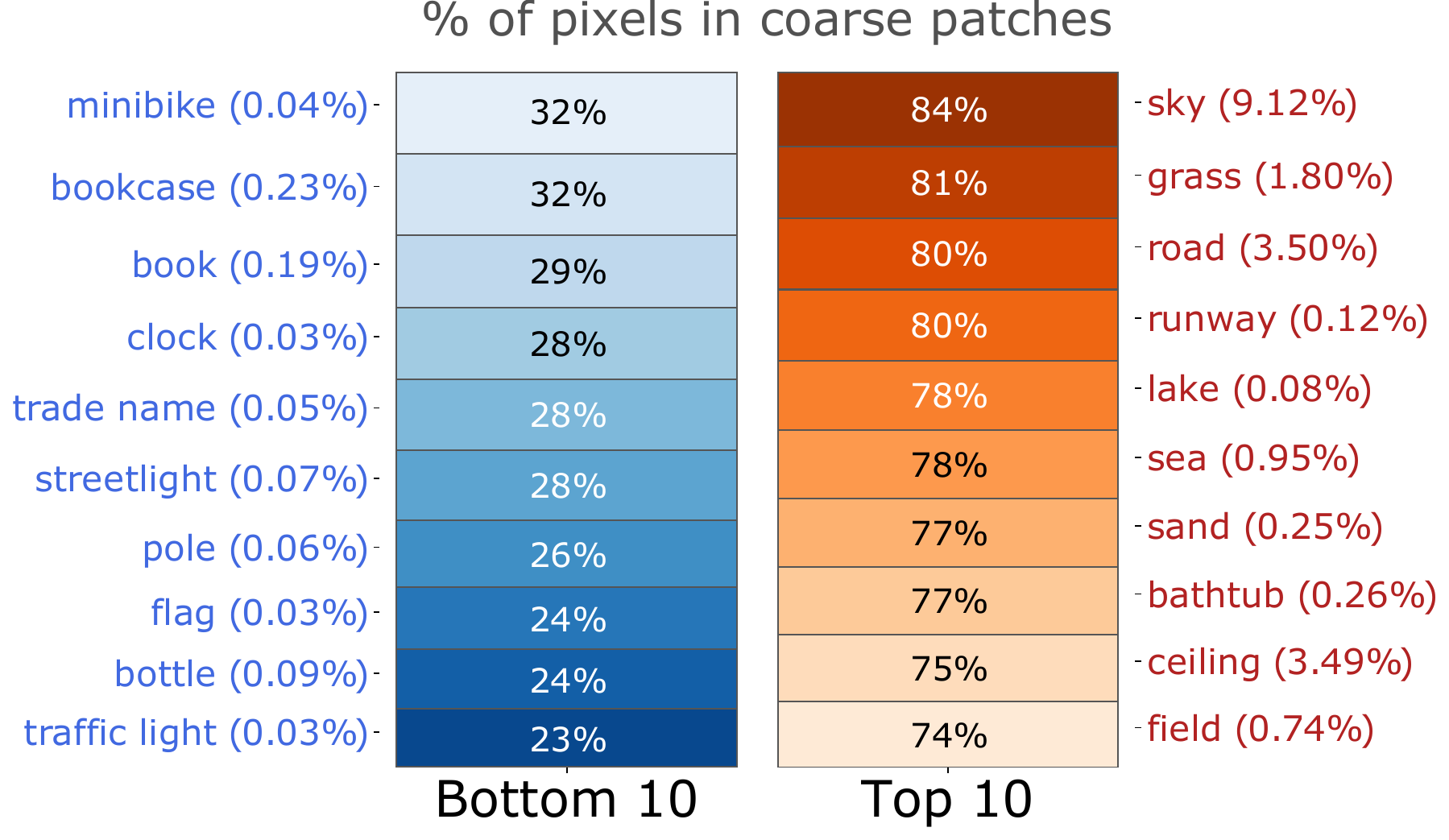}
    \end{center}
    \vspace{-0.5cm}
    {\small \textbf{(c)} ADE20K classes with the highest and lowest percentage of pixels falling in coarse patches. We also write the pixel frequency of each class in the whole dataset next to its label.}\\
    \vspace{-0.2cm}
    \caption{We train Segmenter~\cite{segmenterrepo,strudel2021} on the ADE20k~\cite{Zhou2017ScenePT} dataset, after adding a (frozen) mixed-scale gate trained on ImageNet. We report quantitative results in Table \textbf{(a)}, a qualitative example in \textbf{(b)}, and a break-down of classes most often in coarse regions in \textbf{(c)}}
    \vspace{-0.5cm}
    \label{fig:segmentation}
\end{figure}

We report the results (mIOU, average MAC count, and average latency) in \hyperref[fig:segmentation]{Figure \ref{fig:segmentation} (a, b)}. 
Similar to the classification task, we observe improved accuracy-efficiency trade-offs across different backbone sizes and gate sparsities: For instance with a ViT-S backbone, we can save roughly 40\% MACs for a minor drop of 0.4 in mIoU.
In addition, the scale selection pattern learned on ImageNet is still very meaningful for the images in ADE20k: In \hyperref[fig:segmentation]{Figure \ref{fig:segmentation} (c)}, we show that classes represented via coarse tokens often correspond to uniform regions such as sky or sea, which typically occupy large regions of the image.

\subsubsection{Mixed-scale tokens for token pruning}

\begin{figure}

    \begin{center}
    \begin{minipage}[c]{0.24\textwidth}
    \centering
    \includegraphics[width=\textwidth]{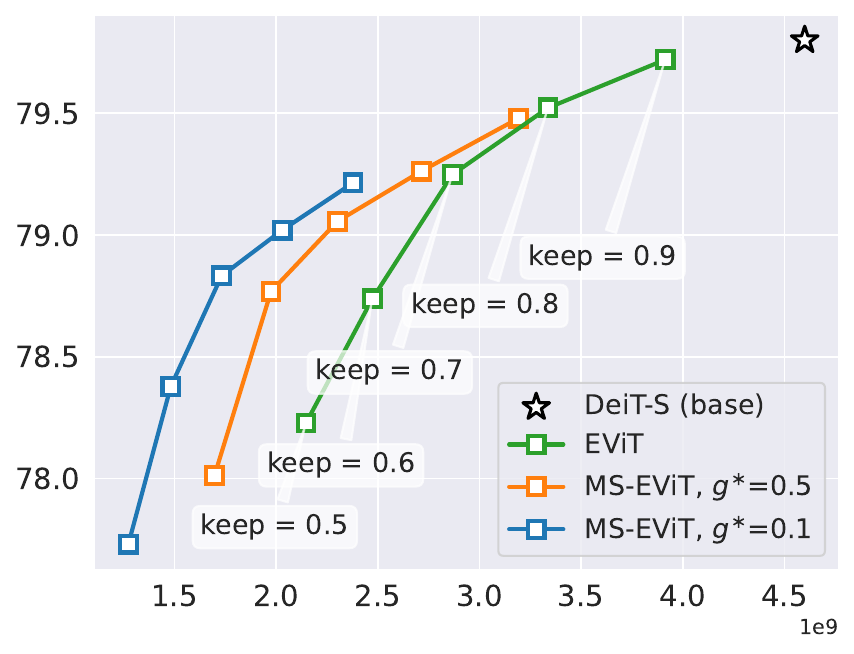}
    
    \vspace{-0.2cm}
    {\small \textbf{(a)} EViT~\cite{evitrepo} + mixed-scale}
    \end{minipage}~
    \begin{minipage}[c]{0.24\textwidth}
    \centering
    \includegraphics[width=\textwidth]{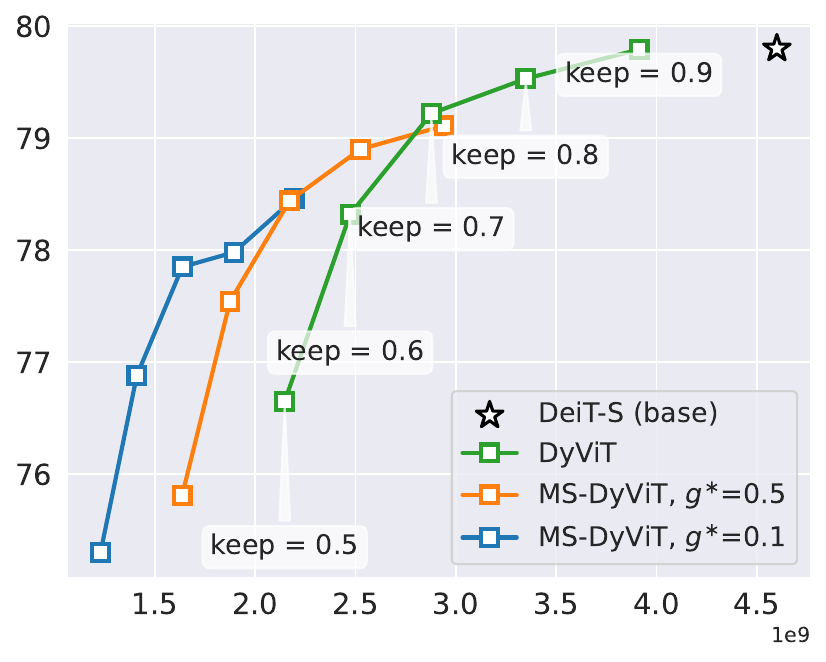}    
    
    \vspace{-0.2cm}
    {\textbf{(b)} \small DyViT~\cite{dyvitrepo} + mixed-scale}
    \end{minipage}
    \end{center}

    \vspace{-0.2cm}
    \caption{Mixed-scale tokenization combine well with token pruning methods, leading to improved efficient/accuracy trade-offs as compared to using token pruning on its own. 
    \vspace{-0.4cm}
    }
    \label{fig:pruning}
\vspace{-0.45cm}
\end{figure}

Token pruning methods iteratively discard a fixed ratio of the tokens in several intermediate layers of the transformer, based on their global class token attention~\cite{xu_evovit_2021, rao_dynamicvit_2021, yin_avit_2022, meng_adavit_2022, renggli_learning_2022, kong_spvit_2022, pan_fast_2022}.
In contrast, MSViT treats every local region individually and reduces the number of tokens before applying any transformer layer, using pixel-level information only, and without discarding any image region. 
As a result, both methods are orthogonal and select active tokens on different criteria. 
To verify how they interact, we augment two SotA pruning methods on DeiT-S, namely EViT~\cite{liang_not_2022,evitrepo} and DyViT~\cite{rao_dynamicvit_2021,dyvitrepo}, with one of our pretrained frozen gates instead of the standard ViT tokenization, and then train each model with their respective original pipeline, for different pruning ratios. We report results in \hyperref[fig:pruning]{Figure \ref{fig:pruning}}.
We observe that mixed-scale tokenization followed by token pruning in the intermediate layers complement one another well, which also introduces an interesting trade-off: Rather than using very high pruning ratios, better accuracy/efficiency performance can be reached by combining mixed-scale tokenization with a token pruning ratio. 

\subsubsection{Mixed-scale tokens for hierarchical ViTs}
Hierarchical (or Multi-scale) Vision Transformers~\cite{liu2021Swin,liu_swin_2022,Fan2021MultiscaleVT,Dong2021CSWinTA,zhang2023hivit} is a popular family of models that draw inspiration from the inductive bias of CNNs to build efficient ViT architectures: 
For instance in Swin, the image is initially split in very small tokens (4x4) which interact through local attention windows (7x7 tokens) and are progressively merged into larger tokens as depth increases.

To incorporate mixed-scale tokens in this scenario, we first run the gate to determine the fine/coarse scale pattern across the image: 
We process fine image regions with the standard Swin paradigm, at a reduced computational cost due to the lower number of tokens and potentially empty attention windows; 
Coarse tokens on the other hand are passed through a single linear embedding and merged back in the stream of tokens at layer $\ell$, once the fine tokens stream has been merged all the way up to the coarse scale. We discuss this process in more details in \hyperref[app:reshapes]{Appendix \ref{app:reshapes}}.
We report results for two values of $\ell$ in \hyperref[tab:swin_results]{Table \ref{tab:swin_results}}: The value of $\ell = 3$ yields better performance than merging  the coarse and fine tokens in an earlier block ($\ell = 2$, bottom table).
Finally, due to the multi-scale design of hierarchical ViTs, we hypothesize that the choice of $\ell$ could be further tuned with respect  to the fine scale, coarse scale and window size, and will consider this direction as future work. 

\begin{table}[tbh]
    \centering
    \resizebox{0.48\textwidth}{!}{ 
    \begin{tabular}{|c||c|c||c|c||c|c|}
    \hline
         \multirow{2}{*}{$\ell = 3$} & \multicolumn{2}{c||}{Base} & \multicolumn{2}{c||}{$g^{\ast} = 0.5$} & \multicolumn{2}{c|}{$g^{\ast} = 0.1$} \\
         \cline{2-7}
         & acc & GMACs & acc & GMACs & acc & GMACs \\
         \hline
         Swin-T & 81.0 & 4.3 & 80.0 & 3.6 & 78.8 & 3.1 \\
         \hline
         Swin-S & 83.4 & 8.6& 82.4& 6.9 & 81.4& 5.9\\
         \hline
         Swin-L & 86.0 & 33.8 & 85.4& 27.7 & 84.7& 23.3\\
         \hline
    \end{tabular}
    }
    \vspace{0.15cm}
    
    \resizebox{0.48\textwidth}{!}{ 
    \begin{tabular}{|c||c|c||c|c||c|c|}
    \hline
         \multirow{2}{*}{$\ell = 2$} & \multicolumn{2}{c||}{Base} & \multicolumn{2}{c||}{$g^{\ast} = 0.5$} & \multicolumn{2}{c|}{$g^{\ast} = 0.1$} \\
         \cline{2-7}
         & acc & GMACs & acc & GMACs & acc & GMACs \\
         \hline
         Swin-T & 81.0 & 4.3 & 80.4 & 4.0& 79.6 & 3.8\\
         \hline
         Swin-S & 83.4 & 8.6& 83.1 & 8.2 & 82.5 & 8.0\\
         \hline
         Swin-L & 86.0 & 33.8 & 85.9 & 32.6 & 85.4& 31.6\\
         \hline
    \end{tabular}
    }
    \vspace{0.05cm}
    
    \caption{We incorporate mixed-scale information in Swin~\cite{liu_swin_2022} by keeping coarse tokens determined by the gate out from the attention mechanism until layer $\ell$. We then train the models at different sizes and gate sparsities in the origina Swin training pipeline.}
    \label{tab:swin_results}
    \vspace{-0.4cm}
\end{table}

\subsection{Ablation experiments}
\label{sec:ablation}

\subsubsection{Generalized batch-shaping loss (GBaS)}

\begin{figure}[tbh]
\begin{minipage}[c]{0.23\textwidth}
\begin{center}
    \includegraphics[width=0.95\textwidth]{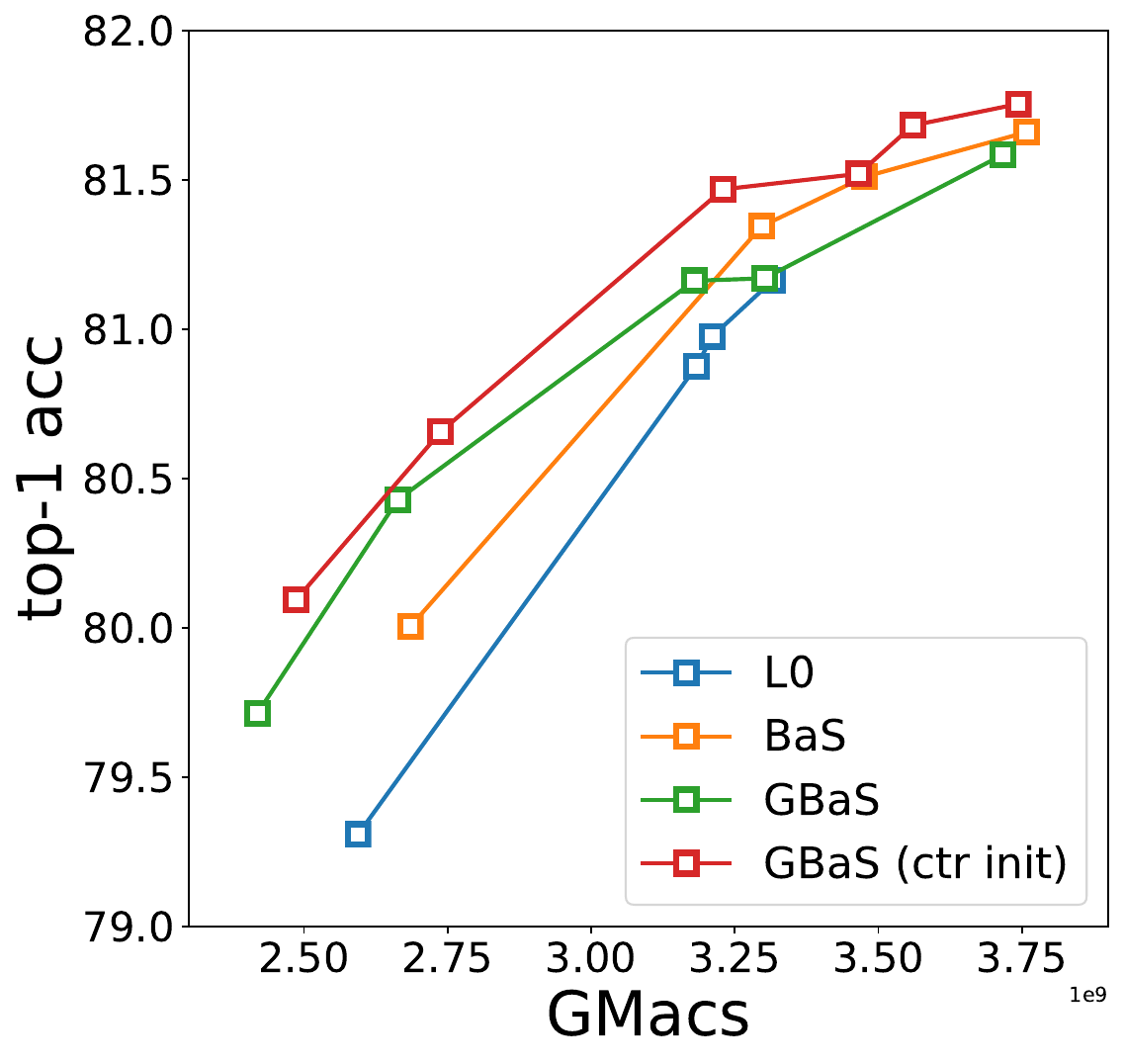}
\end{center}
\vspace{-0.43cm}

\footnotesize \textbf{(a)} Inception-style crops data augmentation (high train/test shift)
\end{minipage}
~
\begin{minipage}[c]{0.23\textwidth}
\begin{center}
    \includegraphics[width=0.95\textwidth]{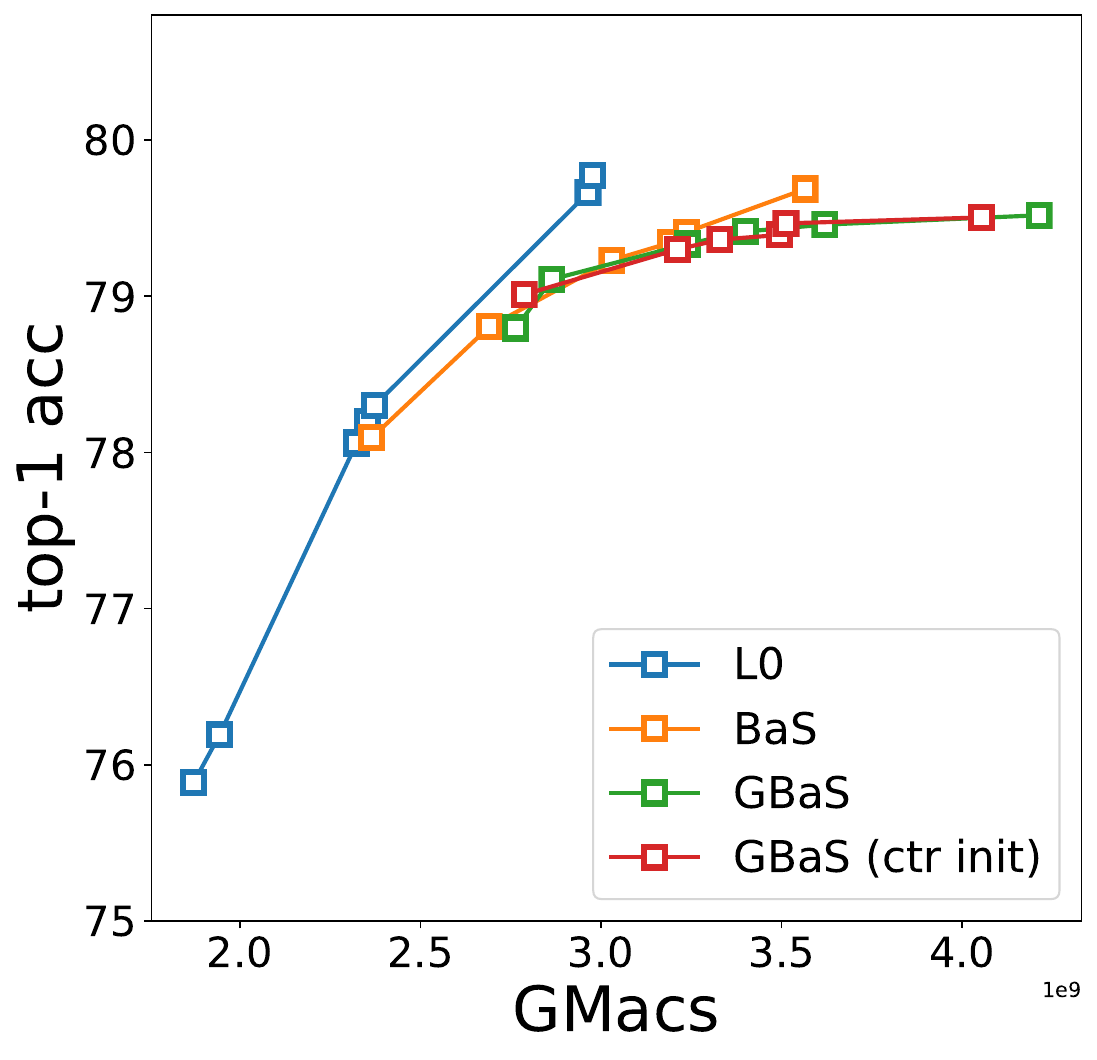}
\end{center}
\vspace{-0.43cm}

\footnotesize \textbf{(b)} Light crops data augmentation (small train/test shift)
\end{minipage}

\vspace{0.1cm}
    \caption{Accuracy-to-MACs comparison on MSViT-S/16,32 of the L0, batch-shaping (\bsloss) and generalized batch-shaping losses, with different random cropping augmentation strategies. 
    \vspace{-0.35cm}}
    \label{fig:lossesaccuracy}
\end{figure}

In \hyperref[sec:gatelearning]{Section \ref{sec:gatelearning}}, we introduced the novel \gbsloss, which allows for more control over the conditional behavior of the gate, and enables us to easily inject prior knowledge about the spatial distribution of the selected scale at initialization.
In \hyperref[fig:lossesaccuracy]{Figure \ref{fig:lossesaccuracy} (a)}, we  confirm that the best trade-off is achieved by \gbsloss{},  further improved when the learned priors are initialized as the inverse normalized distance of each spatial position to the center (\texttt{ctr init} for short).

\begin{figure}[tbh]
    \begin{center}
    \includegraphics[width=0.47\textwidth]{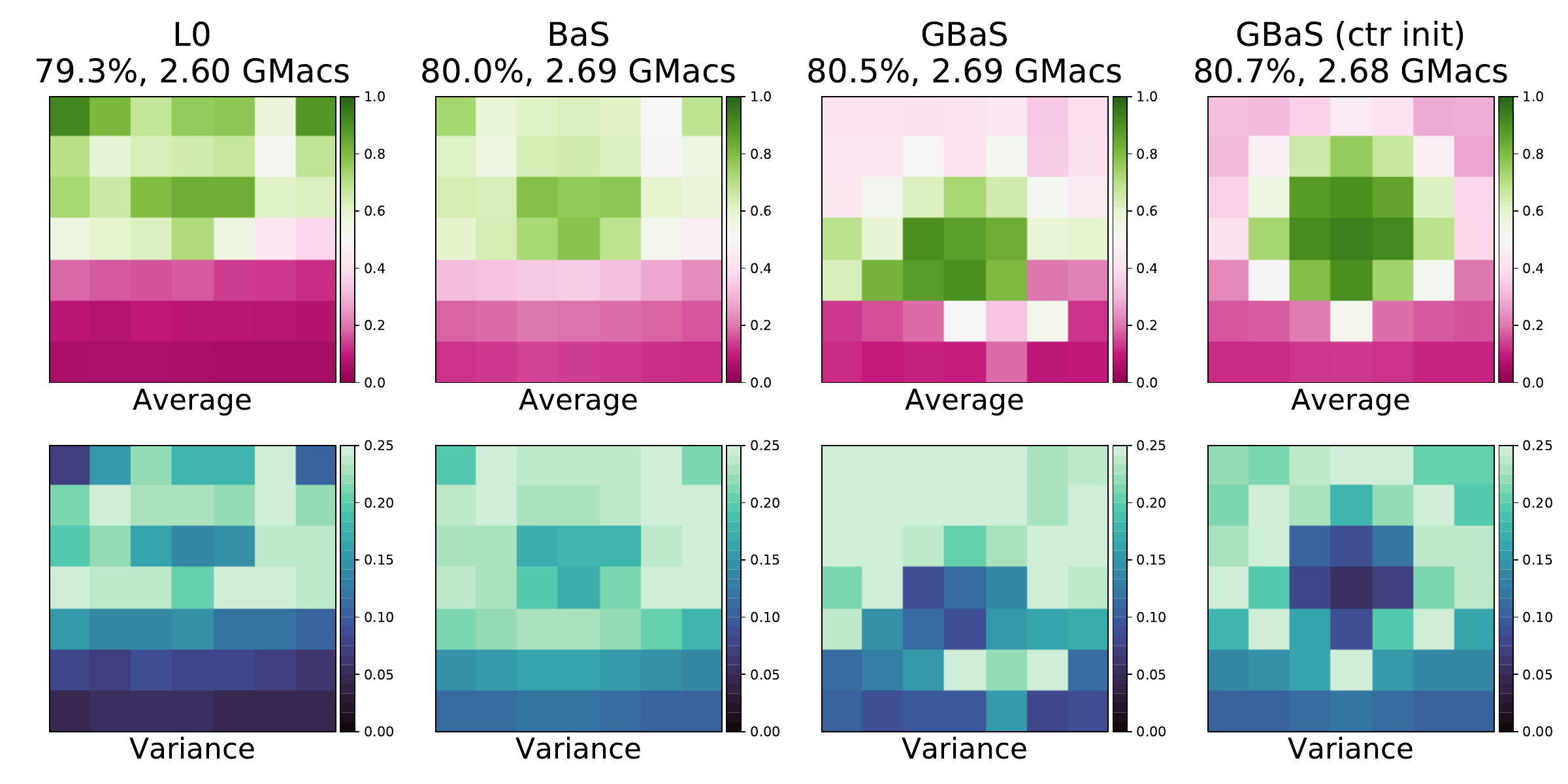}
    \end{center}
    
    \vspace{-0.5cm}
    {\small \textbf{(a)} Average (\textit{top}) and variance (\textit{bottom}) across the validation set of the learned masks selecting between \textbf{\textcolor{dollarbill}{fine}} and \textbf{\textcolor{byzantine}{coarse}} scale.}
    
    \vspace{-0.3cm}

    \begin{center}
    \includegraphics[width=0.47\textwidth]{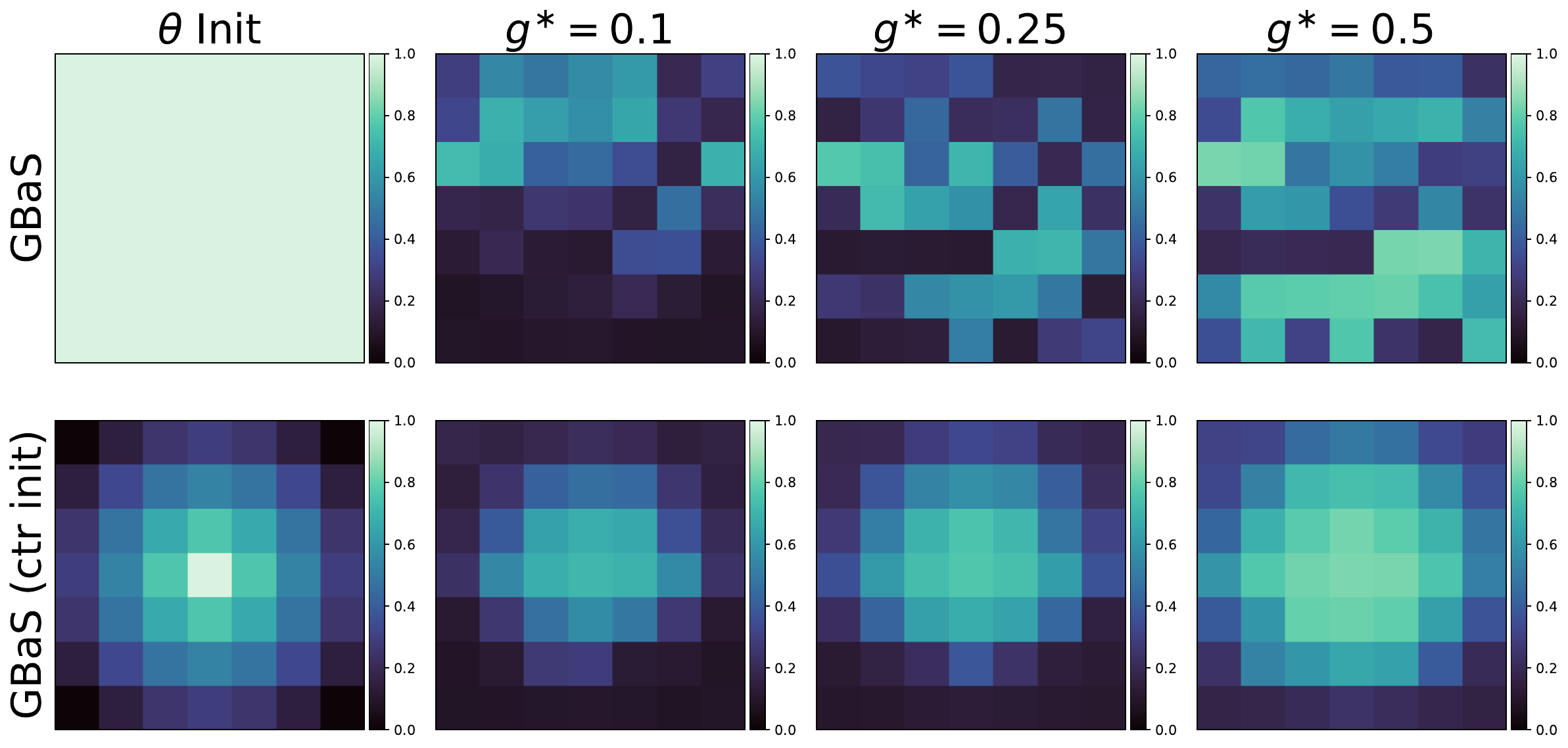}
    \end{center}
    
    \vspace{-0.5cm}
    {\small \textbf{(b)} Prior parameters $\theta$ learned with the \gbsloss{} loss with/without \texttt{ctr init} (\textit{top/bottom}).  
    The first column is initial values of $\theta$.}
    \vspace{0.15cm}
    \caption{Illustration of the masks \textbf{(a)} and priors \textbf{(b)} learned by the model with Inception-style crop data augmentations:  
    The gate is more prone to learn trivial mixed-scale patterns if not controlled properly during training using the \gbsloss{}. 
    In addition, initializing the prior parameters in \gbsloss{} with the \texttt{ctr init} is enough to guide the gate towards a more central pattern, as illustrated in \textbf{(b)}.
    }
    \label{fig:hyperpriorstrong}
    \vspace{-0.1cm}
\end{figure}

In addition, we observe that the cropping data augmentation used during training is a key factor. 
By default, we use the standard "Inception-style" cropping strategy\cite{Szegedy2016RethinkingTI} which leads to a shift between the tokens distributions at train and test time~\cite{Touvron2019FixingTT}. 
This behavior can be observed qualitatively in \hyperref[fig:hyperpriorstrong]{Figure \ref{fig:hyperpriorstrong} (a)}: 
When training with Inception crops, there is high uncertainty on the location of objects, and the L0 loss ends up stuck on a trivial static pattern early during training. 
On the other hand, \gbsloss{} learns more centered mixed-scale patterns, but still captures uncertainty over spatial positions through the learned priors (\hyperref[fig:hyperpriorstrong]{Fig. \ref{fig:hyperpriorstrong} (b) \textit{top row}}), which can be further reduced with  \texttt{ctr init} (\hyperref[fig:hyperpriorstrong]{\textit{bottom row}}).

In contrast, with a lighter cropping strategy, all losses learn that, on average, fine scale tokens are more likely to appear in the center-top of the image, where the object to categorize usually lies (see \hyperref[app:gbas]{Appendix \ref{app:gbas}}). 
As a result, all batch-shaping variants perform equally well, and the  L0 loss even outperforms them in some cases (\hyperref[fig:lossesaccuracy]{Figure \ref{fig:lossesaccuracy} (b)}).

In summary, \gbsloss{} is more robust to train/test discrepancy than other losses; Nevertheless when there is no notable distribution shift, then even a simple L0 sparsity loss can reach a similar or even better performance.

\subsubsection{Benefits of learning a dynamic gate}
\label{sec:dynamic}
\begin{figure}[t]
    \includegraphics[width=0.5\textwidth,trim={0 0 7cm 0},clip]{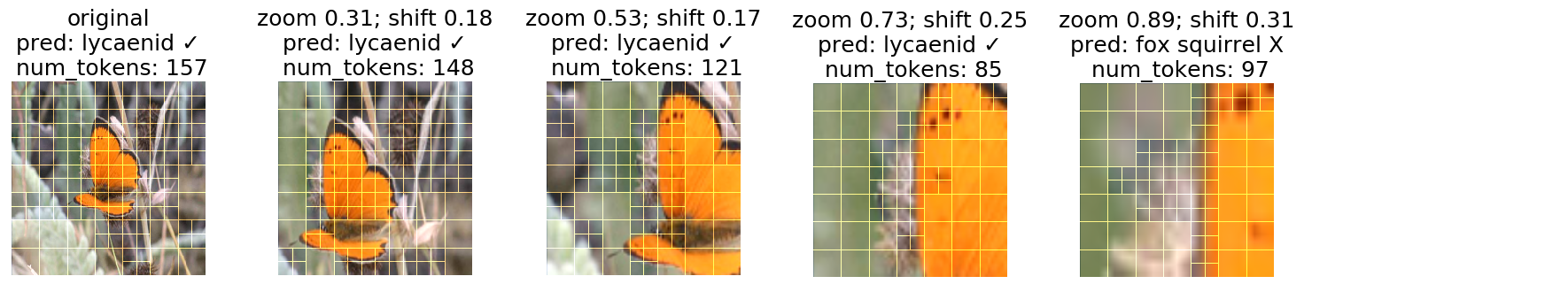}

    \vspace{-0.01cm}
    \caption{Example of the learned dynamic gate outputs when applied on random image zooms and shifts of the validation dataset}
    \label{fig:robustnessshort}
    \vspace{-0.25cm}
\end{figure}

In \hyperref[fig:robustnessshort]{Figure \ref{fig:robustnessshort}}, we illustrate how the learned gate module dynamically adapts the mixed-scale pattern, hence the computation cost, to the input image content. 
We further investigate and highlight this behavior quantitatively in \hyperref[app:robust]{Appendix \ref{app:robust}}, in which we compare using a learned gate versus using a fixed oracle mixed-resolution pattern where all central patches are at the fine scale, and any region further than a certain radius from the center is kept at  coarse scale. 

\section{Conclusions}
In this work, we proposed a dynamic mixed-scale tokenization scheme for ViT, MSViT, via a novel conditional gating mechanism.
The gate is agnostic to the choice of transformer backbone, and is trained jointly with it, in a single-stage, with mixed-scale tokens.
To improve the conditional behaviour of the gate, we proposed a generalization of  batch-shaping~\cite{Bejnordi2020BatchshapingFL}  to better handle \textit{multi-dimensional distributions}. GBaS improves results and allows for easier and better initialization of the gates. 
Our experiments on image classification and semantic segmentation show that the proposed dynamic tokenization enhances computational efficiency by reducing the number of input tokens, with minimal impact on performance. 
For both tasks, the gate learns to represent uniform and background regions with coarse tokens and higher entropy regions with fine ones. 
%

\clearpage
{\small
\bibliographystyle{ieee_fullname}
\bibliography{egbib}
}

\setcounter{section}{0}  
\renewcommand{\thesection}{\Alph{section}}

\onecolumn
\begin{center}
\Large{Appendix}
\end{center}

\section{Additional qualitative results on ImageNet}
\label{app:extra_results}
\begin{figure}[!h]
    \centering
    \includegraphics[width=0.5\textwidth]{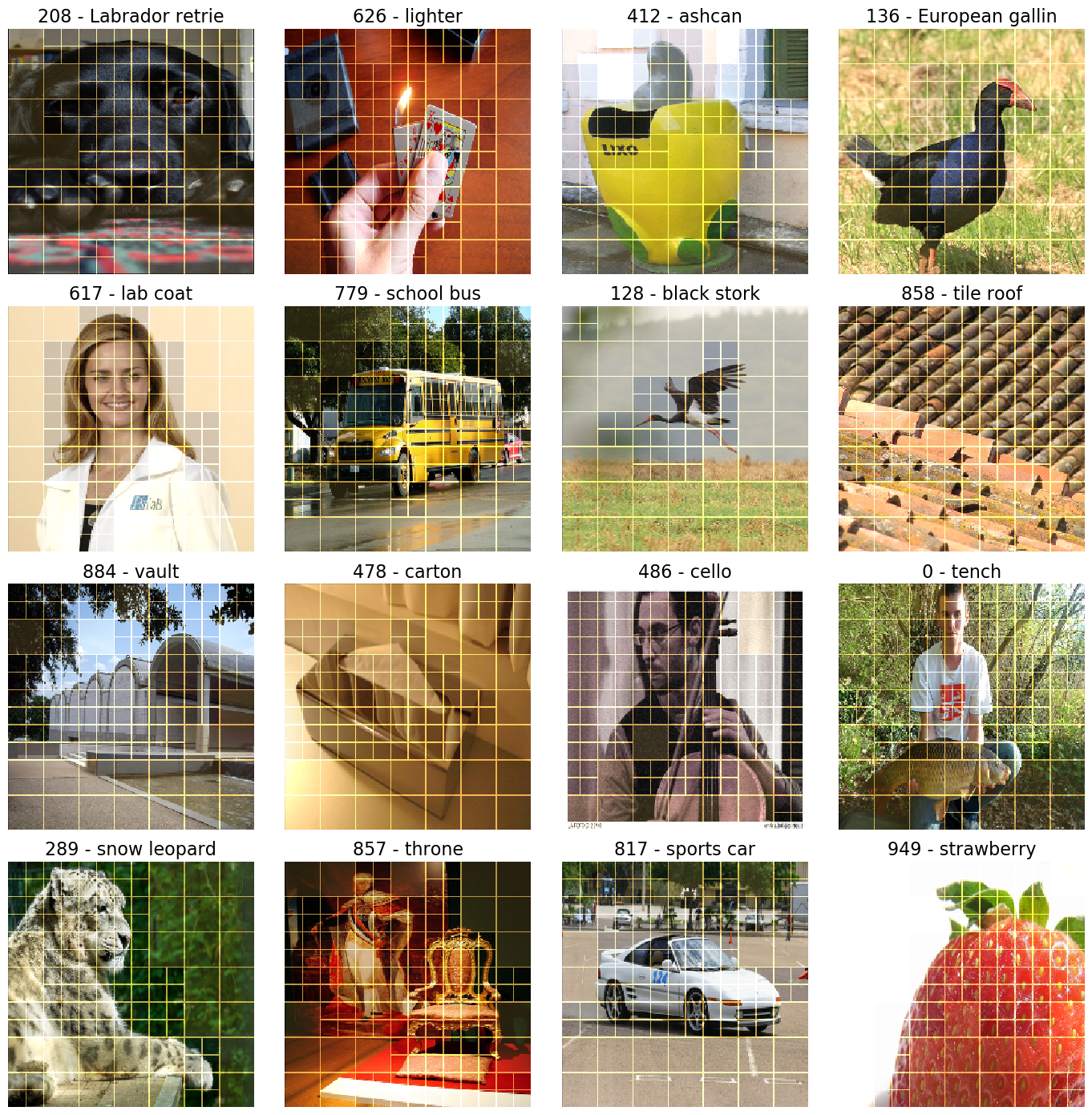}%
    \includegraphics[width=0.5\textwidth]{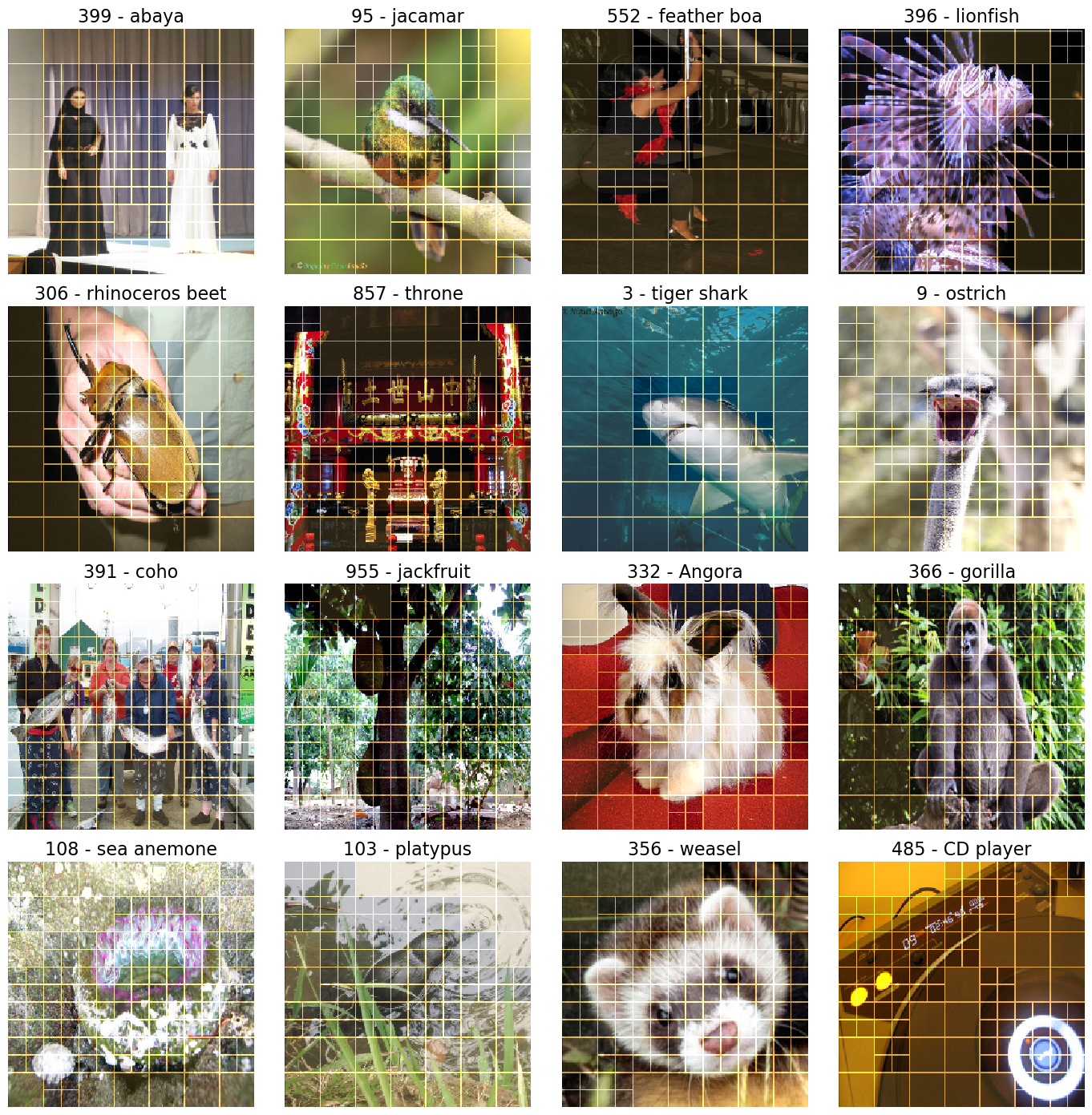}
    
    \includegraphics[width=0.5\textwidth]{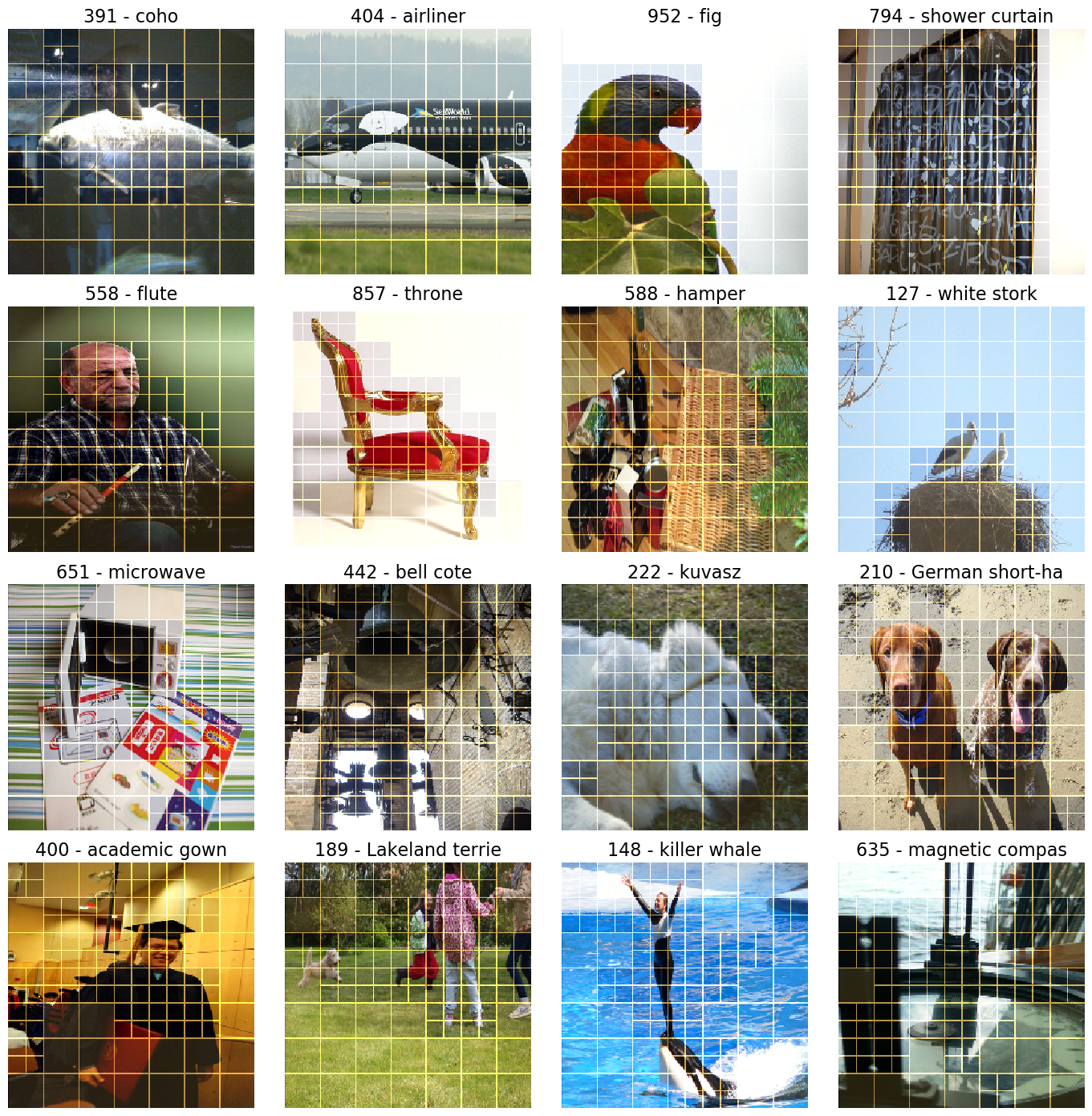}%
    \includegraphics[width=0.5\textwidth]{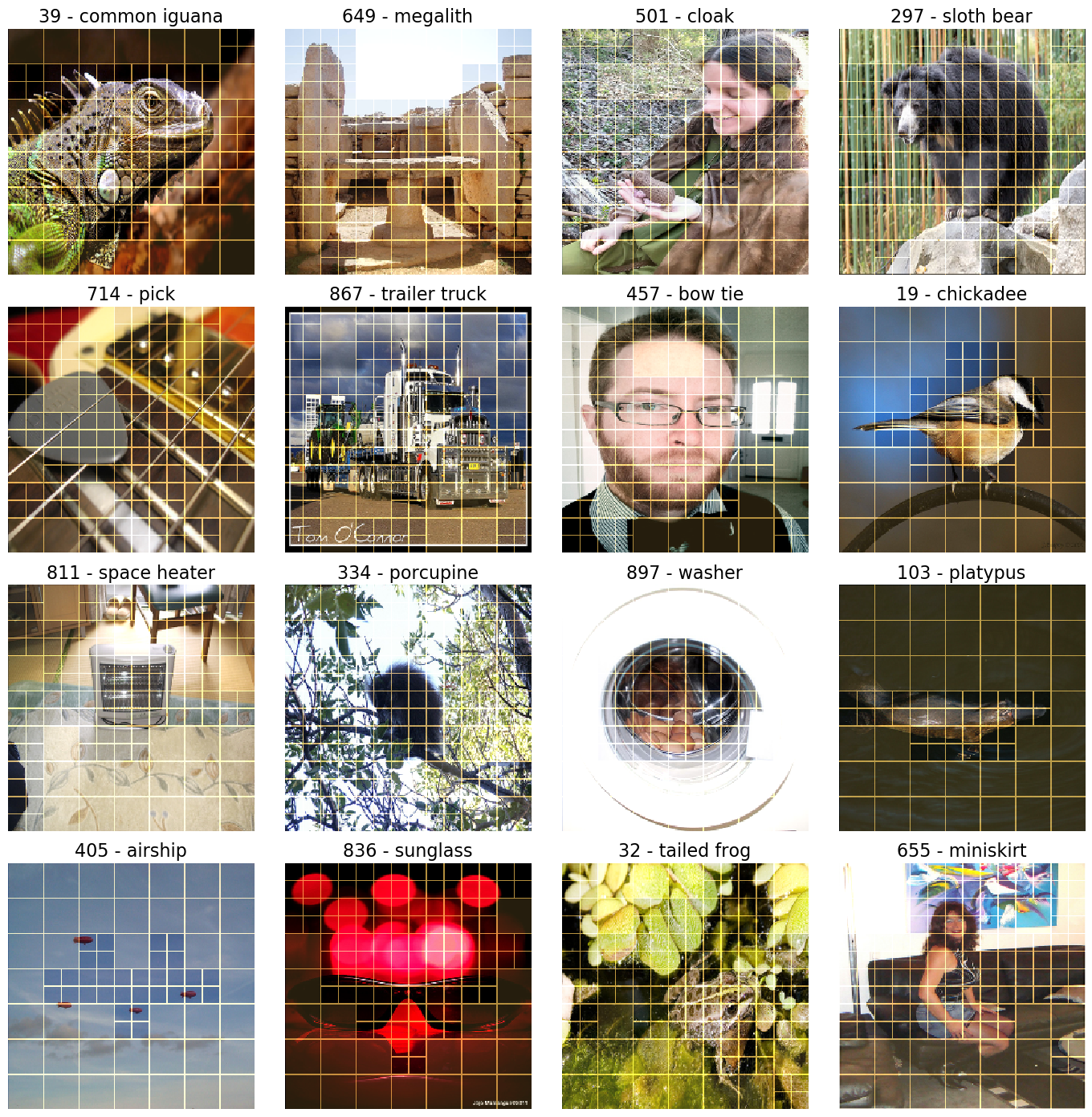}
    \caption{Non-curated qualitative   examples of scale selection masks output by the gating module of MSViT-S/\{16, 32\}. The model was trained on 224px ImageNet images to choose  between the coarse 
    (32px, \tikzinlinecoarse{}) and the fine (16px, \tikzinlinefine{}) token scale. \textit{Best seen zoomed}.
    }
\end{figure}
\clearpage
\twocolumn

\section{Additional results on Image classification}
\label{app:fullclass}
\begin{figure}[!htb]
    \begin{minipage}[c]{0.23\textwidth}
    \centering 
    \includegraphics[width=\textwidth]{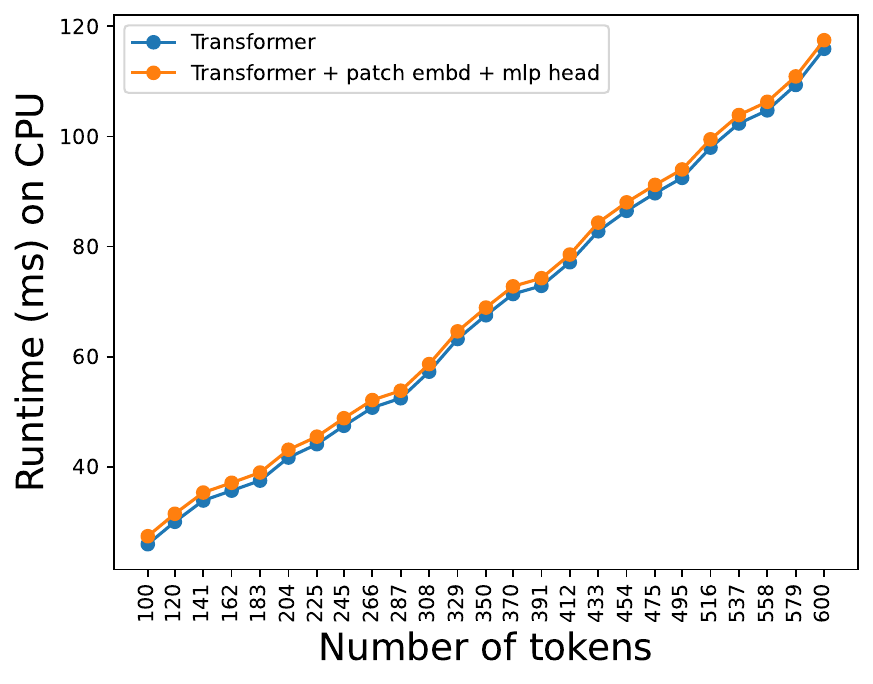}

    \small (a)  CPU
    \end{minipage}
    \begin{minipage}[c]{0.23\textwidth}
    \centering 
    \includegraphics[width=\textwidth]{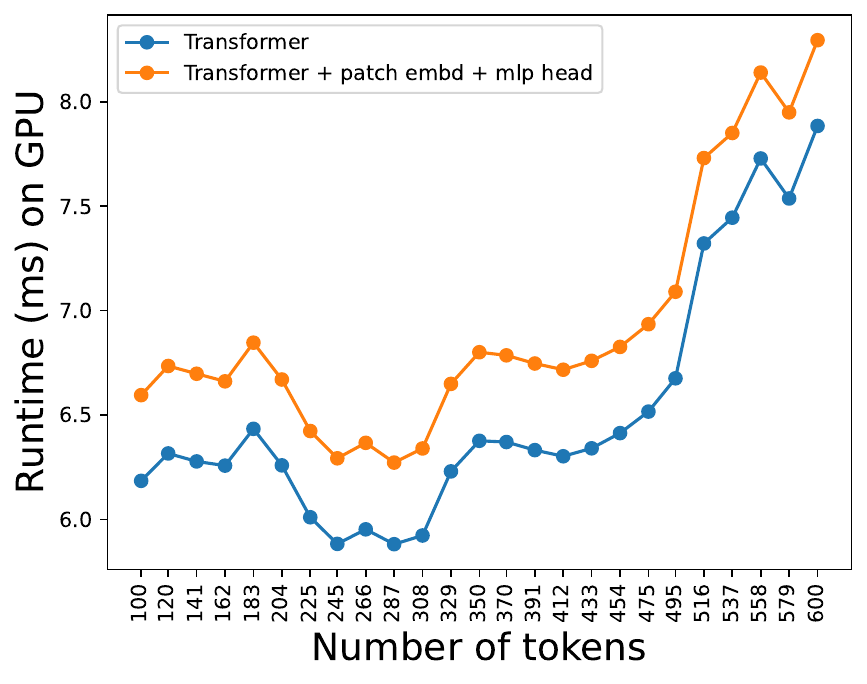}

    \small (b) GPU (RTX 2080 Ti)
    \end{minipage}
    
    \caption{Average runtime in milliseconds of ViT-S for different number of input tokens on two diferent devices, simulated and averaged on 1000 random samples. The \textcolor{blue}{blue} line is the cost of the transformer only, while the \textcolor{orange}{orange} line additionally includes the cost of the patch embedding and MLP classifier.}
    \label{fig:timings}
\end{figure}

In this section we report additional results on the ImageNet classification benchmark.
First, in \hyperref[fig:timings]{Figure \ref{fig:timings}}, we plot the average simulated runtimes of standard ViT-S with different number of tokens. While the exact trend depends on the device due to dedicated optimizations, we verify that reducing the number of tokens leads to concrete runtime improvements.

Then, in \hyperref[tab:extendedinet]{Table \ref{tab:extendedinet}}, we show an extended version of \hyperref[tab:main_results]{Table \ref{tab:main_results}}, with additional results for \textbf{(i)} MS-ViT-L, \textbf{(ii)} more computational budgets and \textbf{(iii)} latency results averaged across the images in the ImageNet validation set.

\begin{figure}[!htb]
    \centering
    \includegraphics[width=0.42\textwidth]{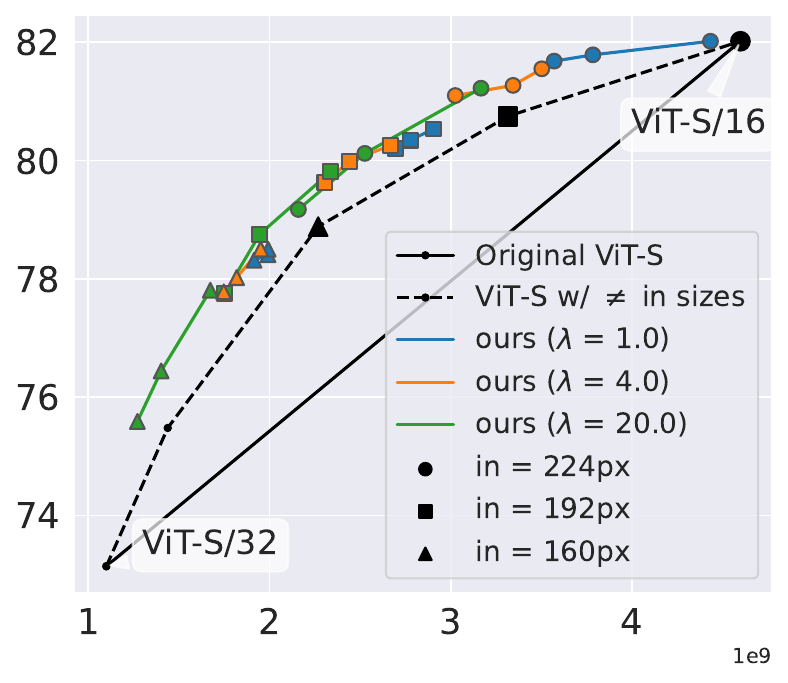}
    \caption{Full hyperparameter sweep for MSViT-S/16 experiments (top-1 accuracy versus MACs). Each line corresponds to a configuration of gate loss weight $\lambda$ and input image size. Each point on a line corresponds to a different gate sparsity target $g^{\ast} \in \{ 0.25, 0.5, 0.75\}$}
    \label{fig:allimagesizes}
\end{figure}

Finally, in \hyperref[fig:allimagesizes]{Figure \ref{fig:allimagesizes}}, we report the results of the full hyperparameter sweep on MSViT-S for different input image sizes: As expected, both the gate sparsity target $g^{\ast}$ and the gate loss weight $\lambda$ can be increased to obtain sparser gates. In addition, we observe that all MSViT points lie on the same Pareto front which suggests the relative performance of MSViT is robust to these hyperparameter choices (gate loss weight, sparsity target and input image size). 

\begin{table}[tbh]
    \centering
  \resizebox{0.48\textwidth}{!}{  
    \begin{tabular}{|c||c|c|c|c|c|||c|c|}
    \hline
         DeiT-Small & Avg \# & GMACs & CPU time  & GPU time  & \multicolumn{2}{c|}{accuracy}\\
         \cline{6-7}
          backbone & tokens & (avg) & (ms) & (ms) & top-1 & top-5\\
         \hline
         \hline
        \rowcolor{mygray} DeiT-S/16 in=160 & 100 & 2.27 & 18.70 & 6.06 & 75.86 & 92.84\\
        \hline
        \textbf{MSDeiT}-S/16,32 in=224 & 94 & 2.20 & 18.01 & 6.00 & 75.90 & 92.68\\
        \textbf{MSDeiT}-S/16,32 in=224 & 97 & 2.27 & 18.22 & 6.02 & 76.99 & 93.38\\
         \hline
         \hline
        \rowcolor{mygray} DeiT-S/16 in=192 & 144 & 3.32 & 24.04 & 6.26 & 77.79 & 93.96\\
        \hline
        \textbf{MSDeiT}-S/16,32 in=224 & 116 & 2.72 & 21.18 & 6.28 & 77.79 & 93.99\\
        \textbf{MSDeiT}-S/16,32 in=224 & 142 & 3.32 & 24.24 & 6.20 & 78.76 & 94.32\\
         \hline
         \hline
        \rowcolor{mygray} DeiT-S/16 in=224 & 196 & 4.60 & 31.65 & 6.07 & 79.85 & 94.57\\
        \hline
        \textbf{MSDeiT}-S/16,32 in=224 & 173 & 4.08 & 27.70 & 6.19 & 79.38 & 94.38\\
        \hline
    \end{tabular}
    }
    
    \vspace{0.35cm}

  \resizebox{0.48\textwidth}{!}{  
    \begin{tabular}{|c||c|c|c|c||c|c|}
    \hline
         ViT-Tiny & Avg \# & GMACs  &  CPU time & GPU time & \multicolumn{2}{c|}{accuracy}\\
         \cline{6-7}
          backbone & tokens & (avg) & (ms) & (ms) & top-1 & top-5\\
        \hline
        \hline
        \rowcolor{mygray} ViT-Ti/16 in=160 & 100 & 0.60 & 9.24 & 5.97 & 71.63 & 90.68\\
            \hline
        \textbf{MSViT}-Ti/16,32 in=224 & 95 & 0.60 & 8.99 & 5.98 & 72.57 & 91.32\\
        \hline
        \hline
        \rowcolor{mygray} ViT-Ti/16 in=192 & 144 & 0.89 & 11.56 & 6.03 & 74.24 & 92.22\\
            \hline
       \textbf{ MSViT}-Ti/16,32 in=224 & 124 & 0.78 & 11.04 & 6.04 & 74.27 & 92.22\\
       \textbf{ MSViT}-Ti/16,32 in=224 & 138 & 0.88 & 11.49 & 6.00 & 74.93 & 92.54\\
        \hline
        \hline
        \rowcolor{mygray} ViT-Ti/16 in=224 & 196 & 1.25 & 13.26 & 5.98 & 76.00 & 93.26\\
            \hline
        \textbf{MSViT}-Ti/16,32 in=224 & 154 & 0.98 & 11.89 & 5.88 & 75.51 & 92.98\\
         \hline
    \end{tabular}
    }
    \vspace{0.35cm}
    
      \resizebox{0.48\textwidth}{!}{  
    \begin{tabular}{|c||c|c|c|c||c|c|}
    \hline
         ViT-Small & Avg \# & GMACs  & CPU time & GPU time & \multicolumn{2}{c|}{accuracy}\\
         \cline{6-7}
          backbone & tokens & (avg) & (ms) & (ms) & top-1 & top-5\\
        \hline
        \hline
            \rowcolor{mygray} ViT-S/16 in=128 & 64 & 1.44 & 15.35 & 5.94 & 75.48 & 93.08\\
            \hline
            \textbf{MSViT}-S/16,32 in=224 & 75 & 1.76 & 16.33 & 5.95 & 77.16 & 94.14\\
        \hline
        \hline
            \rowcolor{mygray} ViT-S/16 in=160 & 100 & 2.27 & 18.60 &  6.06& 78.88 & 94.95\\
            \hline
            \textbf{MSViT}-S/16,32 in=224 & 91 & 2.13 &  17.64 & 5.97 &  78.88 & 95.02\\
            \textbf{MSViT}-S/16,32 in=224 & 98 & 2.30 & 18.60 & 6.04 & 79.51 & 95.33\\
        \hline
        \hline
        \rowcolor{mygray} ViT-S/16 in=192 & 144 & 3.32 & 24.11 & 6.18 & 80.75 & 95.86\\
            \hline
        \textbf{MSViT}-S/16,32 in=224 & 120 & 2.82 & 21.71 & 6.22 & 80.74 & 95.92\\
        \textbf{MSViT}-S/16,32 in=224 & 138 & 3.23 & 23.68 & 6.19 & 81.47 & 96.14\\
        \hline
        \hline
        \rowcolor{mygray} ViT-S/16 in=224 & 196 & 4.60 & 31.46 & 6.08 & 82.02 & 96.45\\
            \hline
        \textbf{MSViT}-S/16,32 in=224 & 187 & 4.43 & 29.30 & 6.25 & 82.02 & 96.44\\
        \hline
        \hline
        ViT-S/16 in=288 & 324 & 7.97 & 53.79 & 6.18 & 83.34 & 96.93\\
            \hline
        \textbf{MSViT}-S/16,32 in=384 & 314 & 7.92 & 52.67 & 6.02 & 83.56 & 97.10\\
        \textbf{MSViT}-S/16,32 in=384 & 286 & 7.16 & 47.53 & 6.09 & 83.34 & 96.99\\
        \hline
        \hline
        ViT-S/16 in=320 & 400 & 10.11 & 68.20 & 6.25 & 83.85 & 97.10\\
            \hline
        \textbf{MSViT}-S/16,32 in=384 & 359 & 9.19 & 60.16 & 6.18 & 83.84 & 97.20\\
        \textbf{MSViT-S}/16,32 in=384 & 382 & 9.80 & 66.12 & 6.21  & 83.93 & 97.18\\
        \hline
        \hline
        ViT-S/16 in=384 & 576 & 15.49 & 104.58 & 6.26 & 84.20 & 97.32\\
            \hline
        \textbf{MSViT}-S/16,32 in=384 & 428 & 11.14 & 76.76 & 6.16  & 84.14 & 97.31\\
         \hline
    \end{tabular}
    }

    \vspace{0.35cm}
      \resizebox{0.48\textwidth}{!}{  
    \begin{tabular}{|c||c|c|c|c||c|c|}
    \hline
         ViT-Large & Avg \# & GMACs  & CPU time & GPU time & \multicolumn{2}{c|}{accuracy}\\
         \cline{6-7}
          backbone & tokens & (avg) & (ms) &  (ms) & top-1 & top-5\\
        \hline
        \hline
            \rowcolor{mygray} ViT-L/16 in=160 & 100 &  31.08 & 185.44 & 12.74 & 81.70 & 96.14\\
            \hline
            \textbf{MSViT}-L/16,32 in=160 & 89& 27.48 & 172.29 & 12.37 & 81.73 & 96.13\\
            \textbf{MSViT}-L/16,32 in=192 & 84& 25.93 & 169.63 & 12.46 & 81.67 & 96.14\\
        \hline
        \hline
        \rowcolor{mygray} ViT-L/16 in=192 & 144 & 44.9 & 233.27 & 14.49 & 82.91 & 96.61\\
            \hline
        \textbf{MSViT}-L/16,32 in=192 & 111 & 34.5 & 195.24 & 12.38 & 82.46 & 96.45\\
            \hline
    \end{tabular}
    }
    \vspace{0.1cm}
    
    \caption{Extended results for \hyperref[tab:main_results]{Table \ref{tab:main_results}} with additional configurations, and average latency per image on CPU and GPU (RTX 2080 Ti). Both the MACs and latencies are estimated with the deepspeed library~\cite{deepspeed}.}
    \label{tab:extendedinet}
\end{table}

\section{Hyperparameters}
\label{app:hyperparameter}
\subsection{VIT backbone}
For ViT experiments, we finetune ImageNet-21k pretrained checkpoints to ImageNet. We use the same finetuning setup as the one from the official ViT repository~\cite{vitrepo}, except we train for 20 epochs instead of 8: 

\begin{verbatim}
batch-size: 512
num-gradacc-steps: 1
data-augmentation: crop+fliplr
num-epochs: 20
optimizer: "SGD"
lr: 0.03
momentum: 0.9
gradient-clipping: 1.0
weight-decay: 0.0
num-warmup-epochs: 0.50 
lr-scheduler: cosine
\end{verbatim}

\subsection{DeiT backbone}
For DeiT, we also follow standard available finetuning pipelines e.g. from~\cite{deitrepo,evitrepo,dyvitrepo}.  In particular, the most notable differences with the ViT finetuning pipeline are:
\begin{itemize}
\item The data loader uses a different normalization and bicubic resizing
\item We use the \texttt{AdamW} optimizer with \texttt{lr = 2e-5} (after sweeping over the range \texttt{lr} $\in \{5e-4, 1e-4, 2e-5 \}$)
\item additional small optimization choices: no gradient clipping, small weight decay and label smoothing with a weight of 0.1
\end{itemize}

\subsection{Gate Hyperparameters}
For training the gate, we use the same optimizer and learning rate as the model features. The only difference is that we use a longer warmup period to account for the fact that the gate is trained from scratch. 
For \gbsloss{}, we observe that the temperature in the Relaxed Bernoulli and the variance of the hyperprior, as long as they do not take very extreme values, do not strongly impact the final learned gate, but rather the training speed.
Therefore, we fix their values in all experiments and instead sweep over the gate loss weight which also directly impacts the gate's training speed.

\begin{verbatim}
num-gate-warmup-epochs: 6
relaxed_bernoulli_temperature: 0.3
hyperprior_variance: 0.1
\end{verbatim}

Finally as mentioned in experiments,  we sweep over the gate target $g^\ast \in \{0.5, 0.25, 0.1\}$ and loss weight $\lambda \in \{1, 4, 20\}$ to obtain models at various compute budgets.

\section{Additional segmentation results}
\label{app:adek}
\begin{figure}[!thb]
    \centering
    \includegraphics[width=0.24\textwidth]{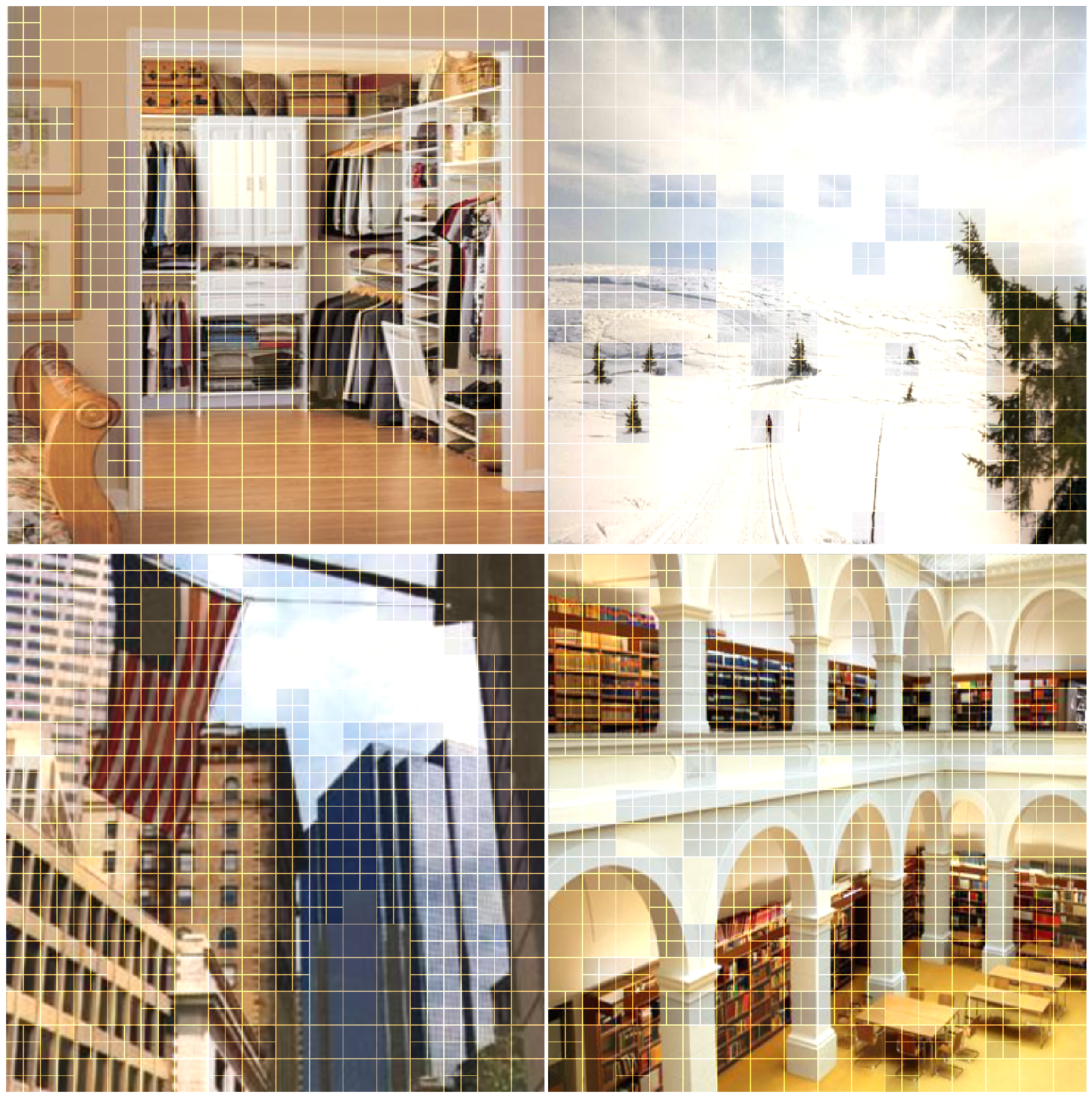}\includegraphics[width=0.24\textwidth]{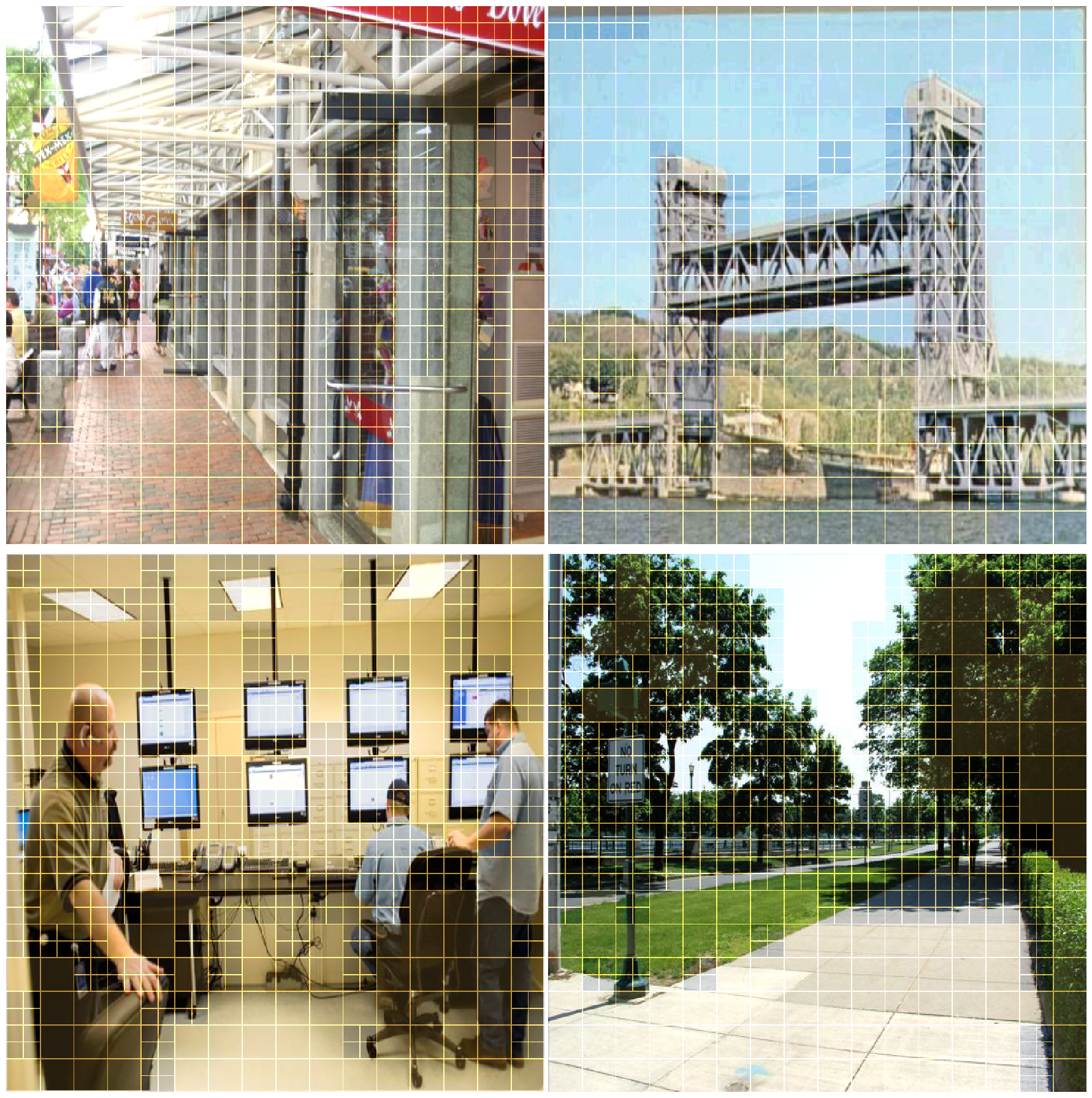}
    
    \includegraphics[width=0.24\textwidth]{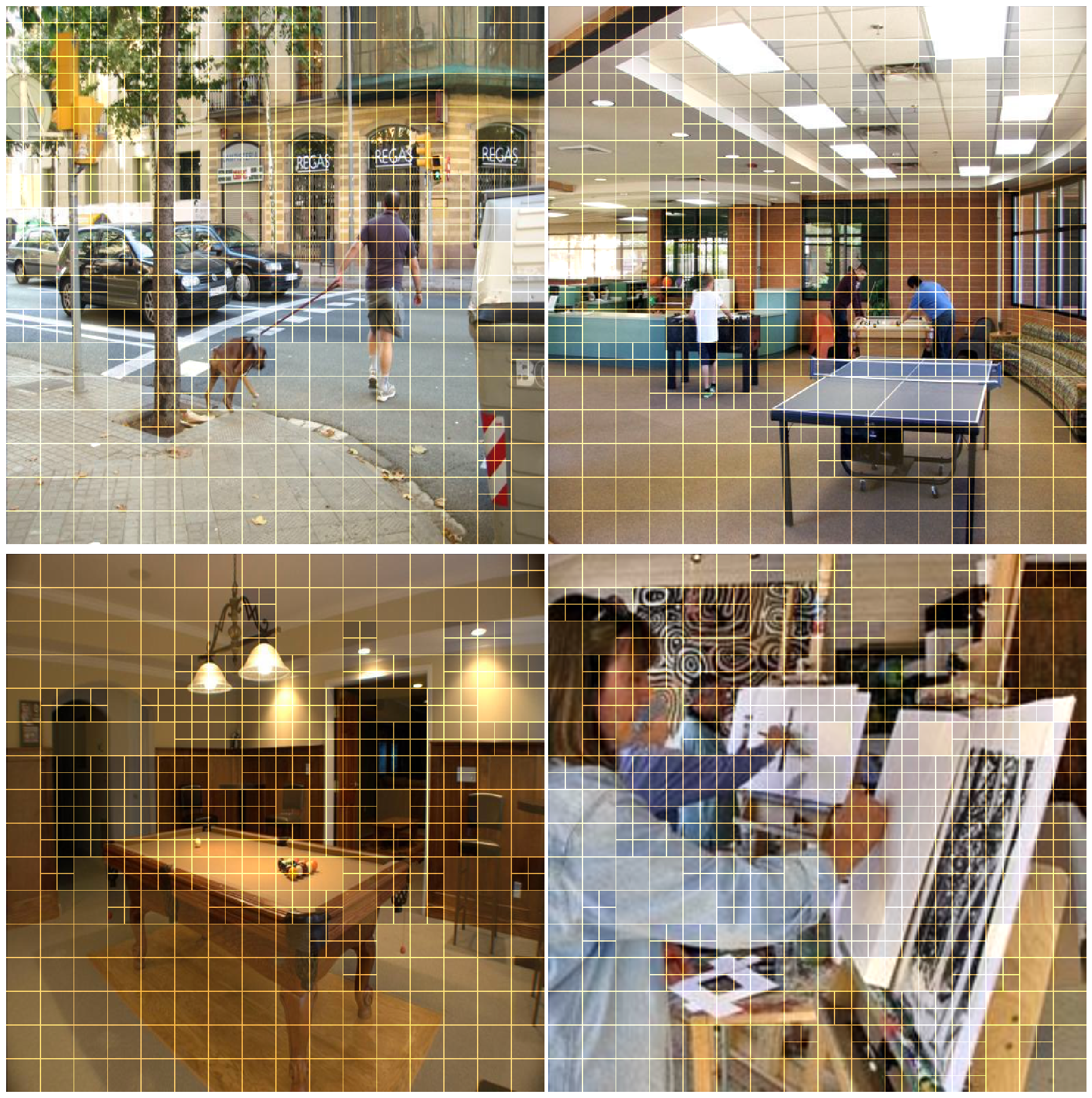}\includegraphics[width=0.24\textwidth]{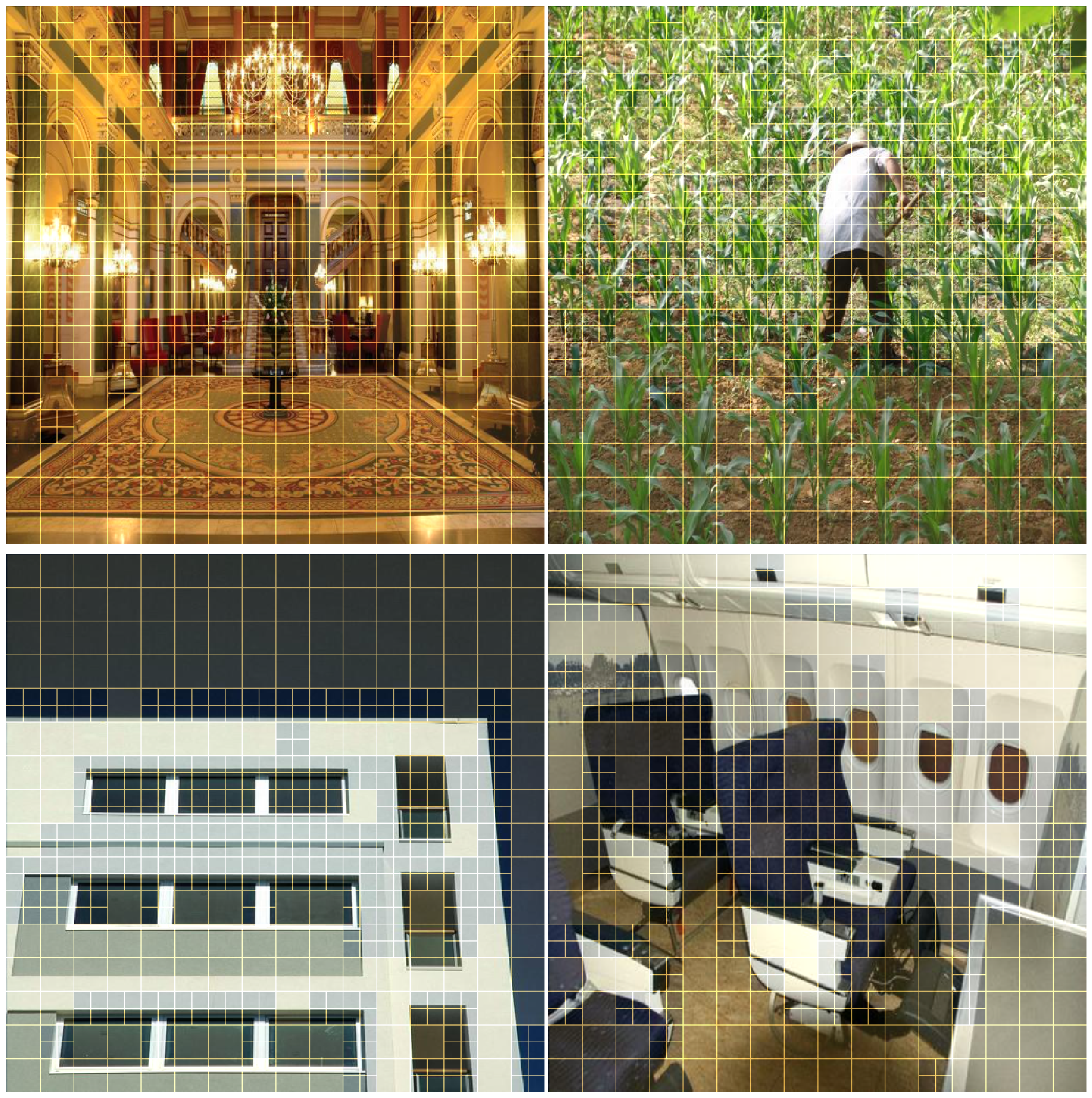}
    \caption{Direct transfer of a gate trained on ViT-S/16 224px images for ImageNet to ADE20k for 512px images}
    \label{fig:segmasks}
\end{figure}

In this section, we report additional  results for the segmentation experiments. 
First, in \hyperref[fig:segmasks]{Figure \ref{fig:segmasks}}, we visualize some masks output by a ViT-S/16 finetuned on ImageNet when directly applied on 512x512 ADE20K~\cite{Zhou2017ScenePT} images,  without extra finetuning on ADE20k.
As we can see, the mixed-scale selection patterns transfer well from ImageNet to ADE20K.

\begin{figure}[!tbh]
    \setlength{\fboxsep}{0pt}%
    \setlength{\fboxrule}{0.4pt}%
    \centering 
    
    \newcommand\idx{199}
    \begin{minipage}[b]{0.155\textwidth}
    \centering
     \footnotesize Mixed-scale mask 
     \vspace{0.023cm}
    \fbox{\includegraphics[width=\textwidth]{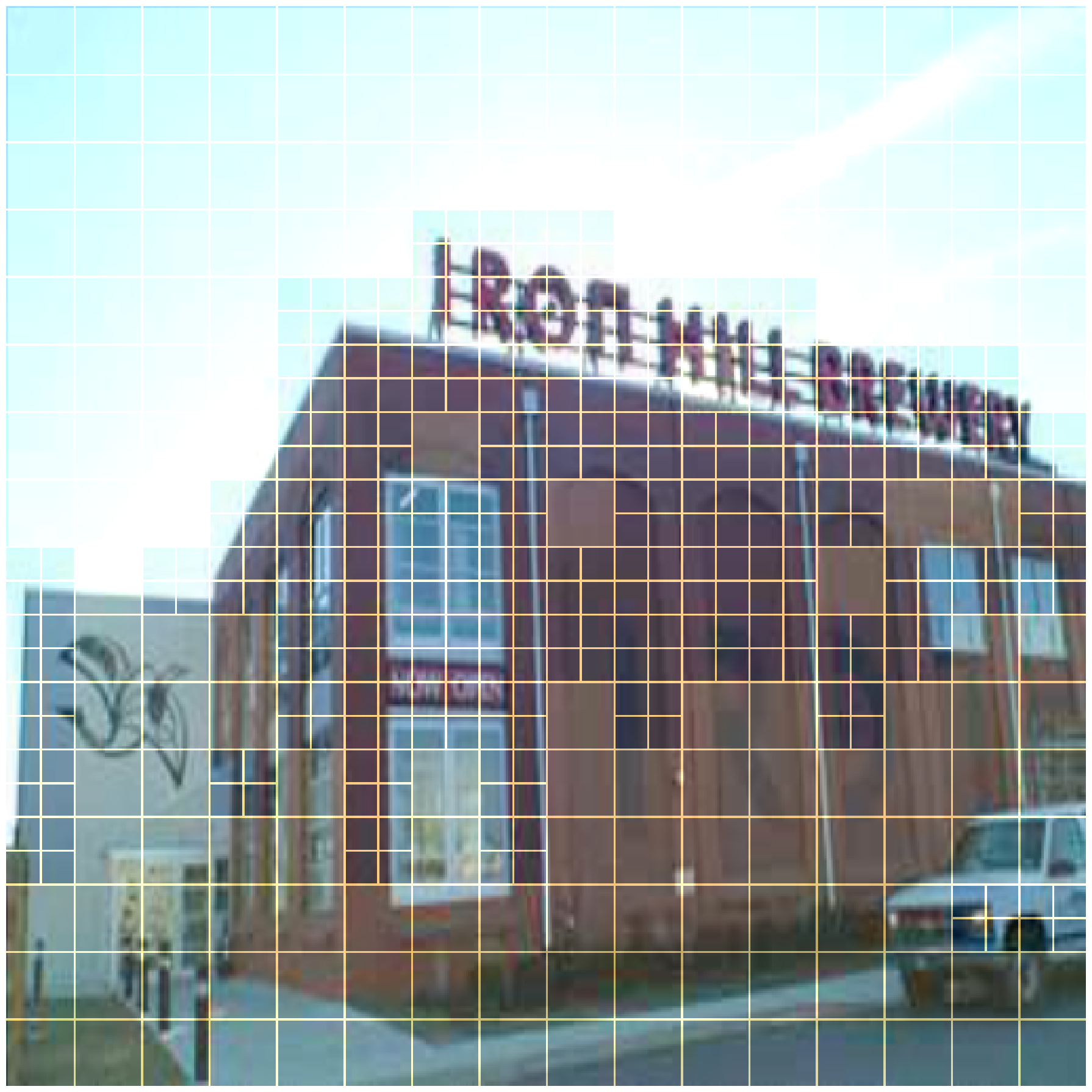}}
    \end{minipage}
    \begin{minipage}[b]{0.155\textwidth}
    \centering
    \small \footnotesize \textbf{MS}Seg-S/16,32 
    \fbox{\includegraphics[width=\textwidth, height=\textwidth]{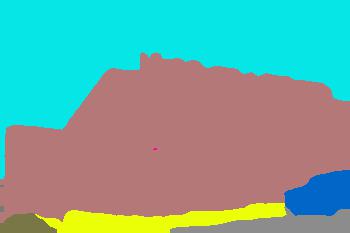}}
    \end{minipage}
    \begin{minipage}[b]{0.155\textwidth}
    \centering
    \footnotesize Seg-S/16
    \fbox{\includegraphics[width=\textwidth, height=\textwidth]{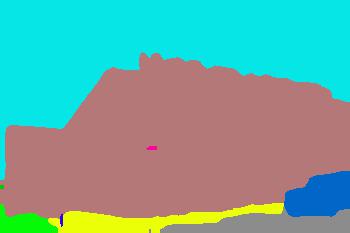}}
    \end{minipage}
    {\small ADE\_val\_00000\idx}

    \renewcommand\idx{109}
    \begin{minipage}[b]{0.155\textwidth}
    \centering
     \vspace{0.023cm}
    \fbox{\includegraphics[width=\textwidth]{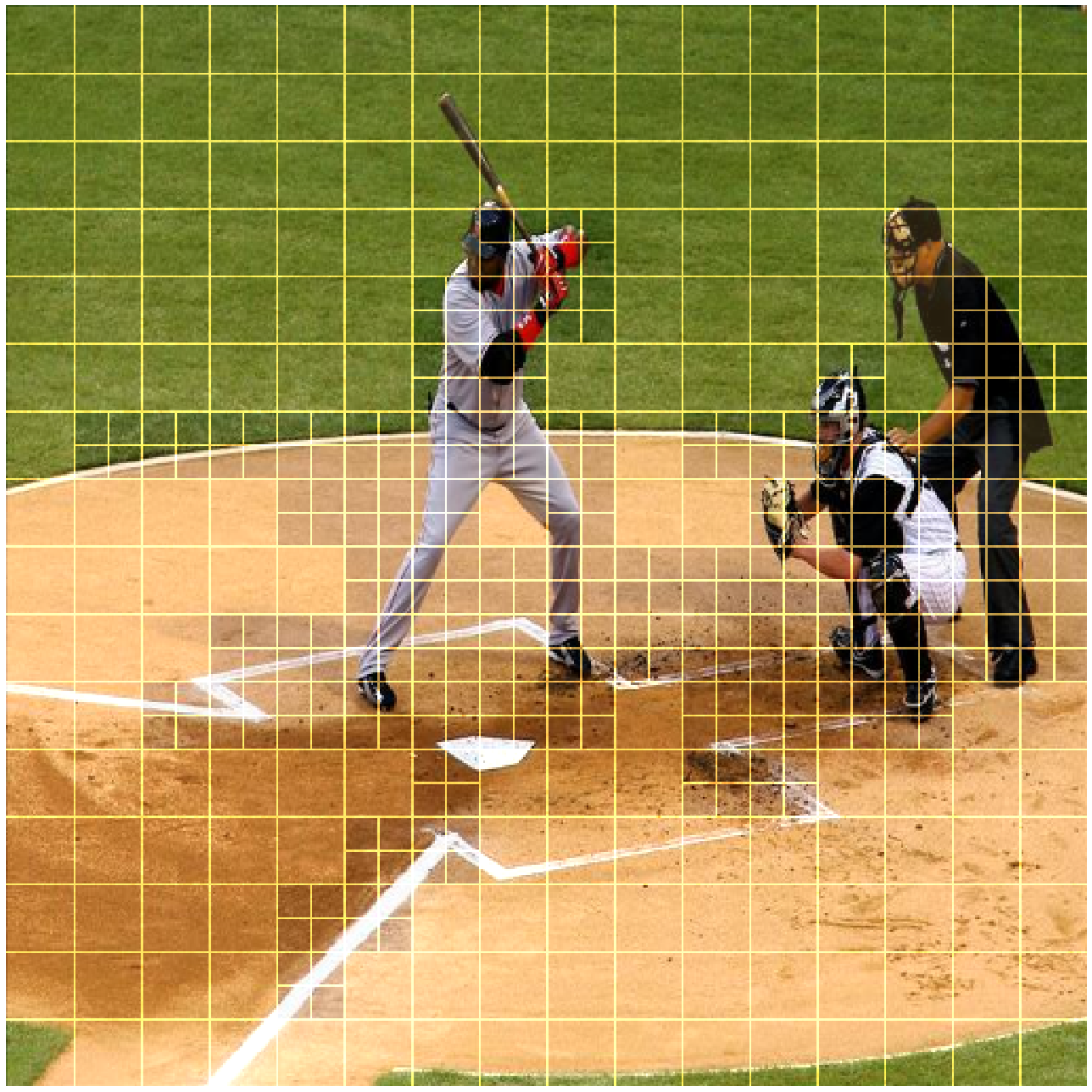}}
    \end{minipage}
    \begin{minipage}[b]{0.155\textwidth}
    \centering
    \fbox{\includegraphics[width=\textwidth, height=\textwidth]{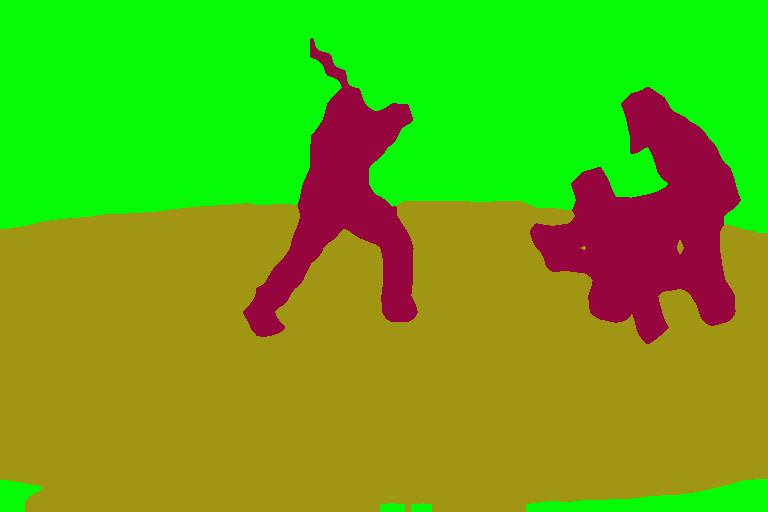}}
    \end{minipage}
    \begin{minipage}[b]{0.155\textwidth}
    \centering
    \fbox{\includegraphics[width=\textwidth, height=\textwidth]{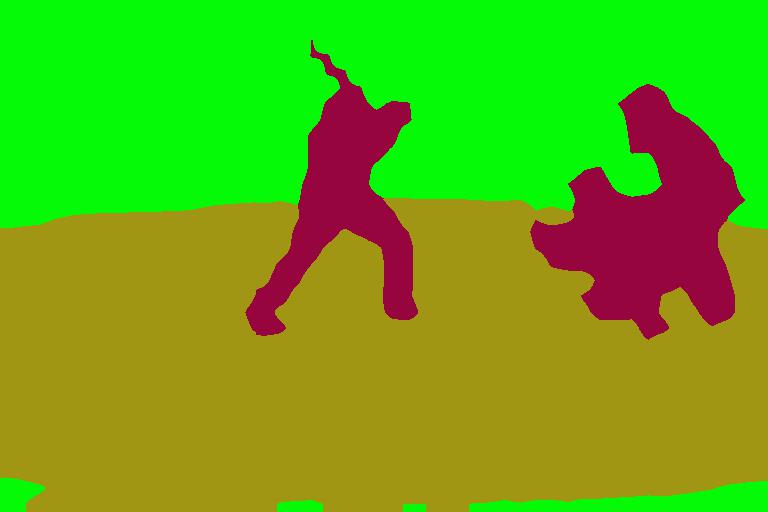}}
    \end{minipage}
    {\small ADE\_val\_00000109}
    
    \renewcommand\idx{503}
    \begin{minipage}[b]{0.155\textwidth}
    \centering
     \vspace{0.023cm}
    \fbox{\includegraphics[width=\textwidth]{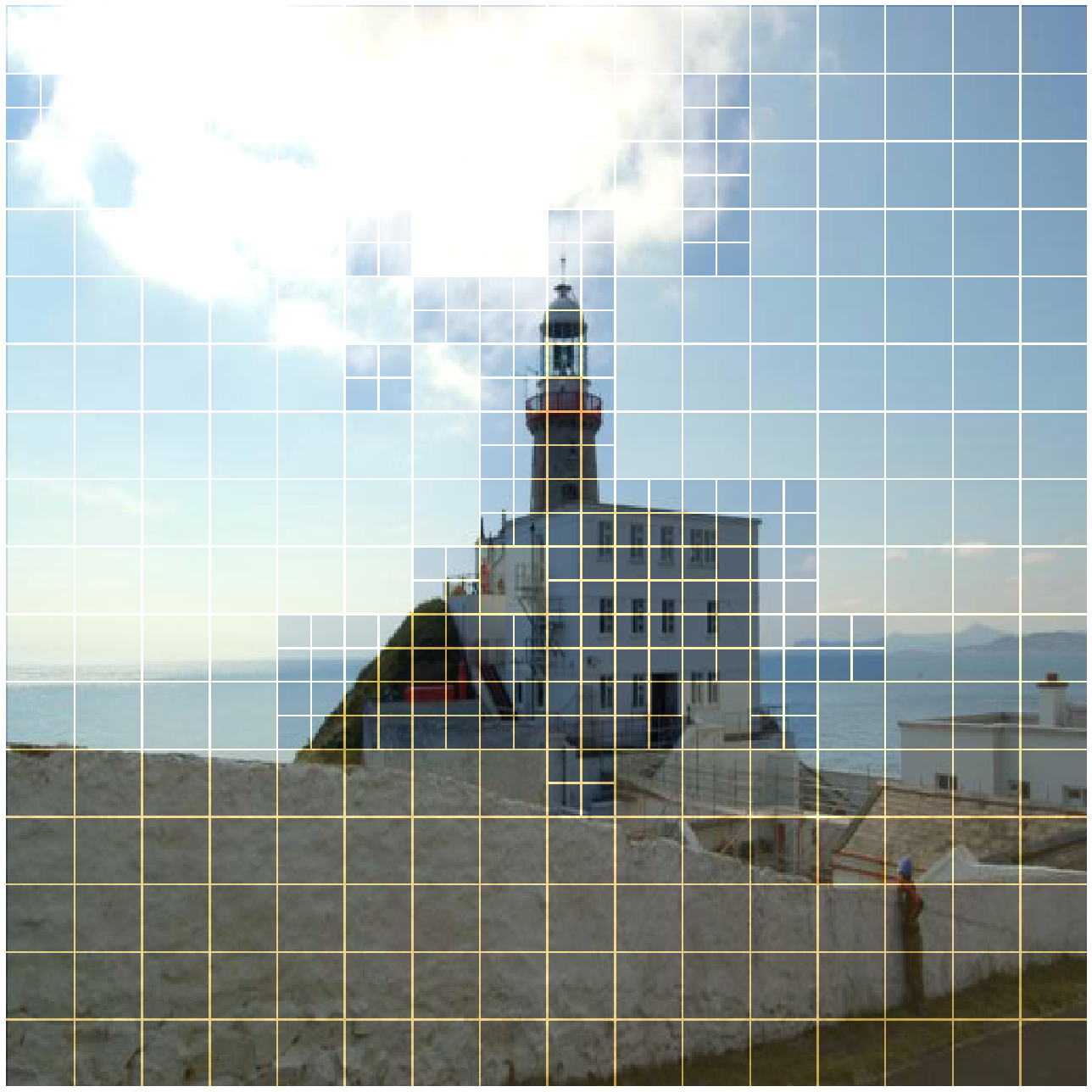}}
    \end{minipage}
    \begin{minipage}[b]{0.155\textwidth}
    \centering
    \fbox{\includegraphics[width=\textwidth, height=\textwidth]{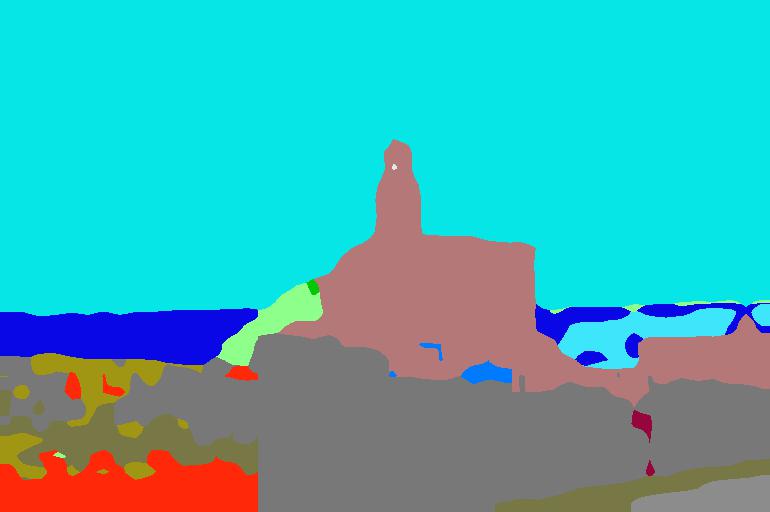}}
    \end{minipage}
    \begin{minipage}[b]{0.155\textwidth}
    \centering
    \fbox{\includegraphics[width=\textwidth, height=\textwidth]{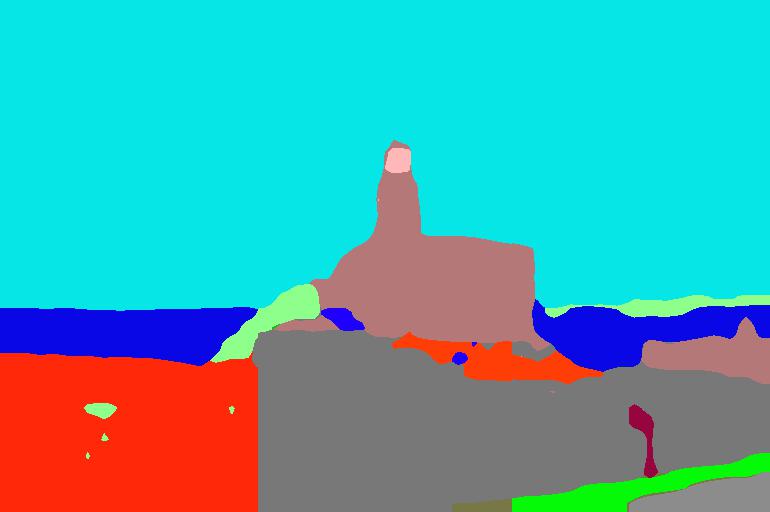}}
    \end{minipage}
    {\small ADE\_val\_00000503}

    \renewcommand\idx{584}
    \begin{minipage}[b]{0.155\textwidth}
    \centering
     \vspace{0.023cm}
    \fbox{\includegraphics[width=\textwidth]{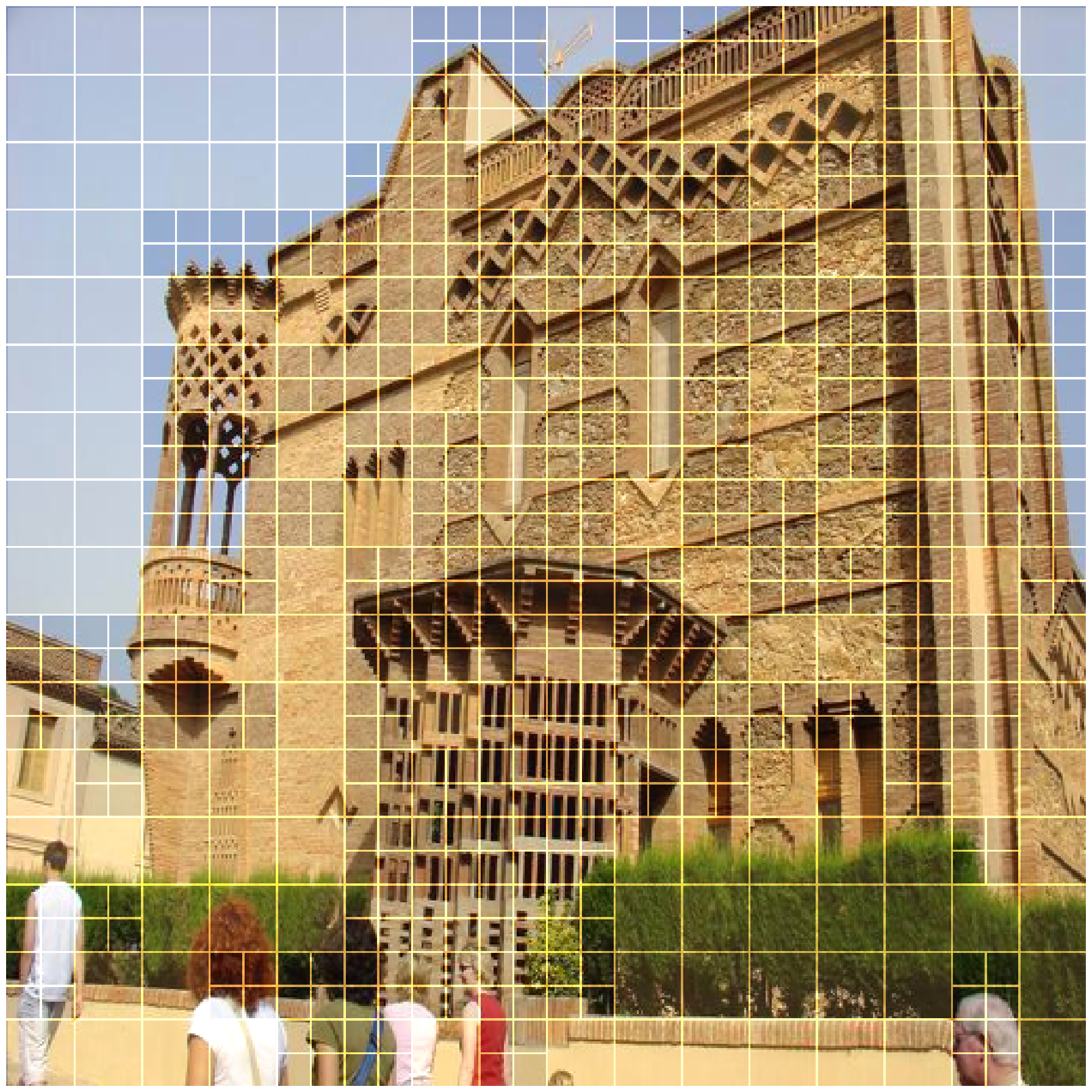}}
    \end{minipage}
    \begin{minipage}[b]{0.155\textwidth}
    \centering
    \fbox{\includegraphics[width=\textwidth, height=\textwidth]{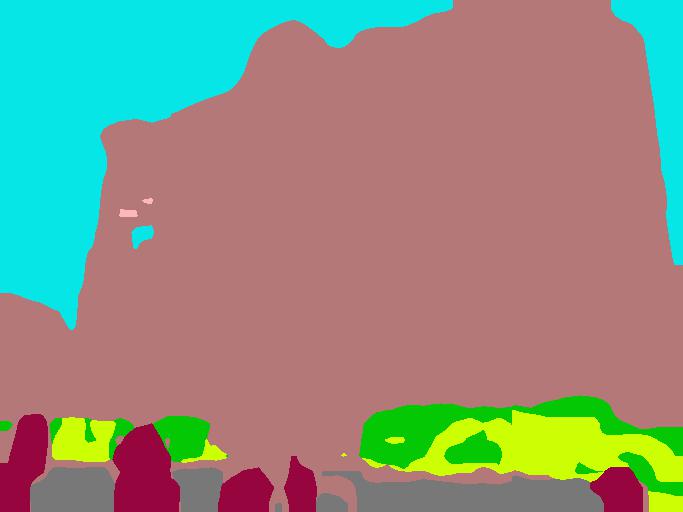}}
    \end{minipage}
    \begin{minipage}[b]{0.155\textwidth}
    \centering
    \fbox{\includegraphics[width=\textwidth, height=\textwidth]{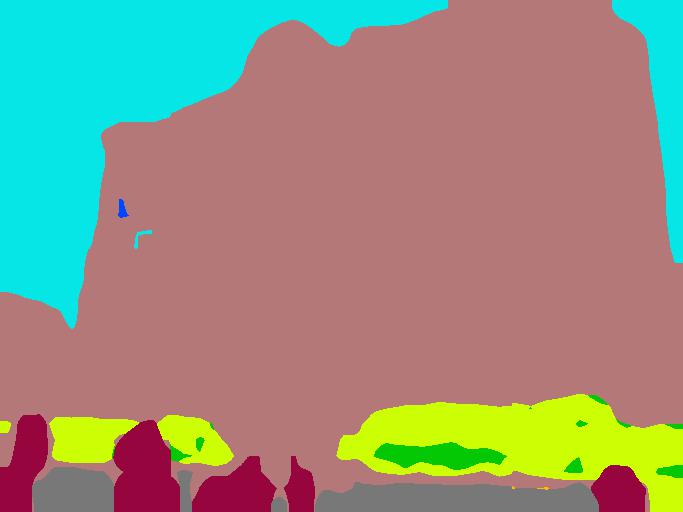}}
    \end{minipage}
    {\small ADE\_val\_00000\idx}

    \renewcommand\idx{107}
    \begin{minipage}[b]{0.155\textwidth}
    \centering
     \vspace{0.023cm}
    \fbox{\includegraphics[width=\textwidth]{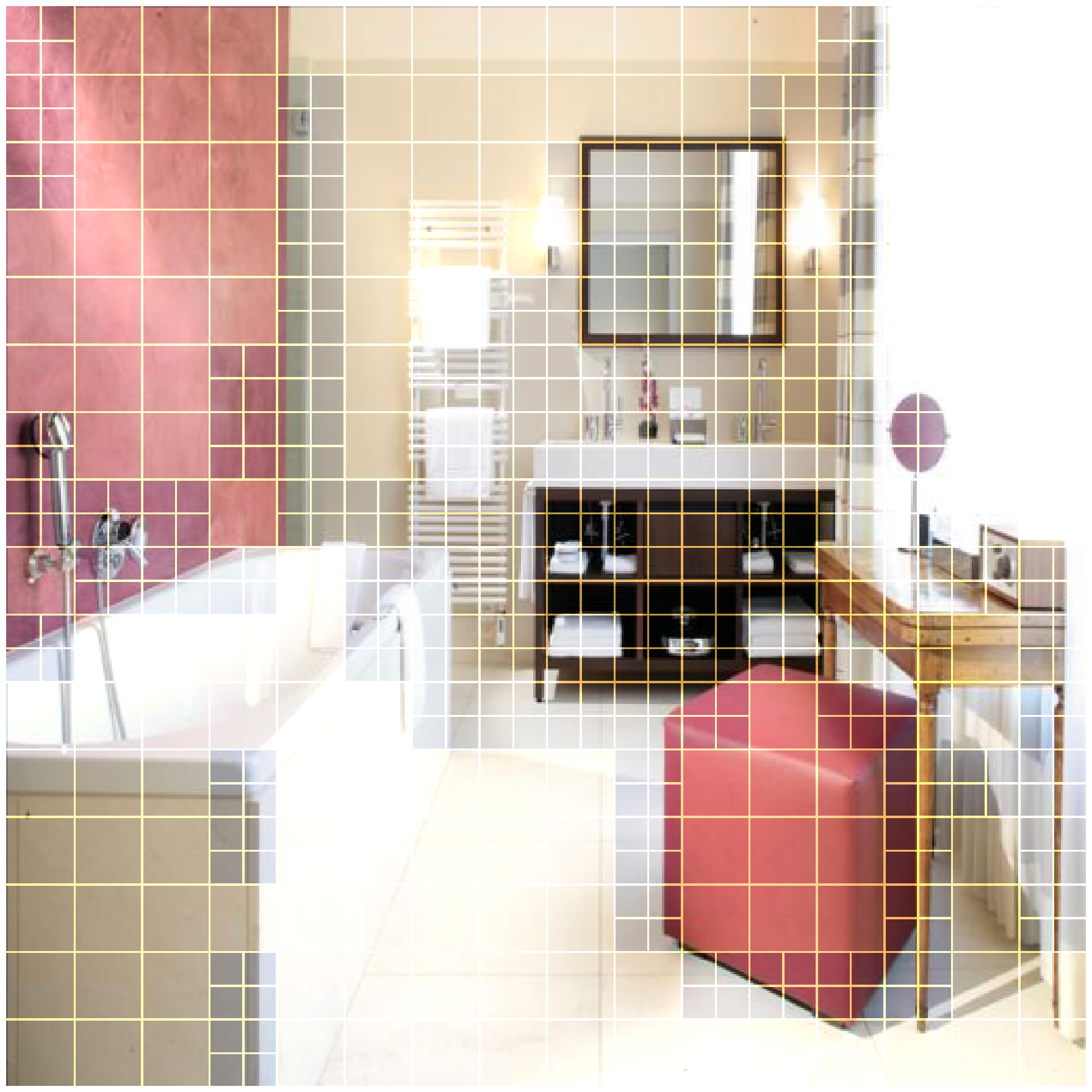}}
    \end{minipage}
    \begin{minipage}[b]{0.155\textwidth}
    \centering
    \fbox{\includegraphics[width=\textwidth, height=\textwidth]{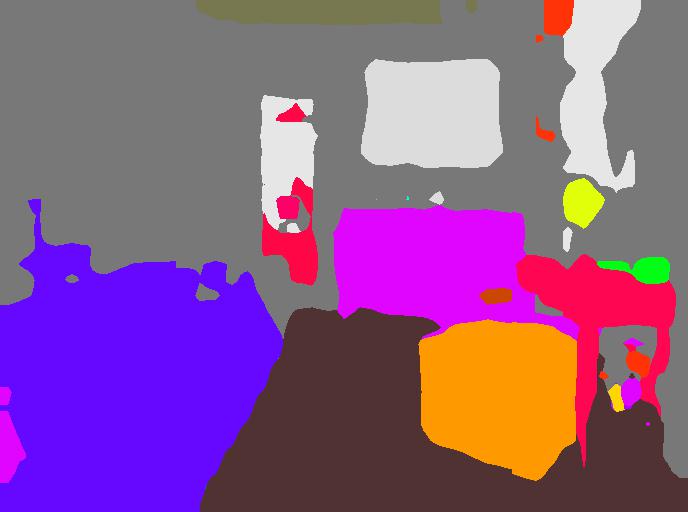}}
    \end{minipage}
    \begin{minipage}[b]{0.155\textwidth}
    \centering
    \fbox{\includegraphics[width=\textwidth, height=\textwidth]{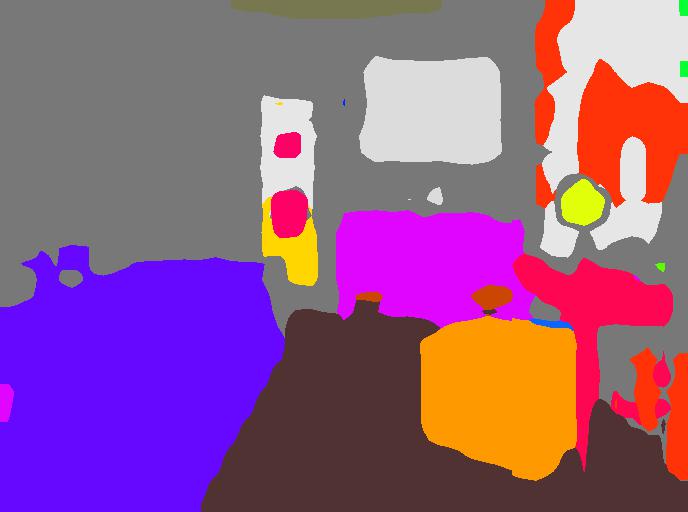}}
    \end{minipage}
    {\small ADE\_val\_00000\idx}

    \renewcommand\idx{113}
    \begin{minipage}[b]{0.155\textwidth}
    \centering
     \vspace{0.023cm}
    \fbox{\includegraphics[width=\textwidth]{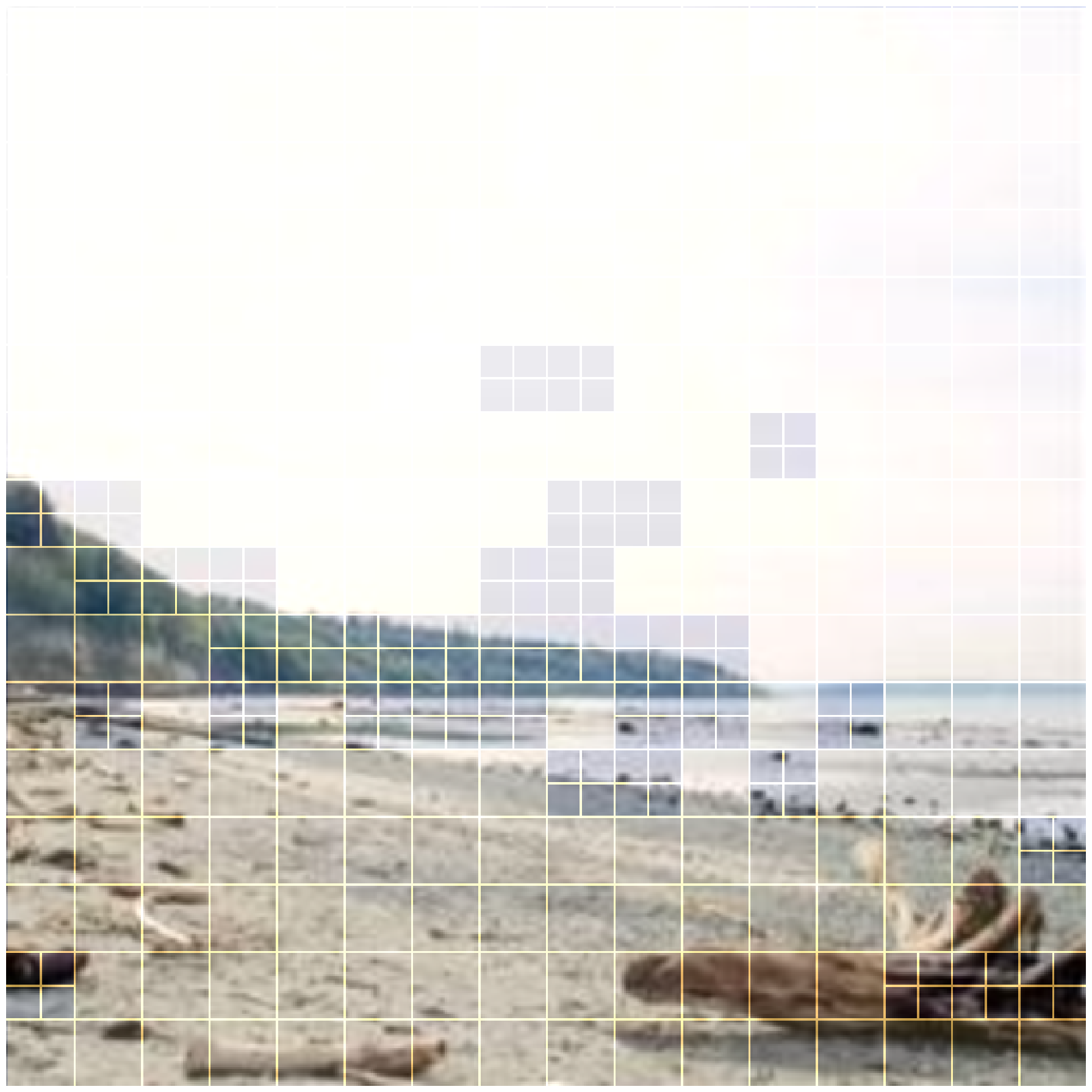}}
    \end{minipage}
    \begin{minipage}[b]{0.155\textwidth}
    \centering
    \fbox{\includegraphics[width=\textwidth, height=\textwidth]{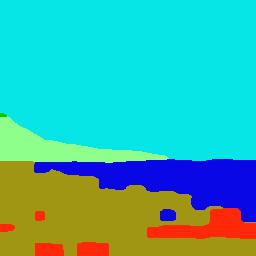}}
    \end{minipage}
    \begin{minipage}[b]{0.155\textwidth}
    \centering
    \fbox{\includegraphics[width=\textwidth, height=\textwidth]{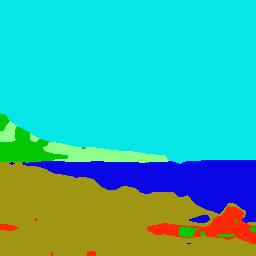}}
    \end{minipage}
    {\small ADE\_val\_00000\idx}

    \caption{Non-curated qualitative results on the segmentation experiments. We display the mixed-scale mask output by the gate (\textit{left}), the final segmentation map output by our MSSeg-S/16,32 trained with target $g^{\ast} = 0.25$ (\textit{middle}) and the segmentation map output by the corresponding backbone baseline Seg-S/16}
    \label{fig:segout}
\end{figure}

Finally, we report extended results in \hyperref[tab:extendedseg]{Table \ref{tab:extendedseg}} (same results as \hyperref[fig:segmentation]{Figure \ref{fig:segmentation} (a)} in the main text but with additional latency results) and visualize some additional qualitative outputs in \hyperref[fig:segout]{Figure \ref{fig:segout}}.

\begin{table}[!thb]
  \resizebox{0.5\textwidth}{!}{  
    \begin{tabular}{|c|c||c|c|c|c||c|}
        \hline
         Backbone & $g^{\ast}$ & \# tokens & MACs & CPU time & GPU time & mIoU \\
         &  & avg & x 1e10 & ms & ms & \footnotesize single-scale \\
         \hline
         \rowcolor{mygray} Seg-T/16 {\footnotesize (512px)}  & - & 1024 & 1.04 & 113.68 & 26.5 & 38.1 \\
         \hline
         \multirow{3}{*}{\textbf{MS}Seg-T/16} & 
         0.5 & 655 & 0.56 & 86.12& 25.6 & 37.9\\
          \cline{2-7}
          & 
         0.25 & 565 & 0.46 & 75.96& 25.0 & 37.3 \\
          \cline{2-7}
          & 
         0.1 & 525 & 0.42 & 69.13 & 24.3& 36.8 \\
          \hline
          \hline
         \rowcolor{mygray} Seg-S/16 {\footnotesize (512px)}  & - & 1024 & 3.17 &  252.09 & 30.7 & 45.3 \\
         \hline
         \multirow{3}{*}{\textbf{MS}Seg-S/16} & 
         0.5 & 684 & 1.92 & 184.81 & 29.6 & 44.9\\
          \cline{2-7}
          & 0.25 & 586 & 1.59 & 153.12 & 29.0 & 44.1\\
          \cline{2-7}
          & 
         0.1 & 552 & 1.48  & 144.02 & 28.5 & 43.3 \\
         \hline
    \end{tabular}
    }
    \vspace{0.15cm}
    \caption{Segmentation results from \hyperref[fig:segmentation]{Figure \ref{fig:segmentation} (a)} in the main text with extended timing results on \textbf{(i)} CPU and \textbf{(ii)} GPU (Tesla V100-SXM2-32GB), both reported in milliseconds }
    \label{tab:extendedseg}
    \end{table}

\section{Mixed-scale tokens for non-standard ViTs}
\label{app:reshapes}
In \hyperref[sec:transfer]{Section \ref{sec:transfer}}, we combine a pretrained mixed-scale gate with different ViT backbones. In this section, we describe how we implement these changes in more details.

\subsection{Segmenter}
The \textbf{Segmenter} architecture~\cite{segmenterrepo} is composed of a standard ViT backbone, followed by a small decoder (either linear, or a small transformer head) to generate the segmentation map outputs.
We first replace the ViT backbone by MSViT. We then simply need to recover the original spatial resolution from the stream of mixed-scale tokens at the end of the backbone: 
More specifically, once the transformer backbone has been executed, we replicate every coarse token 4 times, to compensate for the fine token it replaces.
We then feed this sequence of tokens to the decoder, without making any changes to its original architecture.

\subsection{Token pruning}
Most SotA token pruning methods builds off the DeiT architecture, and implement some form of token binary masking, similar to how we use masked  attention (Eq. \ref{eq:maskedatt}).
Thus adding mixed-scale tokenization to these models is straight-forward: For instance, in \textbf{DyViT}, we simply use the binary mask output by the mixed-scale gate as the initial "pruning policy" of the model (instead of the default initialization which a mask of all ones).
In \textbf{EViT}, the tokens are sorted by decreasing class-attention and a fixed ratio of the lower tokens is pruned in certain layers. We simply apply the same sort-and-prune operation to the mixed-scale mask as the one applied to the tokens and propagate them to the next layer.

\subsection{Hierarchical Transformers}
Unlike ViT, hierarchical transformers such as Swin integrate multiple scale. We denote by $s_\ell$ the scale of the $\ell$-th block in Swin; where each block is a sequence of transformer layers, and in practice $s_\ell = 4 \times 2^{\ell - 1}$. 
The transition from a block to the next is done with a Patch Merging operation: Groups of 4 neighboring tokens are concatenated together then linearly embedded to form a unique token. 
As a result, as we transition through block, the number of tokens decreases (i.e., the patch scale increases) and the number of channels increases, similar to the usual CNN architecture design. 
In addition, Swin implements local attention is computed across windows of $w \times w$ tokens ($w = 7$).

Given a pretrained mixed-scale gate with coarse scale $S_c$, we first run it on the input image: This yields a binary decision for each $S_c \times S_c$ coarse patch in the image. 
We use this binary mask to guide the flow of tokens through the Swin transformer: Every input token that falls in a fine scale region follows the standard Swin paradigm. For the rest of the tokens (coarse scale regions), we feed them to a simple linear embedding, and reintegrate them to the flow of fine tokens in the $\ell$-th block such that $s_\ell = S_c$.

In summary, the mixed-scale gate decides whether a token should be processed at a fine-grained level (early layers of the Swin transformer with small receptive field). The gain in computational cost comes from \textbf{(i)} coarse tokens skipping FFNs in the early layers, and \textbf{(ii)} due to the absence of coarse tokens in the early layers some local attention windows are empty, hence can be entirely skipped.

Finally, there is an interesting interaction between the base patch scale $s_1$, the attention window scale $w = 7$ and the coarse scale of the gate ($S_c = s_\ell$), as they all impact the scale of the tokens. 
In our experiments, we consider varying the parameter $\ell$ and show that it directly impacts the MACs-accuracy trade-off.

\section{Training dynamics of adaptive trimming}
\label{app:trimvsfullmask}
In \hyperref[sec:mixedrestraining]{Section \ref{sec:mixedrestraining}} we introduce the adaptive trimming strategy  (\textbf{AT}) for reducing training overhead. 
In this section we analyze how it impacts the gradients received by the gate. 
For simplicity, let us consider a simple transformer with a single attention layer and a class token at index 0. 

\subsection{Without Adaptive Trimming.}
The full process of MSViT can be summarized as:

\textbf{1. Obtain coarse level mask}
\begin{align}
\forall j \in [1, N_{S_{\idxcoarse}}],\  &m_j = \text{GumbelSigmoid}(g_{\psi}(x_j))\\
&\overline{m}_j = \text{STE}(m_j)
\end{align}

\textbf{2. Deduce fine level mask}
\begin{align}
\forall i \in [N_{S_{\idxcoarse}}& + 1, N_{S_{\idxcoarse}} + N_{S_{\idxfine}} ],\  \overline{m}_i = 1 -  \overline{m}_{C(i)}\label{eq:map}
\end{align}

\textbf{3. Embed patches and add position embeddings}\\

\textbf{4. Masked attention}
\begin{align}
z_i &= e^{Q_0 K_i^T}\\
y_0 &= \sum_{i=1}^{N = N_{S_\idxcoarse} + N_{S_\idxfine}} \frac{\overline{m}_i
 z_i}{\sum_{p=1}^N \overline{m}_p z_p } V_i 
\end{align}
where $Q, K, V$ denotes the query, key and value embeddings of the tokens $x'$; and $C$ is the mapping from fine to coarse tokens.\\

\textbf{5. Feed $y_0$ to linear classification head}\\

For simplicity, we will write the partition function as $Z(\psi) = \frac{1}{\sum_{p=1}^N \overline{m_p} z_p}$.
Using the link between coarse and fine tokens from \hyperref[eq:map]{Equation \ref{eq:map}} we can decompose the equation in \textbf{step 4} as follows:

\begin{align}
y_0 &= Z(\psi) \sum_{i=1}^N \overline{m}_i z_i V_i\\
y_0 &= Z(\psi) \left( \sum_{j=1}^{N_{S_{\idxcoarse}}} \overline{m}_j z_j V_j +  \sum_{i=N_{S_{\idxcoarse}} + 1}^{N} ( 1 - \overline{m}_{C(i)}) z_i V_i  \right)\\
y_0 &= Z(\psi)  \left[\sum_{j=1}^{N_{S_{\idxcoarse}}} \underbrace{\overline{m}_j \left( z_j V_j - \sum_{\substack{i=N_{S_{\idxcoarse}} + 1\\ C(i) = j}}^{N} z_i V_i \right)}_{A_j(\psi)} + \underbrace{\sum_{i=N_{S_{\idxcoarse}} + 1}^{N} z_i V_i}_{B} \right] \label{eq:final}
\end{align}

Because of \textit{straight-through}, we have $\frac{\partial \overline{m_j}}{\partial \psi} = \frac{\partial m_j}{\partial \psi}$, therefore every token contributes to the gradient with respect  to the gate parameters, $\frac{\partial y_0}{\partial \psi}$, even if the token wass masked with $\overline{m_j} = 0$ in the forward pass.
In particular, the fine and coarse tokens of each region directly interact through $A_j(\psi)$, where their attention value (wrt. the class token) are compared to each other. 

\subsection{With Adaptive Trimming}
With adaptive trimming, we reorder the tokens according to their value  of $\overline{m}_i$ and trim the number of tokens to the maximum sequence length in the batch in \textbf{step 4}.
This essentially mean that some terms will now be omitted from both the forward \textit{and backward pass} in \hyperref[eq:maskedatt]{Equation \ref{eq:maskedatt}} and  in \hyperref[eq:final]{Equation \ref{eq:final}}. As a result, these terms also disappear from the gradients of $Z(\psi)$ and $A_j(\psi)$.
In particular, if the coarse token $j$ is active and all corresponding fine tokens are trimmed, then:
\begin{align}
\frac{\partial A_j(\psi)}{\partial \psi} = \frac{\partial m_j}{\partial \psi} z_j V_j \label{eq:before}
\end{align}

In the opposite case (fine scale active and corresponding coarse token is trimmed) then:
\begin{align}
\frac{\partial A_j(\psi)}{\partial \psi} = - \frac{\partial m_j}{\partial \psi}  \sum_{\substack{i=N_{S_{\idxcoarse}} + 1\\ C(i) = j}}^{N} z_i V_i \label{eq:after}
\end{align}

In other words, in the masking scenario (\hyperref[eq:final]{Equation \ref{eq:final}}) the transition from coarse to fine scale for a token is smoothly captured in the straight-through gradients   $\frac{\partial m_j}{\partial \psi}$. 
In contrast, with adaptive trimming, flipping from coarse to fine tokens may sometimes lead to a sudden change in gradients from \eqref{eq:before} to \eqref{eq:after}.  
Thus Adaptive trimming leads to a  noisier optimization process. However, as we will see in the next section, this is not a significant issue in practice.

\subsection{Adaptive trimming in practice}
\label{app:trimming}
In \hyperref[tab:trainingtimes]{Table \ref{tab:trainingtimes}}, we report a comparison of training times for MSViT, with and without the adaptive token trimming (\textbf{AT}) strategy introduced in  \hyperref[sec:mixedrestraining]{Section \ref{sec:mixedrestraining}}. 
As expected, AT leads to faster training times, in particular for lower values of the gate sparsity target $g^{\ast}$. 
Furthermore, in practice  we observe that AT generally yields comparable trade-off (or only incurs a minor drop in accuracy for comparable MAC counts), which is why we make it the default for all training runs in our main experiments.

\begin{table}[t]
    \centering
\renewcommand{\arraystretch}{1.1}
  \resizebox{0.48\textwidth}{!}{  
    \begin{tabular}{|c|c||c|c|c|c|}
    \hline
         \multicolumn{2}{|c||}{Model} & train time & \# tokens & GMACs & Acc. \\
         \multicolumn{2}{|c||}{{\small (ViT-S/16 backbone) }} &  [min] & (average) & & [\%] \\
         \hline
         ViT & in = 224 & 29.4 & 196 & 4.60 & 82.02\\
         \hline
         \hline
         \parbox[t]{2mm}{\multirow{4}{*}{\rotatebox[origin=c]{90}{\small Mixed-Scale}}}  & $g{\ast} = 0.5$, \textbf{AT} & 31.8 & 147 & 3.43 & 81.53 \\
         \cline{2-6}
          & $g{\ast} = 0.5$, \textbf{full} & 36.0 & 155 & 3.62 & 81.71\\
         \cline{2-6}
         \noalign{\vskip\doublerulesep
         \vskip-\arrayrulewidth}
         \cline{2-6}
         & $g{\ast} = 0.1$, \textbf{AT}& 28.8 &  117 & 2.73 & 80.63 \\
         \cline{2-6}
         & $g{\ast} = 0.1$, \textbf{full}&  36.0 & 132 & 3.09 & 80.96\\
         \hline
    \end{tabular}
    }
    \caption{Average training time per epoch (in minutes) for our mixed-scale MSViT-S/16, with (\textbf{AT}) and without (\textbf{full}) adaptive trimming during training. 
    In practice, ATP leads to faster training time, and only a minor drop in accuracy for comparable MAC count.
    We also report the original VIT backbone timings for reference.
    }
    \label{tab:trainingtimes}
\end{table}

\section{Additional ablation experiments}
\label{app:gbas}
\subsection{Benefits of a dynamic gate}
\label{app:robust}
\begin{figure}[t]
    \begin{center}
    \includegraphics[width=0.5\textwidth,trim={0 0 7cm 0},clip]{figures/robustness_illustration_sorted.png}
    
    \small \textbf{(a)} Example gate outputs given random image zooms and shifts.
    \end{center}

    \includegraphics[width=0.5\textwidth]{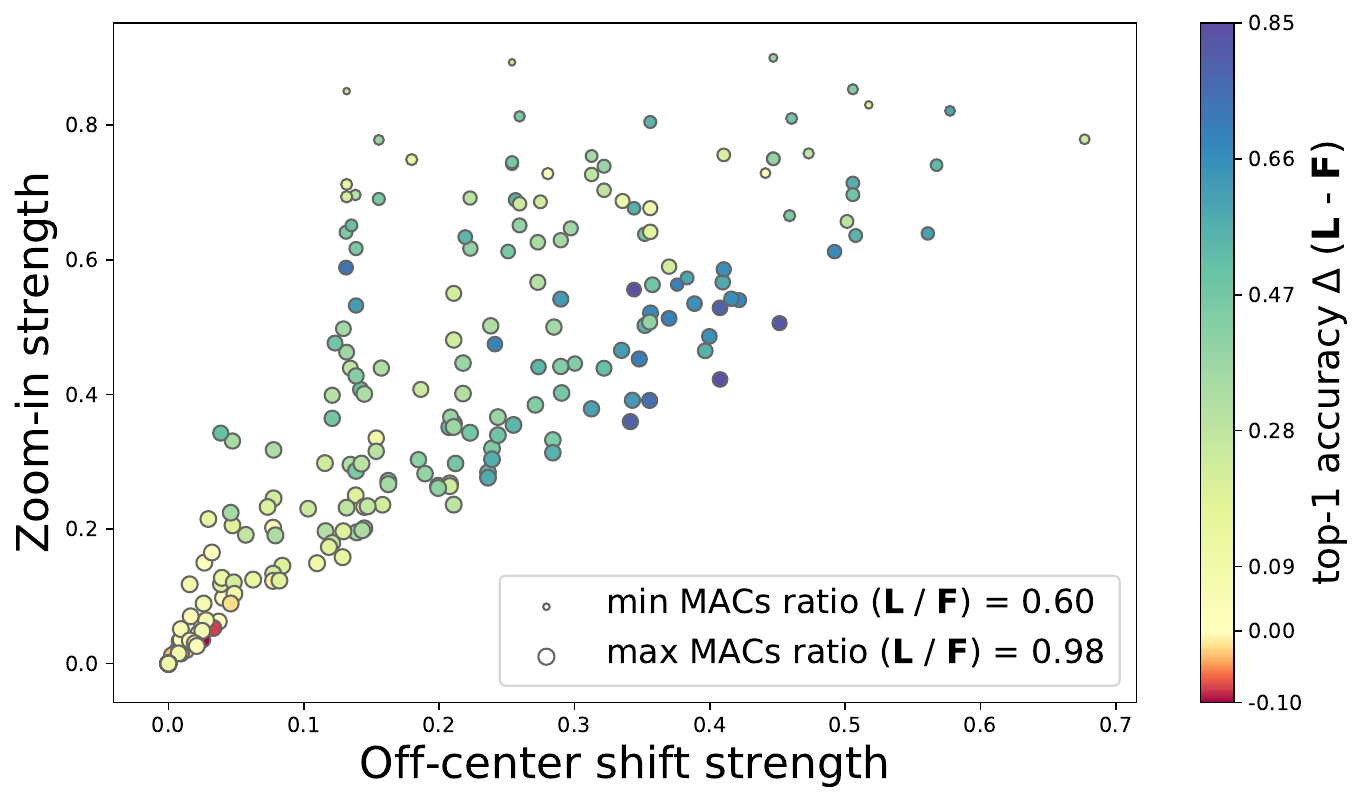}
    \small \textbf{(b) } Each point corresponds to a different cropping transform applied to all images of the validation set, and both models have similar starting performances (83.17 accuracy at 8.50 GMACs for \textbf{L}; 83.06 at 8.75GMACs for \textbf{F}).
    The colors encode accuracy improvement of \textbf{L} compared to \textbf{F} (the bluer the better), and the dot size encodes the efficiency improvement (the smaller the better) of the learned gate over the fixed gate.

    \caption{Performance of a learned gate model (\textbf{L}) versus a fixed radial masking pattern (\textbf{F}). 
    We observe that in most scenarios \textbf{L} provides better accuracy and automatically adapts its computational cost accordingly: For instance, highly zoomed-in images tend to contain more background/flat color patches, which are set to coarse scale by the learned gate, leading to lower MACs.}
    \label{fig:robustness}
\end{figure}

As described in \hyperref[sec:dynamic]{Section \ref{sec:dynamic}}, the learned  dynamic gate in MSViT is able to adapt the model's computational cost based on the input image content, in contrast to using a fixed mixed-scale pattern.  
This is illustrated in \hyperref[fig:robustness]{Figure \ref{fig:robustness}}: 
Here, we generate several random geometric shifts of the validation set, and evaluate both a fixed and learned gate model. We then report in \hyperref[fig:robustness]{Figure \ref{fig:robustness} (b)} their difference in accuracy (color of the scatter plot) and in MAC counts (size of the scatter plot).
We observe that:
\begin{itemize}
\item (i) The learned gate model generally outperforms the fixed gate one and this is more pronounced when the random transformation has a strong off-center shift; In fact, in those cases, the prior of the fixed gate that objects lie in the center of the image breaks.
\item (ii) The fixed scale selection pattern leads to computational
waste when applied on these simple affine geometric shifts that mimic more realistic ”in-the-wild” image inputs. In fact the computational cost of the fixed gate model is constant; while the cost of the learned gate model is significantly reduced when we have strong shifts, as they generally lead to more background regions present in the image, as visualized in \hyperref[fig:robustness]{Figure \ref{fig:robustness} (a)}.
\end{itemize}

\subsection{Generalized Batch Shaping loss}
\begin{figure}[!tbh]
    \includegraphics[width=0.5\textwidth]{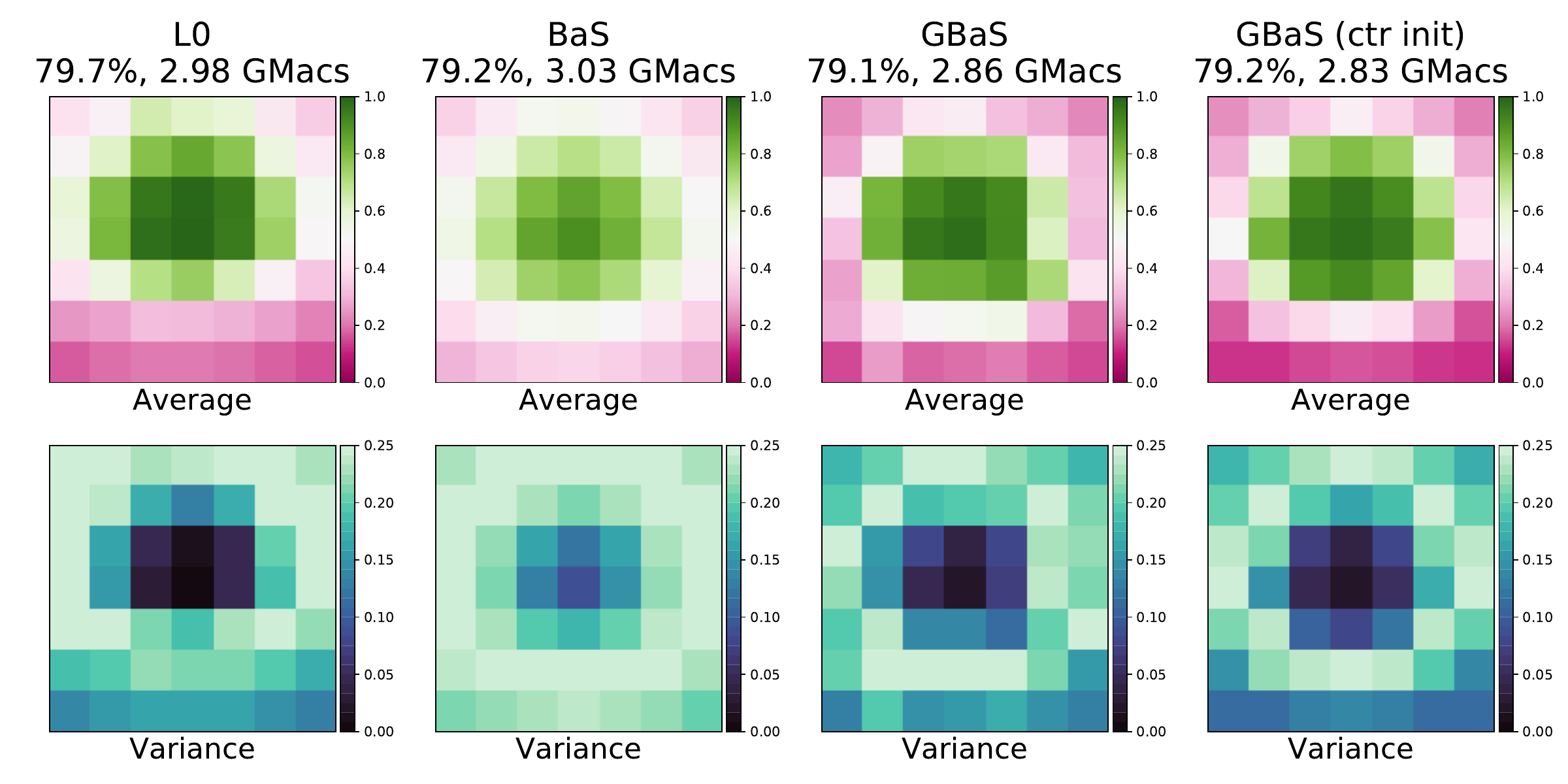}
    
    {\small \textbf{(a)} Average (top row) and variance (bottom row) of the learned scale selection masks across the validation set (A value above 0.5 means that  the corresponding image patch will be kept at fine scale) for different gate sparsity losses.}

    \includegraphics[width=0.5\textwidth]{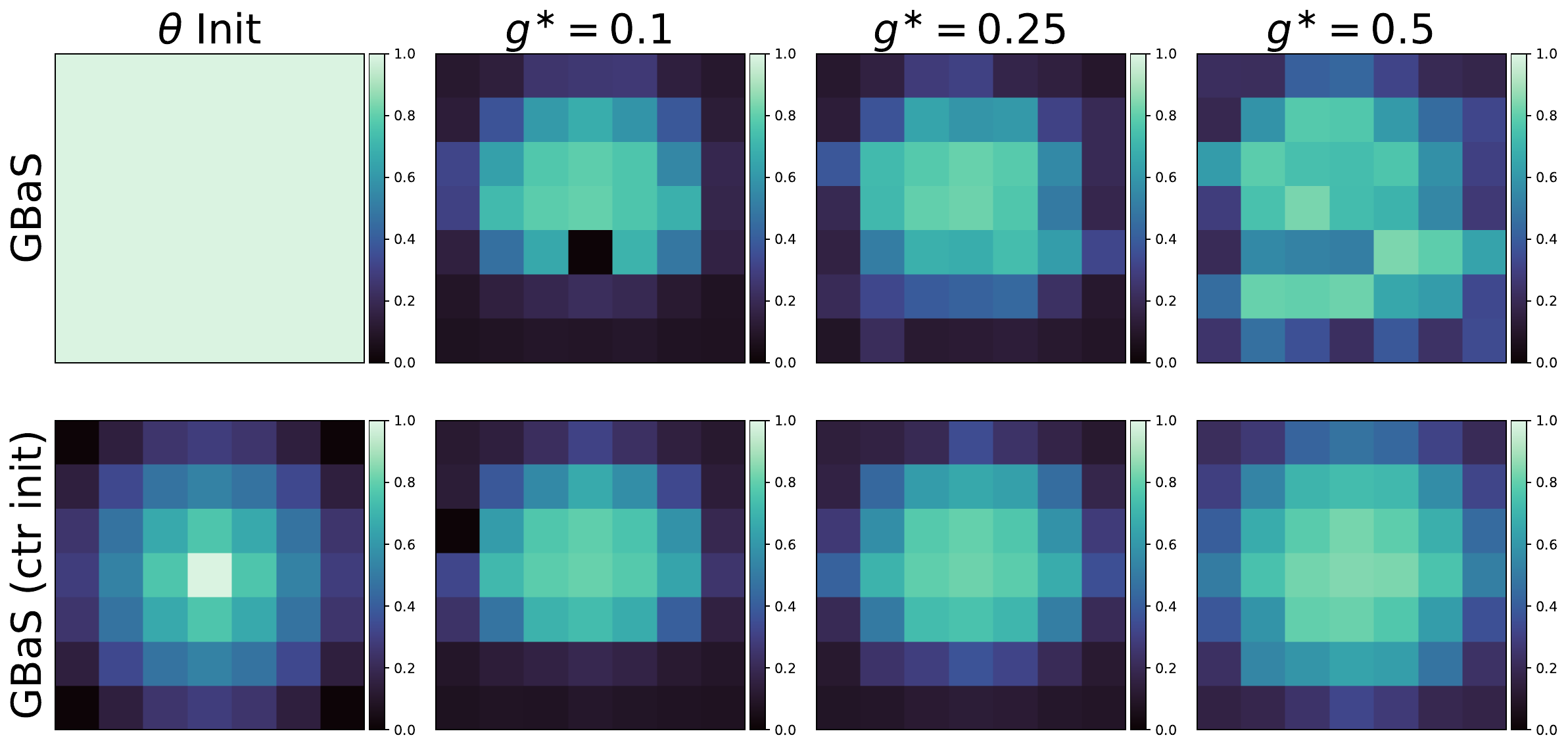}
    
    {\small \textbf{(b)} Prior parameters $\theta$ learned with the \gbsloss{} loss with/without \texttt{ctr init} (\textit{top/bottom}). 
    The first column is initial values of $\theta$.}
    
    \caption{Illustration of the masks learned by the model with light crops data augmentation, leading to little to no shift between the train and test distribution of the tokens input to the gate}
    \label{fig:light}
\end{figure}

In \hyperref[fig:light]{Figure \ref{fig:light}}, we report the counterpart of \hyperref[fig:hyperpriorstrong]{Figure \ref{fig:hyperpriorstrong}} for light croppings data augmentations. 
As mentioned in the main text, in that setting, there is little to no shift in spatial distribution between train and test time. 
As a result, all gate losses manage to capture the prior that objects tend to lie in the center of the image in ImageNet (see \hyperref[fig:light]{Figure \ref{fig:light} (a)}). 
Similarly, for \gbsloss{}, even without dedicated initialization the learned priors also fit the central locality insight (\hyperref[fig:light]{Figure \ref{fig:light} (b)}). 
All losses perform similarly in that setting, and the fast-converging L0 loss is even able to outperform \bsloss{} and \gbsloss{} in \hyperref[fig:lossesaccuracy]{Figure \ref{fig:lossesaccuracy} (b)}.

\subsection{Rescaling the position embeddings with lienar interpolation}
\label{app:resizetrick}

\begin{figure}[!thb]
    \centering
    \includegraphics[width=0.5\textwidth]{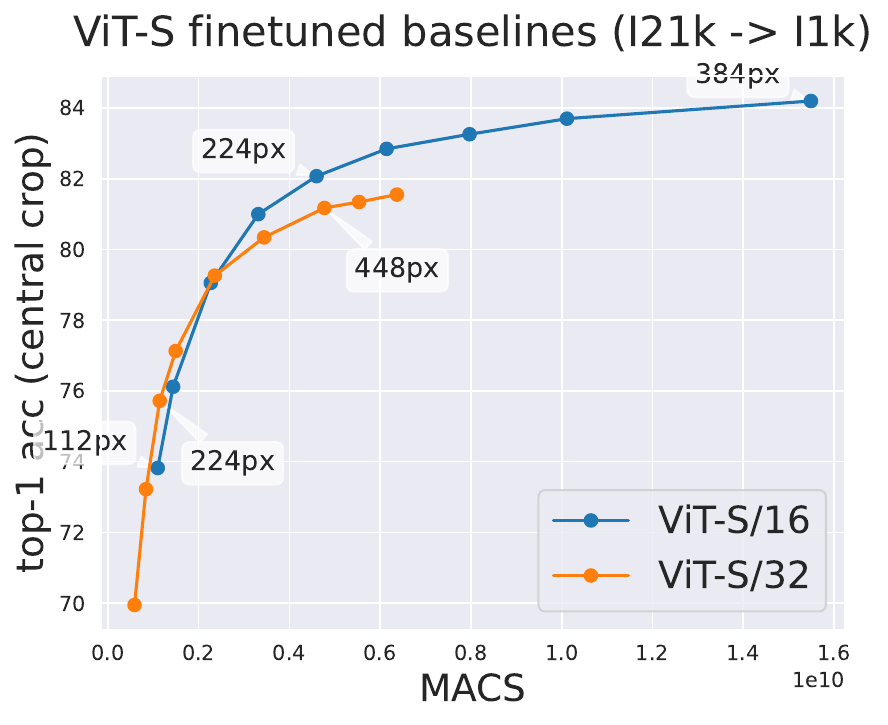}
    \caption{Comparison of the performance of ViT-S with patch size 32 and patch size 16, trained for different input image sizes using the linear interpolation rescaling trick of the position embeddings. While ViT-S/16 generally yields better trade-offs, the trend starts to invert itself around the threshold of 100 tokens}
    \label{fig:patchingtrick}
\end{figure}

In \hyperref[fig:patchingtrick]{Figure \ref{fig:patchingtrick}} we show that, when using the standard ViT finetuning pipeline with the linear interpolation of position encodings leads to an interesting observation: For a low number of tokens,  ViT-S/32 on image size $X$ (scenario \textbf{A}) performs better than a ViT-S/16 model on image size $X // 2$ (scenario \textbf{B}), despite them having the same number of tokens. 

\begin{figure}[!thb]
    \centering
    \includegraphics[width=0.48\textwidth]{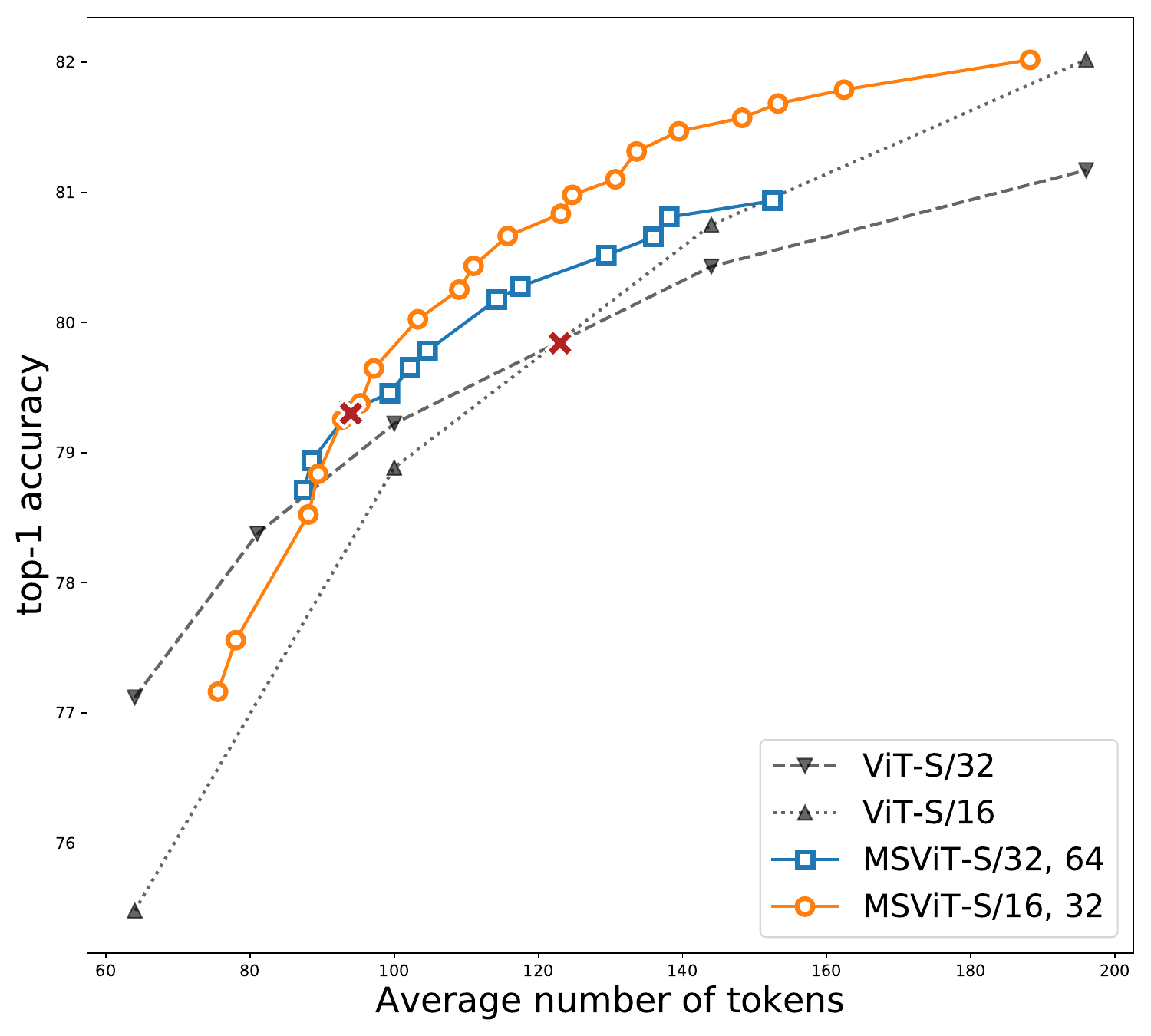}
    \caption{Comparing the effect of patch scale versus input image size: In terms of number of tokens, increasing the patch or decreasing the input image size are equivalent; However, the initial token embeddings and resizing differ in these two settings; As we can see from this plot, this can lead to large differences in the low token regimes for the standard ViT backbone ( $\sim$ 130 tokens, indicated by {\textcolor{red}{X}}), and we see the same shift starting to appear for MSViT models around 90 tokens. }
    \label{lastfig}
\end{figure}

We then investigate whether this behavior also occurs in MSViT. 
In \hyperref[lastfig]{Figure \ref{lastfig}}, we report results for the setting described in the main paper: ViT-S/16 backbone at different input image sizes, and MSViT-S/\{16, 32\} for different gate loss objectives.
In addition, we also report results on ViT-S/32 and MSViT-S/\{32, 64\}, run on a subset of the search space.

As we see from the figure, the impact of patch size is in fact the same as in ViT: In the low regime of the number of tokens (around 95), MSViT-S/32, 64 ran on larger images starts to outperform ViT-S/16, 32.  
This also indicates that the token embedding and resizing algorithm may impact the model's accuracy in for a low number of tokens, and motivates further investigation of this behavior for vision transformers in general.

\end{document}